\documentclass[Afour,sageh,times]{sagej}

\usepackage{moreverb,url}

\usepackage{algorithm}
\usepackage{algorithmic}
\usepackage{amsfonts}
\usepackage{amsmath}
\usepackage{subcaption}

\usepackage{makecell}
\usepackage[table]{xcolor}

\newcommand{\vect}[1]{{\boldsymbol{#1}}}

\newcommand{\Real}{\mathbb{R} }
\newcommand{\E}{\mathbb{E} }

\newcommand{\cost}{\mathcal{C}}
\newcommand{\KL}{D_{\textrm{KL}}}

\newcommand{\new}{\textrm{new}}
\newcommand{\nul}{\textrm{null}}

\usepackage{graphicx}  
\usepackage{url}

\usepackage[colorlinks,bookmarksopen,bookmarksnumbered,citecolor=red,urlcolor=red]{hyperref}

\newcommand\BibTeX{{\rmfamily B\kern-.05em \textsc{i\kern-.025em b}\kern-.08em
T\kern-.1667em\lower.7ex\hbox{E}\kern-.125emX}}

\def\volumeyear{2019}
\begin{document}

\runninghead{Osa}

\title{Multimodal Trajectory Optimization for Motion Planning}

\author{Takayuki Osa\affilnum{1}\affilnum{2}}

\affiliation{\affilnum{1}Kyushu Institute of Technology, Japan\\
\affilnum{2}RIKEN Center for Advanced Intelligence Project, Japan}

\corrauth{Takayuki Osa, Kyushu Institute of Technology
Behavior Learning Systems Loboratory,
Hibikino 2-4,
Wakamatsu, Kitakyushu, Fukuoka,
808-0135, Japan.}

\email{osa@brain.kyutech.ac.jp}

\begin{abstract}
	Existing motion planning  methods often have two drawbacks: 1) goal configurations need to be specified by a user, and 2) only a single solution is generated under a given condition.
	In practice, multiple possible goal configurations exist to achieve a task.
	Although the choice of the goal configuration significantly affects the quality of the resulting trajectory, it is not trivial for a user to specify the optimal goal configuration.
	In addition, the objective function used in the trajectory optimization is often non-convex,
	and it can have multiple solutions that achieve comparable costs.
	In this study, we propose a framework that determines multiple trajectories that correspond to the different modes of the cost function.
	We reduce the problem of identifying the modes of the cost function to that of estimating the density induced by a distribution based on the cost function.
	The proposed framework enables users to select a preferable solution from multiple candidate trajectories,
	thereby making it easier to tune the cost function and obtain a satisfactory solution.
	We evaluated our proposed method with motion planning tasks in 2D and 3D space.
	Our experiments show that the proposed algorithm is capable of determining multiple solutions for those tasks. 
\end{abstract}


\maketitle

\section{Introduction}
\label{sec:intro}
Motion planning is an essential component in robotics.
When handling an object with a robotic manipulator, it is necessary to plan a smooth and collision free trajectory to perform a desired task.
However, existing motion planning methods often have the following limitations; 1) a motion planner generates only a single solution under a given setting, and 2) goal configurations need to be specified by a user.

\begin{figure}[t]
	\centering
	\includegraphics[width=\columnwidth]{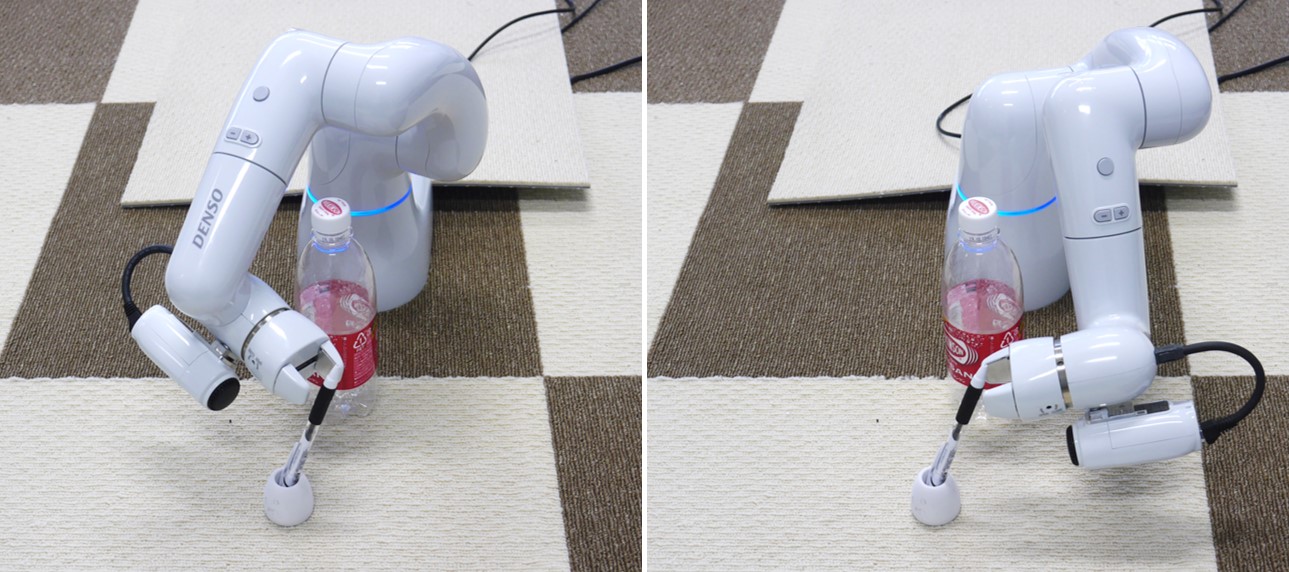}
	\caption{There are often multiple ways to avoid obstacles and reach a target position in practice. In such a case, the cost function for motion planning is often multimodal. Left and right figures show two goal configurations found by our motion planning method for a bottle grasping task.
	} 
	\label{fig:intro}
\end{figure}

In practice, the objective function used in trajectory optimization is often non-convex and has multiple modes.
In such cases, there should be multiple solutions that perform a given task.
For instance, there are multiple ways to avoid an obstacle and  reach the desired position in a scene as shown in Fig.~\ref{fig:intro}.
However, existing methods often ignore this multimodality of the objective function and provide only a single solution.
If a user would like to obtain another solution, she/he needs to tune the objective function or try another initialization of the motion planner.
However, tuning the objective function usually requires expert knowledge on the employed trajectory optimization method. 
Furthermore, when ignoring the multimodality of the objective function, an optimization process can fall into the local optima, which does not exhibit satisfactory performance.
Thus, it is preferable to consider the multimodality of the objective function during optimization so that it generates multiple solutions that correspond different modes.

Additionally, it is not a trivial task to specify a goal configuration in motion planning.
For example, when grasping a cylindrical object, an orientation of the end effector for grasping the given object is not unique as shown in Fig.~\ref{fig:intro}.
Although existing motion planners often require a user to specify a goal configuration, the end effector orientation that leads to the shortest and smoothest collision-free trajectory is not obvious in practice.
Therefore, it is understood that it is necessary to optimize the trajectory including the orientation at the goal point.
Otherwise, a user might need to manually tune the goal configuration by running the motion planning software several times.

In this work, we present the stochastic multimodal trajectory optimization~(SMTO) algorithm that can generate multiple solutions in motion planning. 
Our method estimates the multiple modes of the objective function and determines trajectories that correspond to each mode. 
In the proposed framework, the trajectory including the goal configurations is optimized.
Thus, our method do not require manual tuning of the goal configuration, and the solutions can be obtained even if a given goal configuration has a collision with the environment.
We derived this approach by formulating the trajectory optimization problem as a density estimation problem and introducing importance sampling based on the cost function.
One can interpret that our approach divides the manifold of the trajectories based on the modes of the cost function and finds local solutions in each region. 
The proposed method was evaluated with a 2D-three-link manipulator and manipulators with six and seven degrees-of-freedoms~(DoFs).

In our previous work, we presented a hierarchical reinforcement learning method that learns multiple option policies corresponding to the modes of the reward function~\citep{Osa18}. The method finds the modes of the reward function by performing density estimation with importance sampling.
While the aim of our prior work is to learn a policy, which is defined as the density of actions for a particular state, we extend the concept for the reward-maximization problem to the trajectory optimization in this study. 

Finding multiple solutions gives a user an opportunity to select one of the solutions based on her/his preferences, which is often hard to encode in the objective function.
In addition, our framework can optimize a trajectory, including the orientation at a goal point. 
This framework reduces the burden of users to tune and specify the orientations of end points themselves, and it should be beneficial especially when grasping a planer or cylindrical object.

\section{Related Work}
\subsection{Motion Planning Methods}
There are three popular classes of motion planning methods: 1) optimization-based methods, 2)sampling -based methods, and 3) imitation-learning-based methods.
The work by~\citet{Khatib86,Quinlan93,Brock02} using potential fields is the seminal work on optimization-based motion planning.
More recent optimization-based methods include CHOMP~\citep{Zucker13}, STOMP~\citep{Kalakrishnan11}, TrajOpt~\citep{Schulman14}, and GPMP~\citep{Mukadam18}.
These methods explicitly optimize the trajectory with respect to the objective function.
Sampling-based methods include Probabilistic RoadMap~(PRM)~\citep{Kavraki96,Kavraki98},
Rapidly-exploring Random Trees~(RRT)~\citep{LaValle01,LaValle06},  and RRT*\citep{Karaman11}.
These sampling-based methods can find a feasible trajectory in very complicated environments.
The third class is imitation-learning-based methods.
Dynamic Movement Primitives (DMP)~\citep{Ijspeert02}, Probabilistic Movement Primitives (ProMP)~\citep{Paraschos13} and Kernelized Movement Primitives~(KMP)~\citep{Huang19} are categorized in this class.
These methods learn the trajectory model from demonstrated trajectories to adapt to a new situation.
However, it is said that these imitation-based methods are not efficient under the existence of obstacles and that they often require additional process for obstacle avoidance~\citep{Osa18b}.
These three classes of motion planning are not completely separated, and recent studies have proposed methods that combine the benefits of each category.
\citet{Dragan15} show that the relation between DMP and trajectory optimization based on CHOMP, and studies such as \cite{Koert16,Osa17,Rana17} proposed methods that combine the optimization-based and imitation-learning-based  approaches.
Likewise, \citet{Ye11} proposed a method that combined the sampling-based and imitation-learning-based approaches. 
Although we categorized STOMP as an optimization-based method, it leverages a sampling process and optimizes a trajectory in a stochastic manner.
Likewise, while RRT* and PRM* proposed by \cite{Karaman11} are based on sampling-based methods, i. e., RRT and PRM, they are related to the optimization-based method in the sense that they find optimal solutions with respect to a given cost function.

As discussed above, optimization-based methods, sampling -based methods, and imitation-learning-based methods are related to each other.
Amongst these categories of motion planning methods, we focus on the optimization-based approach, and our method is closely related to CHOMP and STOMP.
CHOMP employs gradient-based optimization, while STOMP employs stochastic gradient-free optimization.
In our proposed trajectory optimization, we iterate the gradient-based and gradient-free optimization steps. 
Our gradient-free optimization step enables us to find multiple modes of the cost function via density estimation. 
However, the gradient-free step does not directly minimize the cost function, and the constraints such as the joint limits are not explicitly integrated. 
Our gradient-based optimization step directly minimizes the cost function and projects solutions onto the constraint solution space.

Regarding the problem of optimizing the trajectory including the end points, \citet{Dragan11} proposed a CHOMP-based method that can plan a feasible trajectory even if the goal configuration given by a user was actually infeasible. 
However, the work by~\citet{Dragan11} does not address the multimodality of the cost function.

The main contribution of our study to draw attention to the mutimodality of the objective function used in motion planning.
Since the optimization of the objective function is integrated in various ways in all of the motion planning method categories, we think that our study contributes to not only the optimization-based approaches but also to other approaches of motion planning.

\subsection{Multimodal Optimization}
In the field of reinforcement learning~(RL), there exist several previous studies discussing the problem of learning diverse skills~\citep{Zhang19,Eysenbach19}.
These studies propose methods of learning multiple ways of solving a given task; these implicitly correspond to different modes of the objective function.
Likewise, the concept of learning options in hierarchical reinforcement learning~\citep{Daniel16,Bacon17,Henderson18} can be interpreted as a way of learning option policies that correspond to the different modes of the Q-function.
Although the concept of learning diverse behaviors is found in both our work and the RL studies cited above, the targeted problem setting is different.
While the aim of these RL methods is to learn a control policy that determines control inputs such as torque inputs to a system's joints, the aim of our study is to plan a trajectory, which is a sequence of the desired states. 
Therefore, the method presented in these RL studies are not directly applicable to our problem setting.

The previous studies conducted by \citet{Goldberg87,Deb10,Stoean10,Agrawal14} have proposed to address the problems of multimodal optimization using evolutionary methods.
Although prior studies have reported that these methods can find multiple modes of the objective function, the dimensions of the parameters are often limited up to approximately 100.
The work by \citet{Agrawal14} shows that diverse behaviors can be obtained by maximizing the objective function that encodes the diversity of solutions.
These approaches based on black-box optimization methods can be applied to a wide range of problems. However, it is not a trivial task to directly apply these evolutionary methods to trajectory optimization in robotics. 
In trajectory optimization, there is a need to optimize hundreds of trajectory parameters.
For example, when the trajectory of a robot with 7 DoFs is represented by 50 time steps, then it is represented as a vector with 350 dimensions.
To cope with such high dimensional parameters, a combination of gradient-based and gradient-free procedures is utilized in the proposed method. Further, a structured exploration strategy is employed to achieve efficient sampling for motion planning.

In addition, the evolutionary algorithms are usually built using heuristic optimization methods, the relation of the algorithm and minimization of the objective function is not clear.
In this work, we illustrate our method of reducing the trajectory optimization problem into a density estimation problem, 
while also demonstrating how our algorithm is related to the minimization of the objective function.

\subsection{Challenges in Multimodal Optimization}

\begin{figure}
	\centering
	\begin{subfigure}[t]{0.465\columnwidth}
		\centering
		\includegraphics[width=\textwidth]{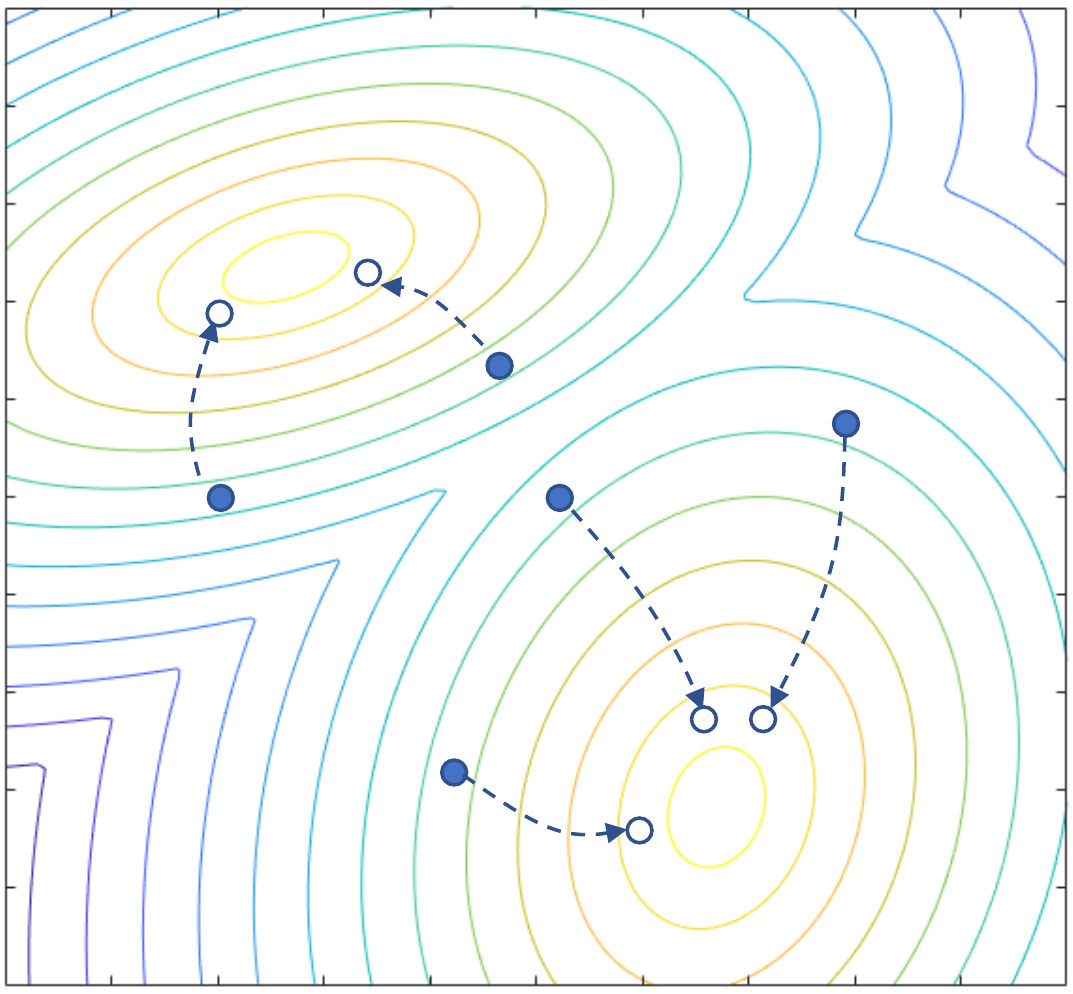}
		\caption{Although multiple solutions can be obtained by running an optimizer with different random seeds, the amount of modes that actually exist cannot be known.}
	\end{subfigure}
	\begin{subfigure}[t]{0.465\columnwidth}
		\centering
		\includegraphics[width=\textwidth]{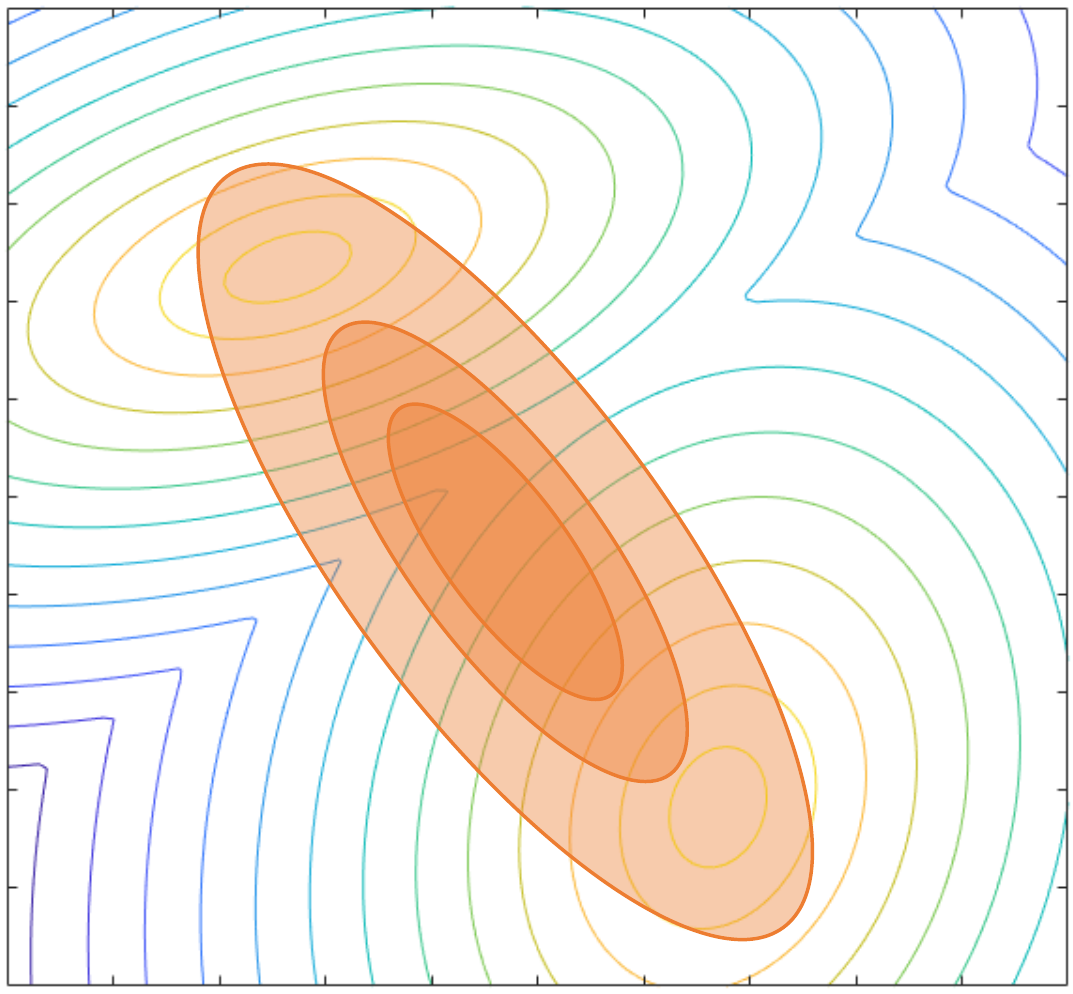}
		\caption{If a unimodal distribution is fitted for sampling, the optimization procedure may fall into a local minimum. }
	\end{subfigure}
	\caption{Schematic of challenges of multimodal optimization. The contour plot indicates a bi-modal objective function.}
	\label{fig:challenge}
\end{figure}

	There exist several challenges in multimodal optimization, as shown in Figure~\ref{fig:challenge}.
	Multiple solutions can be obtained by running an existing optimization method from different initial seeds, as depicted in Figure~\ref{fig:challenge}(a).
	For example, if CHOMP~\citep{Zucker13} or TrajOpt~\citep{Schulman14} is run ten times with different random seeds, then ten numerically different solutions can be obtained even if the objective function unimodal.
	However, the difference of the solutions may be attributed to numerical computation, and 
	the amount of  distinctively different modes that are contained in the obtained solutions is not known.
	It is possible to cluster the solutions after running optimization several times; however, it is quite time-consuming in practice.
	Although the proposed method estimates the modes of the objective function by density estimation, which can be viewed as clustering, it does not require optimization for all the seeds/samples.
	Therefore, the proposed method is much more computationally efficient than the na\"ive method that runs an optimizer numerous times with different random seeds.

	In addition, gradient-free methods, such as the cross entropy method~\citep{Mannor03,Boer05}, reward-weighted regression~\citep{Peters07}, PI\textsuperscript{2}~\citep{Kappen07,Theodorou10}, often employ a unimodal distribution for sampling in each iteration. 
	However, as indicated by \citet{Daniel16}, such methods may fall into a local minimum when applied to a multimodal objective function, as shown in Figure~\ref{fig:challenge}(b).
	To manage this issue, the gradient-free update in the proposed method systematically fits the multimodal distribution for sampling.

\section{Multimodal Trajectory Optimization for Motion Planning}
\subsection{Problem Setting}
The goal of motion planning for manipulation tasks is to plan a trajectory between the start configuration $\vect{q}_0 \in \Real^{D}$ and the goal configuration $\vect{q}_T \in \Real^{D}$
where $D$ is the number of joints in a robotic manipulator and $T$ is the number of time steps in a trajectory.
In this paper, $\vect{\xi} = [\vect{q}_0, \ldots,\vect{q}_T  ] \in \Real^{D \times T} $ denotes a trajectory in configuration space.
We also denote by $\vect{x}_{\textrm{end}}(\vect{q})$ the position of the end effector in task space, given a configuration $\vect{q}$.
In our formulation, a cost function $\mathcal{C}(\vect{\xi})$ quantifies the quality of a trajectory $\vect{\xi}$.
Like other optimization-based methods~\citep{Zucker13,Kalakrishnan11,Schulman14}, we consider an iterative trajectory optimization process.
We denote by $\vect{\xi}^k = [\vect{q}^k_0, \ldots,\vect{q}^k_T  ] $ a trajectory estimated at the $k$th iteration.
The initial trajectory is represented by $\vect{\xi}^0$, and the start and goal configurations given by a user are denoted by $\vect{q}^0_0$ and $\vect{q}^0_T$, respectively.
We use the initial trajectory obtained by interpolating between $\vect{q}^0_0$ and $\vect{q}^0_T$ on the configuration space, unless otherwise stated.

While existing motion planning methods often assume that $\vect{q}_0$ and $\vect{q}_T$ are given and fixed,
this may not be the case in practice.
As discussed in the introduction, the optimal goal configurations are not be obvious on tasks such as grasping a cylindrical or planer object.
Therefore, we also consider a case where rotation of the end effector is allowed around  an axis $\vect{r}_g=[r^g_x, r^g_y, r^g_z]$ at the goal point.
We present a method that optimizes a trajectory including the goal configuration $\vect{q}_T$, while maintaining the end effector positions on the task space at the goal point, $\vect{x}_{\textrm{end}}(\vect{q}_T)$.

\subsection{Overview of the Proposed Multimodal Trajectory Optimization Method}
The goal of our trajectory optimization is to find trajectories that correspond to the modes of the cost function $\mathcal{C}(\vect{\xi})$.
However, it is challenging to estimate the location of the modes of the cost function.
To make this problem tractable, we divide this trajectory optimization problem into two steps:
1) finding the location of modes of the cost function approximately and 2) locally optimizing each trajectory that corresponds to a mode of the cost function.

The proposed stochastic multimodal trajectory optimization (SMTO) algorithm is summarized in Algorithm~\ref{alg:MulTrajOpt}.
To find trajectories corresponding to the modes of the cost function in the first step, we reduce the problem of finding the modes of the cost function to that of estimating the density of the trajectory induced by a distribution based on the cost function, which we refer to as the \textit{cost-weighted density estimation}.
We obtain multiple trajectories by approximating a multimodal distribution using an importance sampling approach (Lines 3-12 in Algorithm~\ref{alg:MulTrajOpt}).
In our implementation, the density estimation is performed using variational Bayes expectation maximization (VBEM).
In the second step, we locally optimize each trajectory obtained in the first step using a gradient-based method.
While the cost-weighted density estimation in the first step does not directly minimize the cost function,
the second step of our trajectory update directly minimizes the cost function and projects the trajectories onto the space that satisfies the constraints (Lines 13-15 in Algorithm~\ref{alg:MulTrajOpt}).
The gradient-based update in the second step is adapted from the covariant gradient descent proposed by~\cite{Zucker13}.
As the density estimation in the first step of our method is gradient-free, one can interpret that our method alternates between gradient-free and gradient-based trajectory updates.

\begin{algorithm}[t]
	\caption{The Stochastic Multimodal Trajectory Optimization Algorithm (SMTO)}
	\begin{algorithmic}[1]
		\STATE{
			\textbf{Input:} Initial trajectory $\vect{\xi}^0$, Maximum number of solutions $O$ 
		}
		\FOR{$k = 1,\ldots, K$}
		\FOR{$n = 1,\ldots, N$}
		\IF{the goal configuration is fixed}
		\STATE{ Sample trajectories by following \eqref{eq:exploration_traj_ini}, \eqref{eq:exploration_traj_k}}
		\ELSIF{the goal configuration has a rotational freedom}
		\STATE{Sample trajectories by following \eqref{eq:exploration_end}}
		\ENDIF
		\STATE{ Compute the cost of sampled trajectories}
		\ENDFOR \\
		\STATE{ Perform dimensionality reduction of the sampled trajectories, e.g. Laplacian eigenmaps  }		\STATE{ Perform the density estimation using VBEM with importance sampling in~\eqref{eq:importance}}
		\STATE{ Update the trajectories with gradient-descent as in \eqref{eq:chomp_update}}
		\STATE{ Project the trajectories onto the constraint solution space as in \eqref{eq:projection}}
		\STATE{ (Optional) Update the trajectories for optimizing the null space with \eqref{eq:update_null} }
		\ENDFOR \\
		\STATE{ \textbf{Return:} trajectories that correspond to modes of the cost function }
	\end{algorithmic}
	\label{alg:MulTrajOpt}
\end{algorithm}

We explain the details of our algorithm in the following sections.
The formulation and algorithm of the cost-weighted density estimation in the first step is described in Section~\ref{sec:MulTrajOpt}.
Our trajectory sampling strategy used in the cost-weighted density estimation is explained in Section~\ref{sec:exploration}.
Gradient-based optimization of trajectories including their goal configurations is described in Section~\ref{sec:CHOMPend}.

\subsection{Multimodal trajectory Optimization via Density Estimation with Importance Sampling}
\label{sec:MulTrajOpt}
In this section, we describe the manner in which we estimate the location of modes of the objective function.
For this purpose, we reduce the problem of finding the modes of the objective function to that of approximating the density of the trajectory with a multimodal distribution.

We consider a distribution over trajectories in the following form:
\begin{align}
d^{\cost}(\vect{\xi}) = \frac{f( \mathcal{C} (\vect{\xi}) )}{Z},
\label{eq:scale_f}
\end{align}
where $f(\cdot)$ is a monotonically decreasing function with respect to the input variable and always satisfies $f(\cdot) > 0$, and $Z$ is the partition function. 
When following such a distribution, a trajectory with the smaller cost is drawn with a higher probability.
Therefore, finding the optimum of the cost function is equivalent to finding the modes of the distribution $d^{\cost}$.
Thus, we can formulate the problem of finding the modes of the cost function $\mathcal{C}(\vect{\xi})$ as that of estimating the density of trajectories induced by $d^{\cost}$.

However, trajectories drawn from $d^{\cost}(\vect{\xi})$ are not available in practice.
To solve this density estimation problem, we employ the importance sampling approach.
Namely, we sample trajectories $\{ \vect{\xi}_i \}^{N}_{i=1}$ using a proposal distribution $\beta(\vect{\xi})$, evaluate their costs $\mathcal{C}(\vect{\xi}_i)$ for $i=1,\ldots,N$ and estimate $d^{\cost}(\vect{\xi})$ using those samples with the importance weights. 
The importance weight is given by
\begin{align}
W(\vect{\xi}) = \frac{d^{\cost}(\vect{\xi})}{\beta(\vect{\xi})} = \frac{f( \mathcal{C} (\vect{\xi}) )}{Z \beta(\vect{\xi})},
\end{align}
where $\beta(\vect{\xi})$ is an arbitrary proposal distribution for sampling trajectories.
In practice, we normalize the importance weights as
\begin{align}
\tilde{W}(\vect{\xi}_i)
& = \frac{W}{\sum^{N}_{j=1}W(\vect{\xi}_j) } 
= \frac{\frac{f( \mathcal{C} (\vect{\xi_i}) )}{Z \beta(\vect{\xi}_i)}}{\sum^{N}_{j=1}\frac{f( \mathcal{C} (\vect{\xi}_j) )}{Z \beta(\vect{\xi}_j)} } \\
& = \frac{\frac{f( \mathcal{C} (\vect{\xi_i}) )}{\beta(\vect{\xi}_i)}}{\sum^{N}_{j=1}\frac{f(\mathcal{C} (\vect{\xi}_j) )}{\beta(\vect{\xi}_j)} }.
\label{eq:importance}
\end{align}
Since the partition function $Z$ is canceled, we do not have to compute $Z$ in practice. 
We use this importance weight for approximating the multimodal distribution $d^{\cost}(\vect{\xi})$.
In our implementation, we used $f(\vect{x}) = \exp(- \vect{x})$, although our algorithm is not limited to a specific form of $f$.

This problem of matching between the trajectory distribution $d^{\cost}(\vect{\xi})$ and a distribution $d_{\vect{\theta}}(\vect{\xi})$ parameterized by a vector $\vect{\theta}$ is formulated as minimizing the KL divergence,
which is given by
\begin{align}
\vect{\theta}^* = \arg \min_{\vect{\theta}} \KL \left( d^{\cost}(\vect{\xi}) || d_{\vect{\theta}}(\vect{\xi}) \right),
\label{eq:KL}
\end{align}
where $ \KL \left( d^{\cost}(\vect{\xi}) || d_{\vect{\theta}}(\vect{\xi}) \right)$ is the KL divergence, and is given by
\begin{align}
\KL \left( d^{\cost}(\vect{\xi}) || d_{\vect{\theta}}(\vect{\xi}) \right)
& =  \int d^{\cost}(\vect{\xi})\log \frac{d^{\cost}(\vect{\xi})}{ d_{\vect{\theta}}(\vect{\xi})} \textrm{d}\vect{\xi} \nonumber \\
& = \int W(\vect{\xi})\beta(\vect{\xi})\log \frac{ W(\vect{\xi}) \beta(\vect{\xi})}{ d_{\vect{\theta}}(\vect{\xi})} \textrm{d}\vect{\xi}.
\label{eq:kl_p}
\end{align}
The minimizer of $\KL \left( d^{\cost}(\vect{\xi}) || d_{\vect{\theta}}(\vect{\xi}) \right)$ is given by the maximizer of the weighted log likelihood:
\begin{align}
L( d_{\vect{\theta}}, \beta ) & =  \int  W(\vect{\xi}) \beta(\vect{\xi}) \log d_{\vect{\theta}}(\vect{\xi} ) \textrm{d}\vect{\xi} \nonumber \\
& \approx \frac{1}{N} \sum_{ \vect{\xi}_{i} \in \mathcal{D} } \tilde{W}(\vect{\xi}_{i}) \log d_{\vect{\theta}}(\vect{\xi}_{i} ).
\label{eq:log_likelihood}
\end{align}
Therefore, we can solve this density estimation problem by using the importance weight $W(\vect{\xi})$. 
In this study, we call $W(\vect{\xi})$ the \textit{cost-weighted importance}, 
and we refer to the problem formulated in \eqref{eq:KL} as \textit{cost-weighted density estimation}.

To discuss the connection between minimizing the cost function and minimizing the KL divergence in \eqref{eq:KL},
we consider the surrogate objective function $J(\vect{\theta}) = \E_{\vect{\xi} \sim d_{\vect{\theta}}}[- \tilde{R}(\vect{\xi})]$ where  $\tilde{R}(\vect{\xi}) = f( \mathcal{C}(\vect{\xi}))$.
Since $f$ is monotonically decreasing function, the maximizer of $\tilde{R}(\vect{\xi})$ is equivalent to the minimizer of $\mathcal{C}(\vect{\xi})$.
Therefore, we approximated the problem of minimizing $\mathcal{C}(\vect{\xi})$ with the problem of minimizing $J(\vect{\theta})$. 

Since $\tilde{R}(\vect{\xi}) > 0$, in a manner similar to the results shown by \cite{Dayan97,Kober11}, we can obtain 
\begin{align}
\log \big( - J( \vect{\theta}  ) \big) 
& = \log  \int d_{\vect{\theta}}( \vect{\xi} ) \tilde{R}( \vect{\xi} ) \textrm{d}\vect{\xi}
\nonumber \\
&= \log  \int \beta( \vect{\xi} ) \frac{ d_{\vect{\theta}}(\vect{\xi}) \tilde{R}( \vect{\xi})}{ \beta( \vect{\xi} ) } 
\textrm{d}\vect{\xi}\nonumber \\
&\geq   \int \beta( \vect{\xi}) \frac{ \tilde{R}( \vect{\xi})}{ \beta(\vect{\xi} )  } 
\log d_{\vect{\theta}}( \vect{\xi} ) \textrm{d}\vect{\xi} \nonumber \\
& =   Z  \int \beta( \vect{\xi} ) \frac{ \tilde{R}( \vect{\xi})}{ \beta( \vect{\xi} ) Z} 
\log  d_{\vect{\theta}}(\vect{\xi}) \textrm{d}\vect{\xi}  \nonumber \\
& \propto   \int  W(\vect{\xi}) \beta(\vect{\xi}) \log d_{\vect{\theta}}(\vect{\xi} ) \textrm{d}\vect{\xi} 
= L( d_{\vect{\theta}}, \beta ).
\end{align}
From line 2 to 3, we used Jensen's inequality. In line 4, we used $W(\vect{\xi}) = \frac{ R( \vect{\xi})}{ \beta( \vect{\xi} ) Z}$.
Therefore, maximizing the weighted log likelihood $L( d_{\vect{\theta}}, \beta )$ is equivalent to minimizing the upper bound of $J(\vect{\theta})$.

\begin{figure*}
	\centering
	\begin{subfigure}[b]{0.65\columnwidth}
		\centering
		\includegraphics[width=\textwidth]{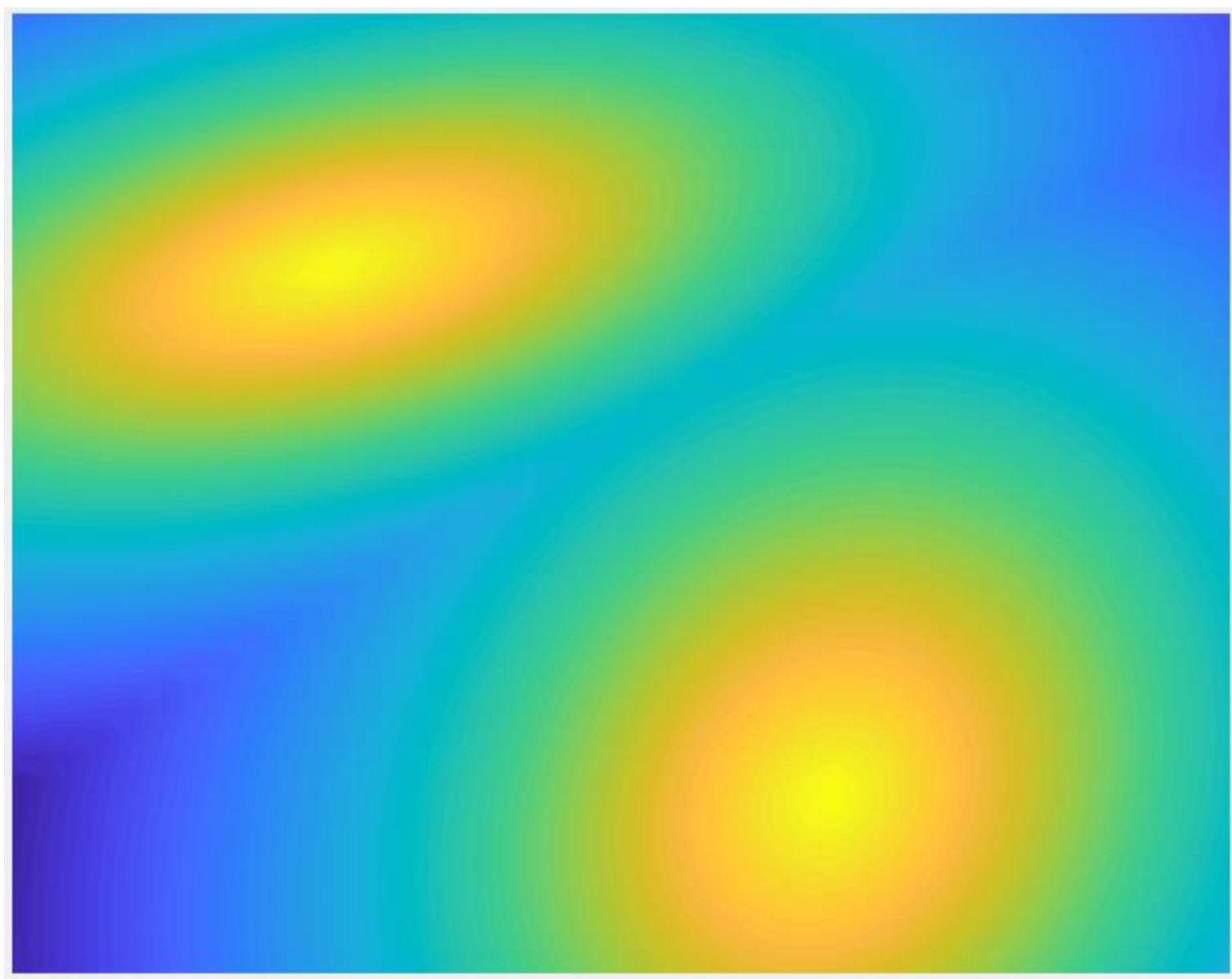}
		\caption{Visualization of the cost function. The warmer color represents the lower cost.}
	\end{subfigure}
	\begin{subfigure}[b]{0.65\columnwidth}
		\centering
		\includegraphics[width=\textwidth]{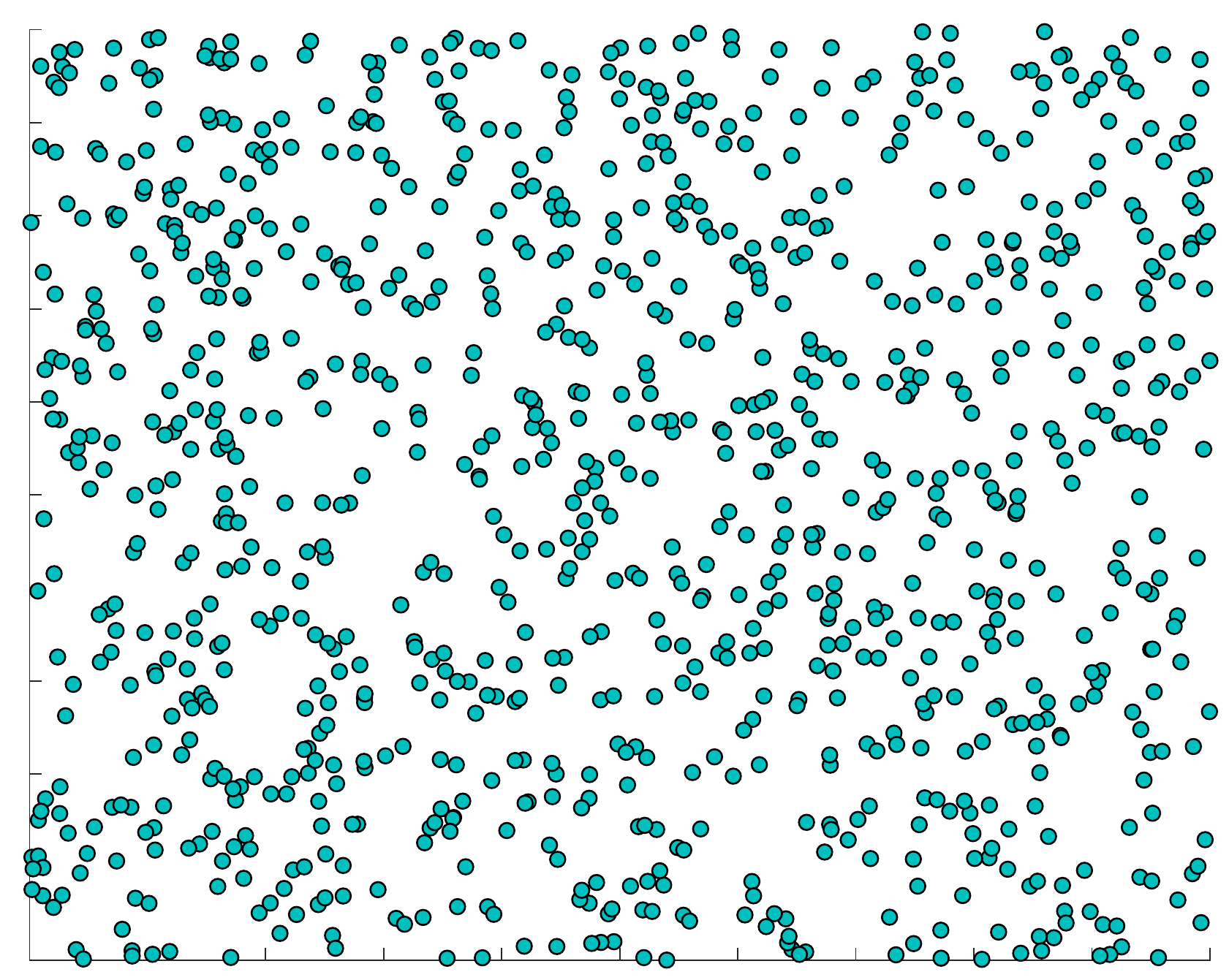}
		\caption{Samples drawn from the uniform distribution.}
	\end{subfigure}
	\begin{subfigure}[b]{0.65\columnwidth}
		\centering
		\includegraphics[width=\textwidth]{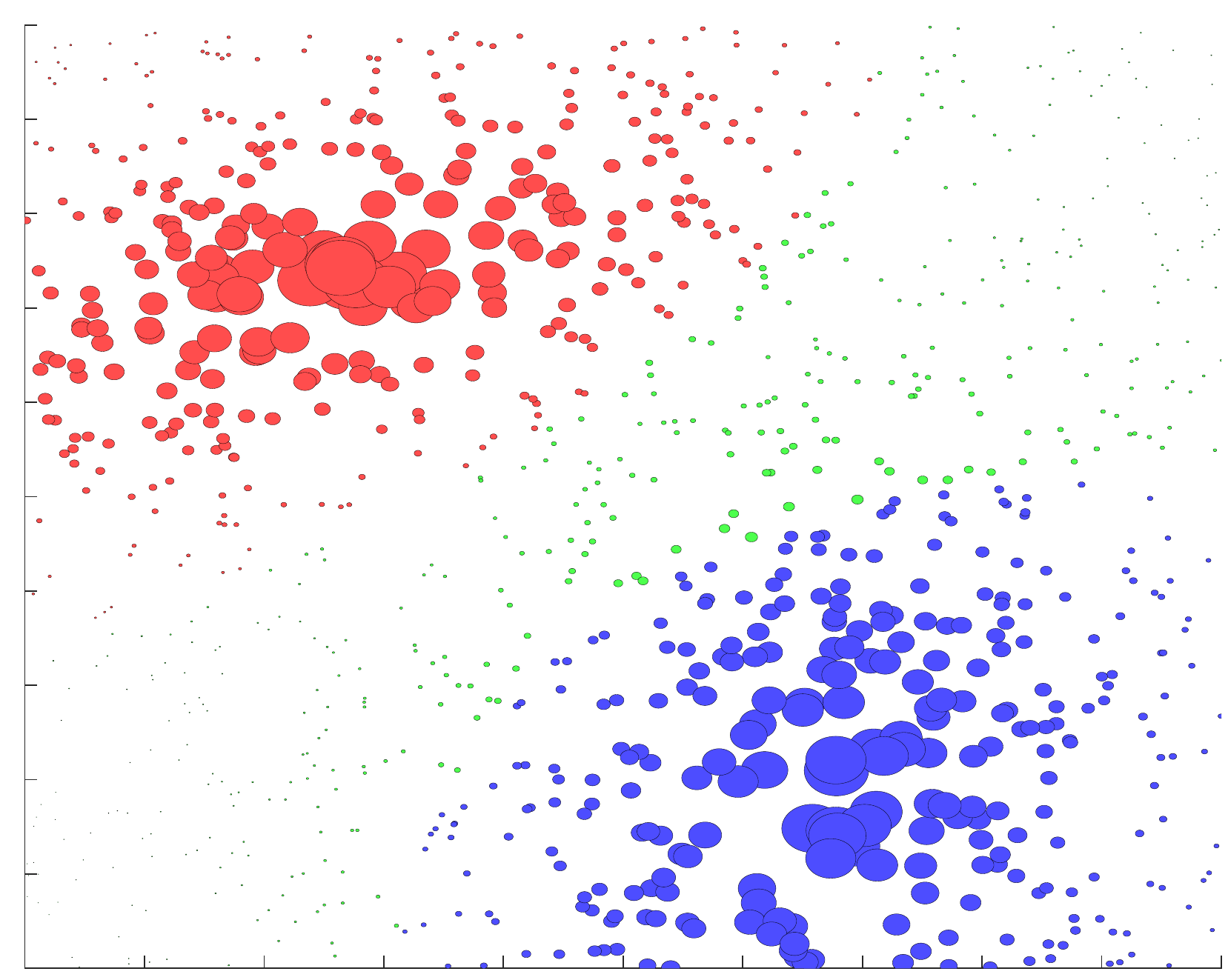}
		\caption{The same samples in (b) with the scaling based on the importance weight.}
	\end{subfigure}
	\caption{Example of estimating modes of the cost function with importance sampling. (a) shows the cost function in 2D space and let us consider the horizontal and vertical axes represent trajectory parameters. The warmer color represents the lower cost in (a). (b) shows samples drawn from uniform distributions. In (c), the same samples in (b) are shown, but samples with higher importance in Eq.~\eqref{eq:importance} is drawn as larger circles. The colors in (c) represent the clusters found by VBEM. One can see that the clusters that correspond to the modes of the cost function appear by using the importance weight.
	} 
	\label{fig:approach}
\end{figure*}

In our implementation, we employed Gaussian Mixture Models (GMMs) to represent a multimodal distribution.
\begin{align}
d_{\vect{\theta}}(\vect{\xi}) = \sum^{O}_{o=1} p(o) p(\vect{\xi} | o)
\end{align}
A popular way of log likelihood maximization for the Gaussian mixture model fitting is the expectation-maximization (EM) algorithm.
We use the variational Bayes expectation-maximization (VBEM) algorithm with importance sampling in this study, whilst the maximum likelihood~(ML) EM algorithm is also applicable~\citep{Bishop06}.
The advantage of the VBEM algorithm over the ML EM algorithm is that the use of the symmetric Dirichlet distribution as a prior of the mixing coefficient leads to a sparse solution.
This property of VBEM is suitable for our trajectory optimization since unnecessary clusters are automatically eliminated and the clusters obtained by VBEM focus on separate modes.
Therefore, the number of solutions is automatically estimated by using VBEM in our multimodal trajectory optimization framework.
We provide the details of VBEM with importance weights in Appendix~\ref{sec:VBEM}.

When fitting GMMs for approximating $d^{\cost}(\vect{\xi})$, the dimensionality of the raw data of a whole trajectory is too high for performing the VBEM algorithm.
For this reason, we reduced the dimensionality by performing Laplacian eigen map~\citep{Belkin03}.
Once a GMM is successfully learned, we can assign sampled trajectories to clusters found by VBEM as
\begin{align}
o(\vect{\xi}) = \arg \max_{o'} p(o' | \vect{\xi}).
\label{eq:cluster_o}
\end{align}
The trajectory corresponding to the $l$th mode of the cost function is given by
\begin{align}
\vect{\xi}_{l} = \frac{\sum^{N}_{i=1} \delta(o(\vect{\xi}_i)-l)W(\vect{\xi}_i)\vect{\xi}_i}{\sum^{N}_{i=1} \delta(o(\vect{\xi}_i)-l)W(\vect{\xi}_i)},
\label{eq:weighted_sum}
\end{align}
where $\delta(\cdot)$ is the delta function, if $o=l$ then $\delta(o-l)= 1$ else $\delta(o-l)= 0$.

Figure~\ref{fig:approach} shows the behavior of the cost-weighted density estimation.
In the example shown in Figure~\ref{fig:approach}, the cost function is bi-modal, and the proposal distribution is uniform.
The modes of the cost function can be clearly indicated by the samples scaled with the importance weight in \eqref{eq:importance}.
The number of clusters are determined by VBEM in this example.

\subsection{Strategy for Sampling Trajectories}
\label{sec:exploration}
To achieve efficient exploration, it is necessary to employ a structured sampling strategy, which is suitable for motion planning.
When the start and goal configuration is fixed, we can use the sampling strategies similar to STOMP~\citep{Kalakrishnan11}.
In the first iteration of the trajectory optimization, we use the following proposal distribution:
\begin{align}
\beta^1_{\textrm{traj}}(\vect{\xi}) = \mathcal{N}(\vect{\xi}^0, a \vect{B})
\label{eq:exploration_traj_ini}
\end{align}	
where $a$ is a constant, $\vect{\xi}^0$ is an initial trajectory, and the covariance matrix $\vect{B}$ is given by the inverse of  $\vect{A}^{\top}\vect{A}$, and the matrix $\vect{A}$ defined as
\begin{align}
\vect{A} & = 
\left[
\begin{array}{cccccc}
0 & 0  & 0 & \ldots & & 0 \\
0 &   2 & -1 &  & &\\ 
0  & -1 & 2 &  & &\\
0 & 0 & -1 & & 0 & 0\\
0 & 0 & &  & -1 & 0\\
&   & &  & 2  &- 1 \\
0 & 0 & &  & -1 & 2
\end{array}
\right].
\label{eq:metric}
\end{align}
\citet{Kalakrishnan11} proposed the use of the above covariance matrix $\vect{B}$ for sampling trajectories, and it is empirically shown that the use of this covariance matrix leads to smooth perturbation of the whole trajectory.
For sampling at the $k$th iteration of trajectory optimization, we use a proposal distribution given by a mixture model 
\begin{align}
\beta^{k}_{\textrm{traj}}(\vect{\xi}) = \sum^{L}_{l=1}U(l)\mathcal{N}(\vect{\xi}^{k-1}_{l}, a\vect{B}),
\label{eq:exploration_traj_k}
\end{align}
for $k > 1$ where $\vect{\xi}^k_{l}$ represents a trajectory that corresponds to the $l$th modes found by the $k$th iteration, $L$ is the number of the solutions found at the $k$th iteration, and $U(l)$ is a discrete uniform distribution.

\begin{figure}[]
	\centering
	\begin{subfigure}[b]{0.465\columnwidth}
		\centering
		\includegraphics[width=\textwidth]{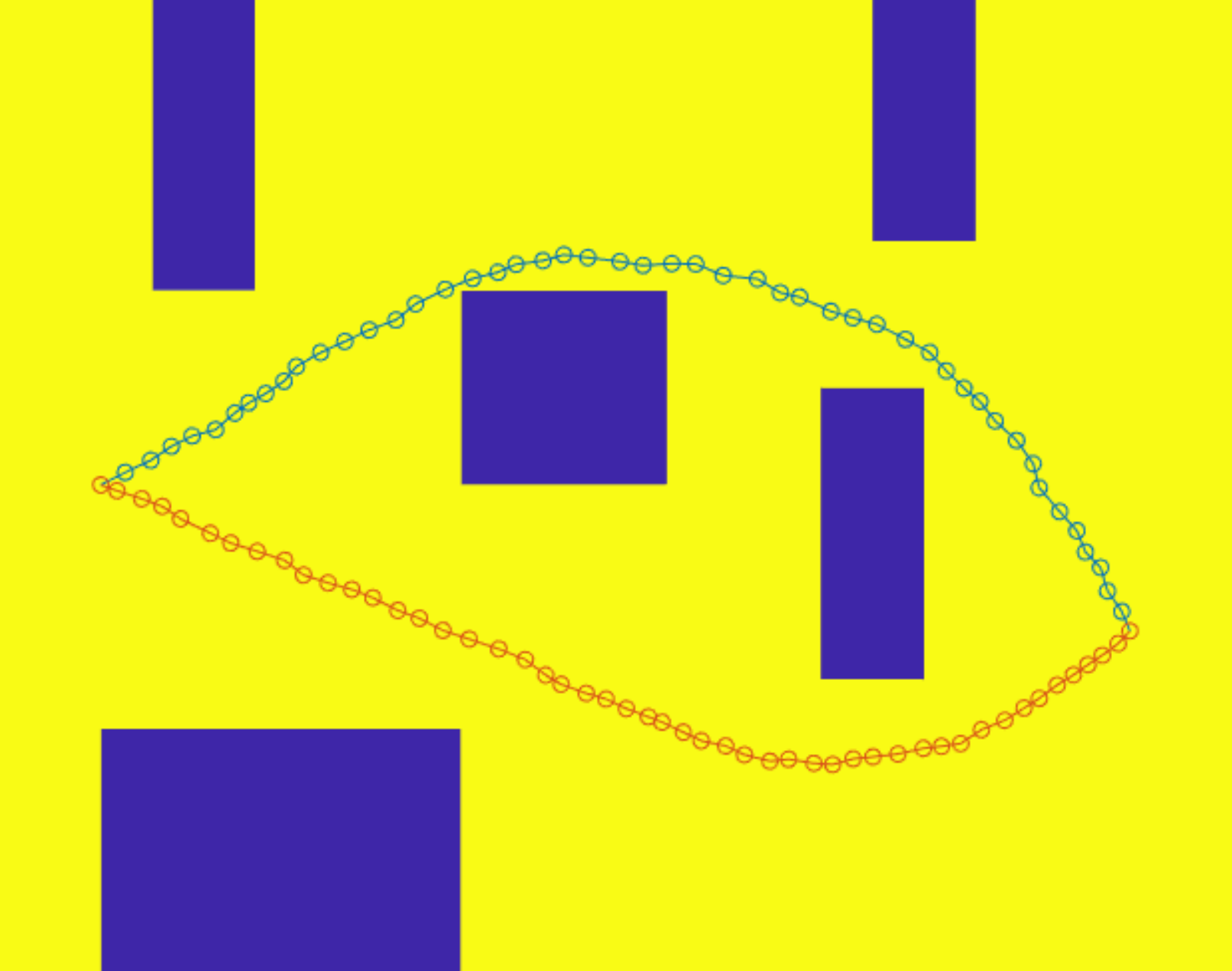}
		\caption{Path planning task $\#$ 1. }
	\end{subfigure}
	\hfill
	\begin{subfigure}[b]{0.465\columnwidth}
		\centering
		\includegraphics[width=\textwidth]{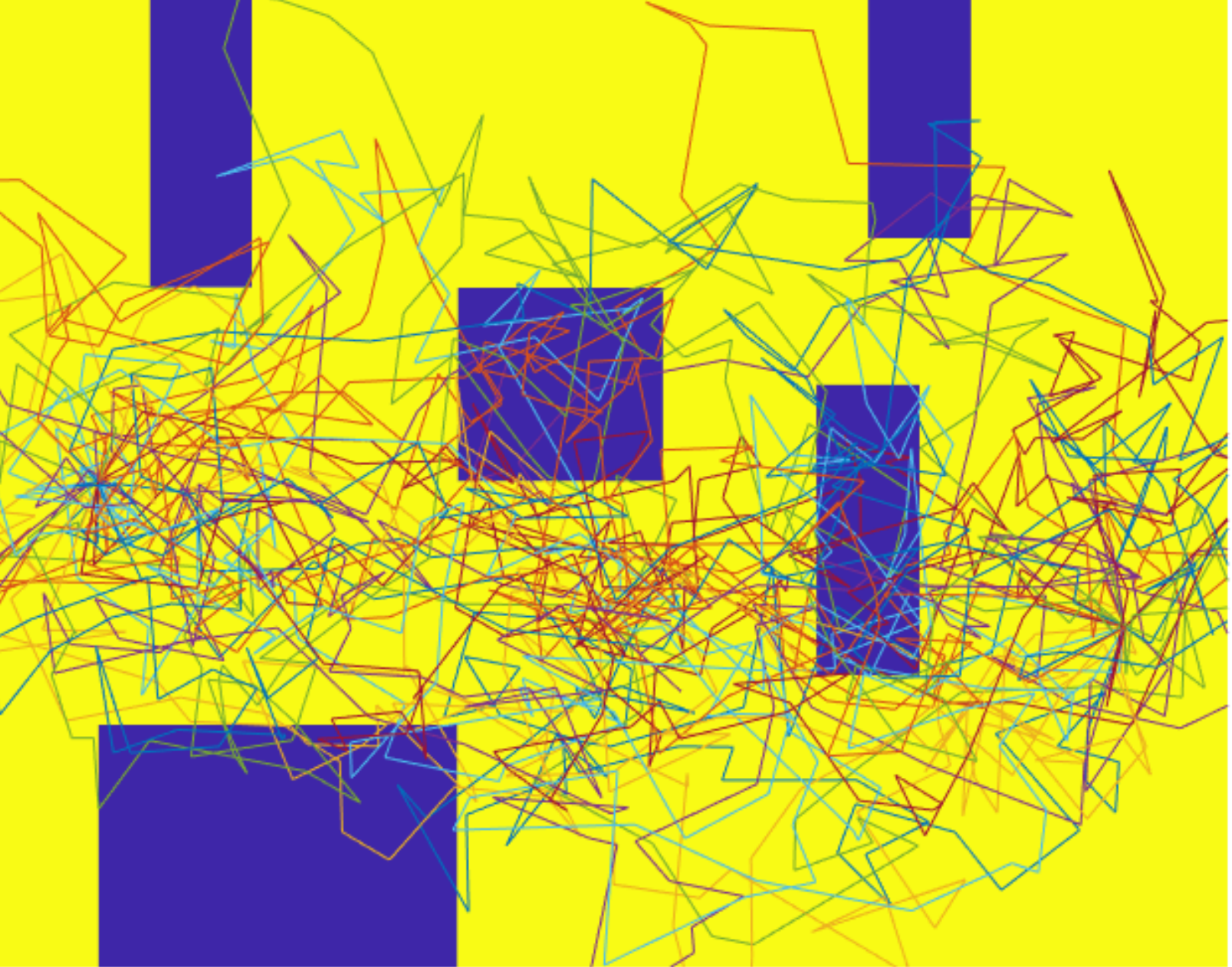}
		\caption{Explored paths in (a).}
	\end{subfigure}
	\hfill
	\begin{subfigure}[b]{0.465\columnwidth}
		\centering
		\includegraphics[width=\textwidth]{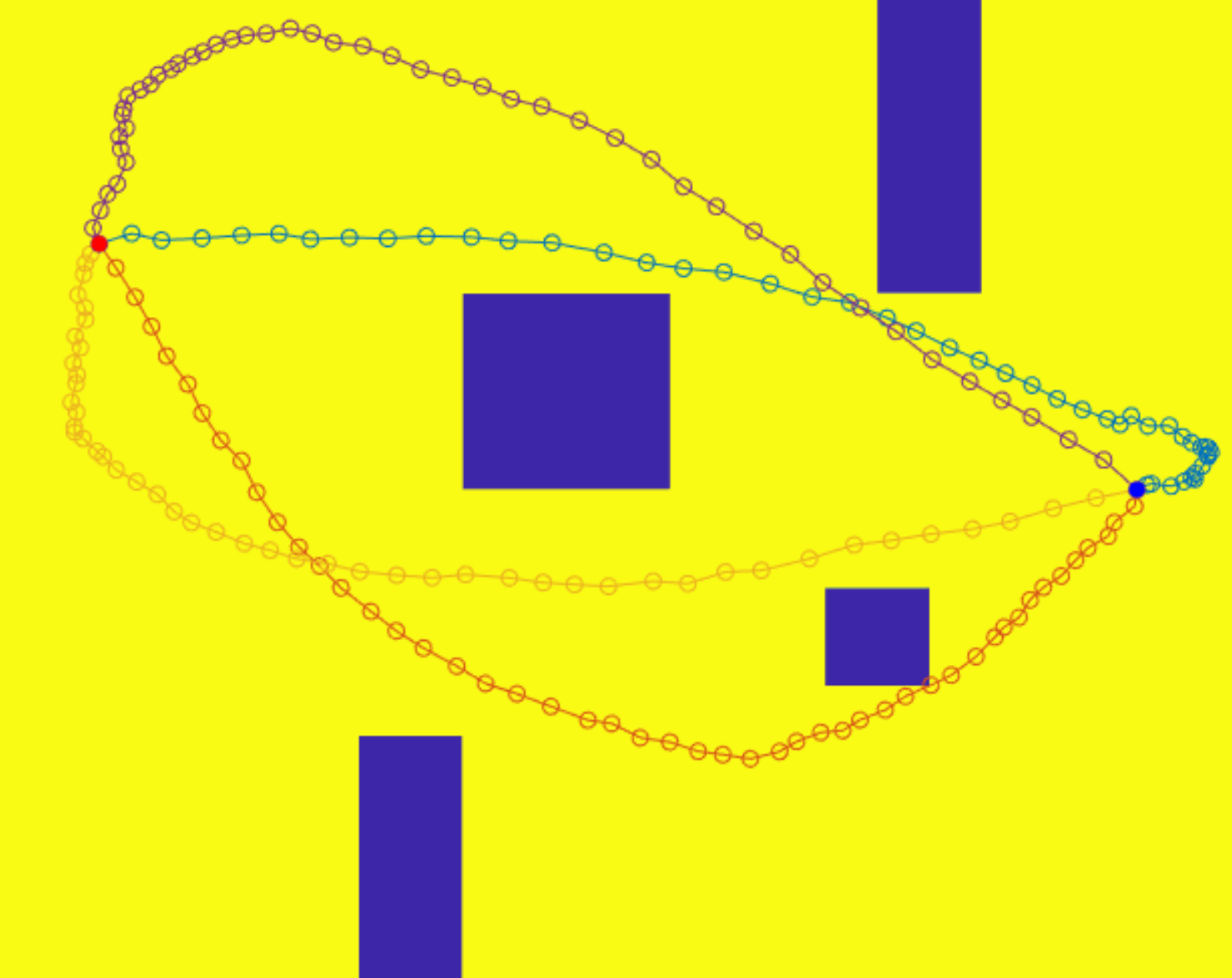}
		\caption{Path planning task $\#$ 2.}
	\end{subfigure}
	\hfill
	\begin{subfigure}[b]{0.465\columnwidth}
		\centering
		\includegraphics[width=\textwidth]{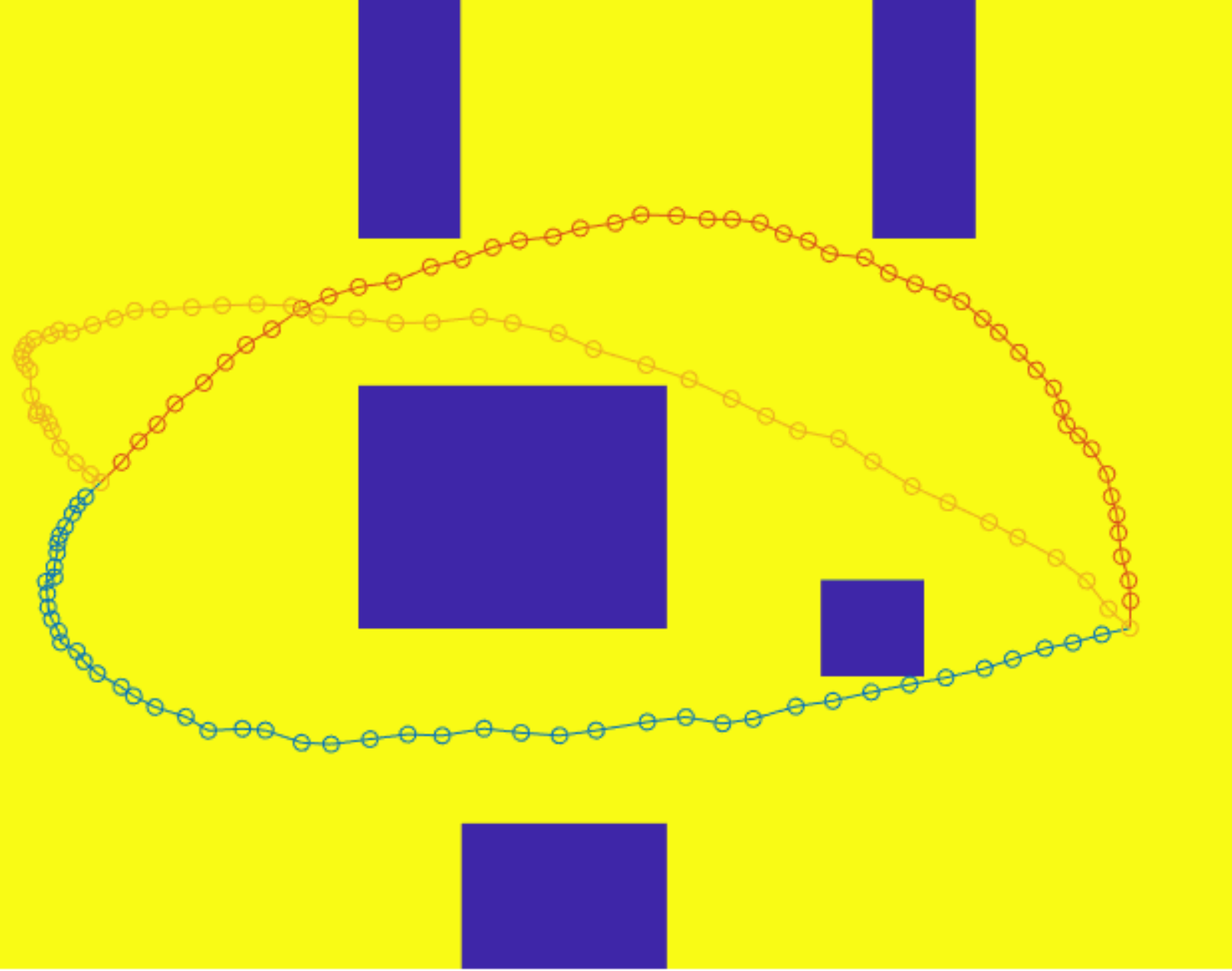}
		\caption{Path planning task $\#$ 3.}
	\end{subfigure}
	\caption{Path planning in 2D space. The yellow region indicates the zero-cost region, and the blue region indicates the non-zero-cost region. The cost function is flat except the boundary of the yellow and blue regions. 
	The length of the trajectory is not penalized in this task.
	} 
	\label{fig:path2d}
\end{figure}

To demonstrate the qualitative performance, we show path planning in 2D space using the cost-weighted density estimation and the exploration strategy described in \eqref{eq:cluster_o}--\eqref{eq:exploration_traj_k}.
The goal of this task is to plan a path that reaches the goal point from the starting point.
Examples of this path planning task are shown in Fig.~\ref{fig:path2d}.
The cost is 0 in the yellow region and 1 in the blue region.
Therefore, the cost function is designed to be flat except the boundary of the yellow and blue regions;
we can see whether the proposed cost-weighted density estimation works for discontinuous cost functions.
The trajectory length is not penalized by the cost function, and the number of time steps $T$ is fixed at $T=50$.
We iterate the sampling and the cost-weighted density estimation to solve this task.
The covariant gradient descent described in Section~\ref{sec:CHOMPend} is not used.

The trajectories shown in Fig.~\ref{fig:path2d} represent the solutions found by iterating the sampling in \eqref{eq:exploration_traj_ini}--\eqref{eq:exploration_traj_k} and the cost-weighted density estimation two times.
It is apparent from Fig.~\ref{fig:path2d} that the proposed method can find multiple solutions despite the presence of discontinuity of given cost functions.

\subsection{Covariant Gradient Descent under Constraints}
\label{sec:CHOMPend}
While the trajectory update based on the cost-weighted density estimation can minimize the upper bound of the cost function, the trajectories corresponding to the mean of each cluster do not correspond to the modes of the cost function. 
In addition, the trajectories obtained from the cost-weighted density estimation do not explicitly satisfy the constraints such as the end effector position or the joint limits.
Since we maintain the end effector position in the task space at the end points, the resulting trajectory should satisfy $\vect{x}_{\textrm{end}}(\vect{q}^0_T) = \vect{x}_{\textrm{end}}(\vect{q}^1_T) = \vect{x}_{\textrm{end}}(\vect{q}^k_T)$.
However, when we explore different goal configurations, the solutions obtained by \eqref{eq:weighted_sum} may deviate from a given end-effector position in the task space.

The prior work by~\cite{Dragan11} proposed a trajectory optimization method that alternates the unconstrained update of the trajectory and projection of the solution onto the constraint subspace.
Likewise, we need to project the solutions obtained in \eqref{eq:weighted_sum} onto the constraint subspace.
To obtain trajectories that correspond to the modes of the cost function and satisfy desired constraints,
we perform a gradient-based trajectory update.
Specifically, we employ the covariant trajectory update based on CHOMP:
\begin{align}
\vect{\xi}^{\new} = \arg \min_{\vect{\xi}} \left\{  \mathcal{C}(\vect{\xi}^c) + \vect{g}^{\top} (\vect{\xi} - \vect{\xi}^c) + \frac{\eta}{2} \left\| \vect{\xi} - \vect{\xi}^c \right\|^2_{\vect{M}}  \right\}
\label{eq:chomp}
\end{align}
where $\vect{g} = \nabla \mathcal{C}(\vect{\xi})$, $\vect{\xi}^c$ is the current plan of the trajectory, 
$\eta$ is a coefficient, 
and  $\left\| \vect{\xi} \right\|^2_{\vect{M}}$ is the norm defined by a matrix $\vect{M}$ as $\left\| \vect{\xi} \right\|^2_{\vect{M}} = \vect{\xi}^{\top}\vect{M}\vect{\xi}$.
This covariant trajectory update can be interpreted as a trust region optimization method where
the trust region is defined by $\vect{M}$.
In our framework, we locally update the estimation of trajectories that correspond to the modes of the cost function.
Therefore, each trajectory update needs to be stable and solutions should not jump to other modes.
The trust region optimization with the norm defined by $\vect{M}$ enables the stable and local trajectory update in our framework.
In our implementation, we used the metric given by $\vect{M} = \vect{K}^{\top}\vect{K}$ and $\vect{K}$ is a finite differencing matrix as in \citep{Zucker13,Dragan11}.

The trajectory update in \eqref{eq:chomp} is equivalent to the update rule
\begin{align}
\vect{\xi}^{\new} = \vect{\xi}^{c} - \frac{1}{\eta} \vect{M}^{-1} \vect{g}.
\label{eq:chomp_update}
\end{align}
When the position of the end effector at the goal position in the current plan of the trajectory is changed from the given one, we can shift the goal configuration and update the whole trajectory using the following equation as discussed in~\cite{Dragan15}:
\begin{align}
\vect{\xi}^{\new} =  \vect{\xi}^{c} + \vect{M}^{-1}[0,\ldots,0, \Delta \tilde{\vect{q}}_T]^\top,
\label{eq:projection}
\end{align}
where 
$\Delta \tilde{\vect{q}}_T$ is given by
\begin{align}
\Delta \tilde{\vect{q}}_T = \vect{J}^{-1} 
\left[
\begin{array}{c}
\vect{x}_{\textrm{end}}(\vect{q}^0_T) - \vect{x}_{\textrm{end}}(\vect{q}^c_T) \\
\vect{0}
\end{array}
\right],
\end{align}
where $\vect{J}$ is the Jacobian matrix.
We used the above update rule in order to project the solution obtained by the cost-weighted density estimation onto the constraint solution space.

\subsection{Extension for Optimizing the Rotational DoF at the Goal Point}
\label{sec:goal_opt}

When we optimize the orientation of the goal configuration, we need to explore trajectories that have different goal orientations for the cost-weighted density estimation.
For this purpose, we sample different goal configurations and propagate the difference of the goal configuration to the whole trajectory.

\begin{figure}[t]
	\centering
	\includegraphics[width=0.7\columnwidth]{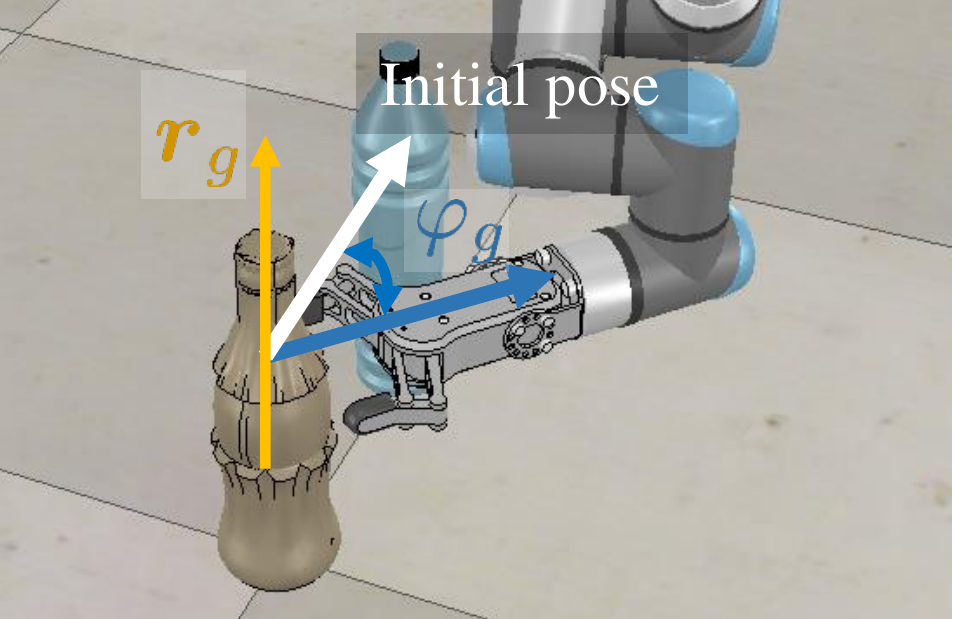}
	\caption{We consider a rotation angle $\varphi_g$ of the end-effector around $\vect{r}_g$ at the goal point.
		We generate trajectories that have different orientations at the goal point, maintaining the end effector position in task space.
	} 
	\label{fig:rotation_notation}
\end{figure}

Herein, we consider a case where we can change the orientation of the end-effector around the rotation axis $\vect{r}_g$ as shown in Fig.~\ref{fig:rotation_notation}.
We sample a rotation angle $\varphi_g$ round $\vect{r}_g$ from a uniform distribution as
\begin{align}
\varphi_g \sim U(\varphi^{\min}_g, \varphi^{\max}_g).
\label{eq:noise_ang}
\end{align}
Subsequently, we compute the configuration $\vect{q}^{\textrm{rot}}_T(\varphi_g, \vect{r}_g)$ which has the orientation rotated $\varphi_g$ round $\vect{r}_g$ from a given goal orientation and satisfies $\vect{x}_{\textrm{end}}(\vect{q}^{\textrm{rot}}_T(\varphi_g, \vect{r}_g))=\vect{x}_{\textrm{end}}(\vect{q}^0_T)$.  

When a robotic manipulator is redundant, it is also necessary to explore the null space of the goal configuration.
For this purpose, we can also employ exploration by sampling a deviation of the configuration
\begin{align}
\Delta \vect{q}_{\textrm{null}} = \sum^{L}_{i=1} \epsilon_i  \vect{e}^{\nul}_i
\label{eq:exp_end_null}
\end{align}
where $\vect{e}^{\nul}_i$ represents a unit basis vector of the null space of the Jacobian $J(\vect{q}_T)$, and $\epsilon_i$ is a constant drawn from a uniform distribution as $\epsilon \sim U( -\epsilon^{\min}, \epsilon^{\max})$.

The noise injected to the goal configuration can be smoothly propagated to the whole trajectory using the following:   
\begin{align}
\Delta \vect{\xi}_{\textrm{end}} =   \vect{M}^{-1}[ 0, \ldots, 0, \Delta \vect{q}_{T}]^{\top}
\label{eq:exp_prop}
\end{align}
where $\Delta \vect{q}_T = \vect{q}^{\textrm{rot}}_T(\varphi_g, \vect{r}_g) - \vect{q}^k_T$,  $\vect{q}^k_T$ is the goal configuration estimated at the $k$th iteration, and $M$ is given by \eqref{eq:metric}.
In the case that a manipulator has redundant DoFs, the noise to the null space configuration can also be added as
$\Delta \vect{q}_T = \vect{q}^{\textrm{rot}}_T(\varphi_g, \vect{r}_g) - \vect{q}^k_T + \Delta \vect{q}_{\textrm{null}}$.

\begin{figure}
	\centering
	\begin{subfigure}[b]{0.465\columnwidth}
		\centering
		\includegraphics[width=\textwidth]{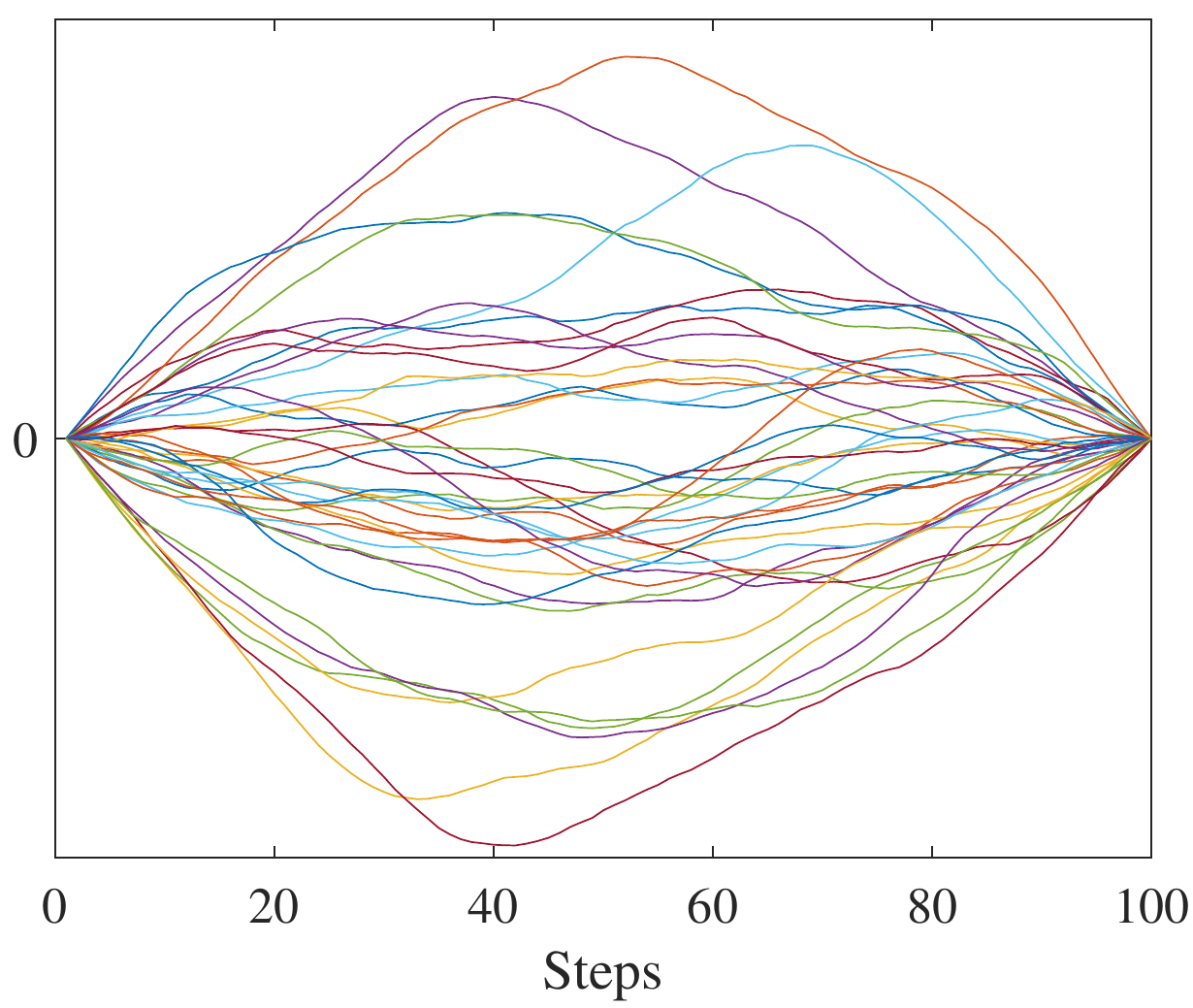}
		\caption{Samples obtained from \eqref{eq:exploration_traj_ini}.}
	\end{subfigure}
	\begin{subfigure}[b]{0.465\columnwidth}
		\centering
		\includegraphics[width=\textwidth]{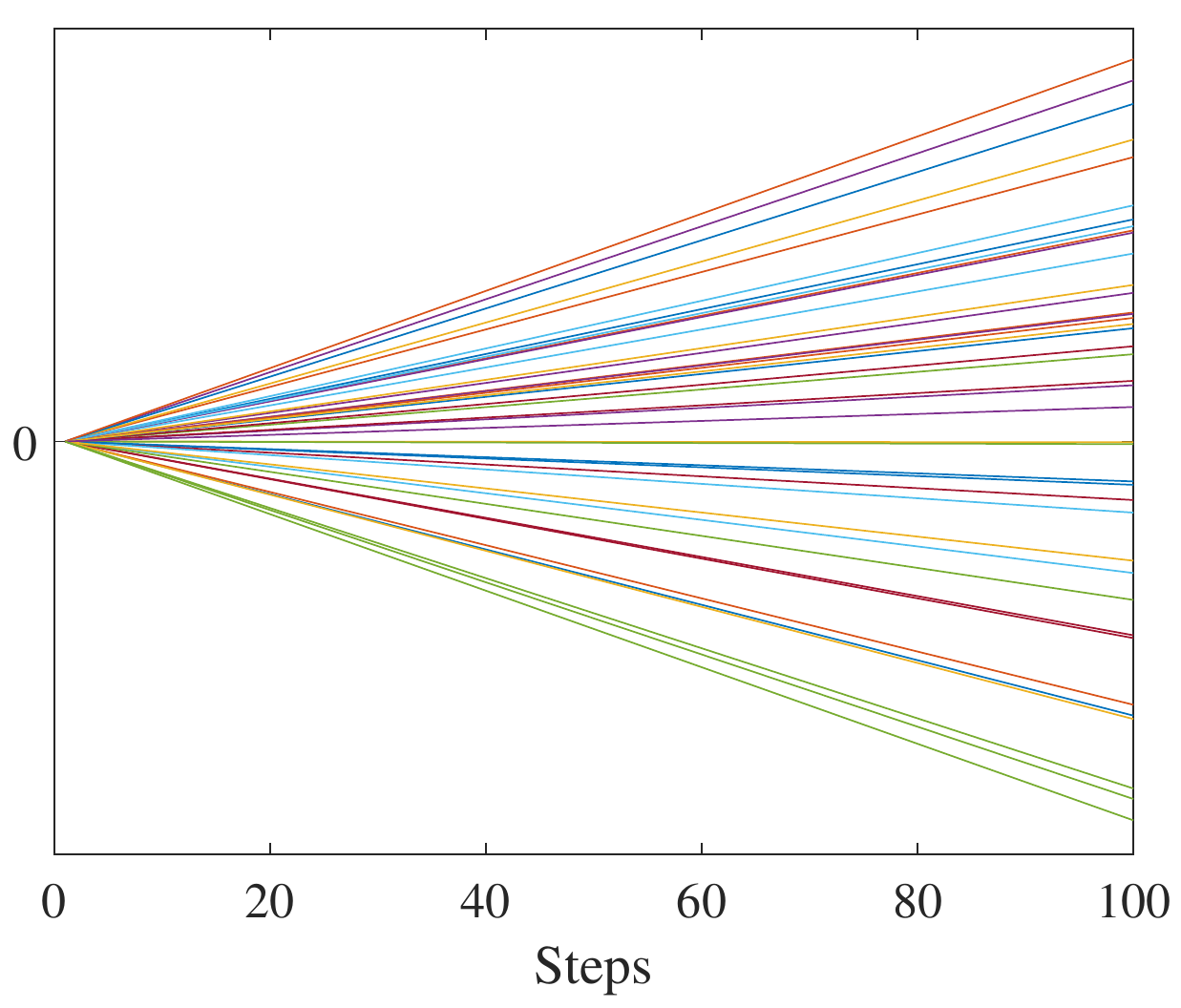}
		\caption{Samples obtained by \eqref{eq:exp_prop}.}
	\end{subfigure}
	\caption{When optimizing the trajectory including the goal configuration, the proposal distribution is given by linear combination of  $\beta_{\textrm{traj}}(\vect{\xi})$ and $\beta_{\textrm{end}}(\vect{\xi})$. 
		(a) shows trajectories drawn from a distribution $\mathcal{N}(\vect{0}, R)$ used in $ \beta_{\textrm{traj}}(\vect{\xi})$.
		(b) shows trajectories drawn from the proposal distribution $\beta_{\textrm{end}}(\vect{\xi})$. The deviation at the goal configuration is linearly propagated to the whole trajectory as shown in (b). 
	} 
	\label{fig:proposal}
\end{figure}

Combining the exploration of the trajectory with the fixed goal configuration and the exploration of goal configuration,
we use the following proposal distribution at the $k$th iteration of the trajectory optimization:
\begin{align}
\beta(\vect{\xi}) = \beta_{\textrm{traj}}(\vect{\xi}) + \beta_{\textrm{end}}(\vect{\xi})
\label{eq:exploration_end}
\end{align}
where 
\begin{align}
\beta_{\textrm{end}}(\vect{\xi}) & =  \vect{M}^{-1}[ 0, \ldots, 0, \Delta \vect{q}_{T}]^{\top}, \\
\Delta \vect{q}_{T} & = \vect{q}^{\textrm{rot}}_T(\varphi_g, \vect{r}_g) - \vect{q}^{k}_T \label{eq:exp_end}, \\
\varphi_g &\sim U(\varphi^{\min}_g, \varphi^{\max}_g),
\end{align}
and $\beta_{\textrm{traj}}(\vect{\xi})$ is given by \eqref{eq:exploration_traj_ini}--\eqref{eq:exploration_traj_k}.
If the manipulator is redundant, \eqref{eq:exp_end} is replaced with 
\begin{align}
\Delta \vect{q}_{T} & = \vect{q}^{\textrm{rot}}_T(\varphi_g, \vect{r}_g) - \vect{q}^0_T + \Delta \vect{q}_{\textrm{null}}, 
\end{align}
where $\Delta \vect{q}_{\textrm{null}}$ is given by \eqref{eq:exp_end_null}.
Fig.~\ref{fig:proposal} visualizes trajectories drawn from  $\mathcal{N}(\vect{0}, aR)$ and $\beta_{\textrm{end}}(\vect{\xi})$.
The deviation at the goal configuration is linearly propagated to the whole trajectory as shown in Fig.~\ref{fig:proposal}(b). 

When applying our motion planning framework to a redundant robotic manipulator, 
the covariant trajectory update can be also modified to optimize the null space configuration at the goal point.
the whole trajectory is updated as follows:
\begin{align}
\vect{\xi}^{\new}  = \vect{\xi}^c - \frac{1}{\eta} M^{-1} [0,\ldots,0, \Delta \tilde{\vect{q}}^{\textrm{null}}_T ]^\top,
\end{align}
where $ \Delta \tilde{\vect{q}}^{\textrm{null}}_T$ is given by
\begin{align}
\Delta \tilde{\vect{q}}^{\textrm{null}}_T = \sum^L_{i=1} a_i 
\frac{ \mathcal{C}(\vect{q}_T + \Delta q^{\nul} \vect{e}^{\nul}_i  ) -  \mathcal{C}(\vect{q}_T)}{\Delta q^{\nul}} \vect{e}^{\nul}_i
\label{eq:update_null}
\end{align}
where $\vect{e}^{\nul}_i$ is a basis vector of the null space of the Jacobian $J(\vect{\vect{q}_T})$, 
$\Delta q^{\nul}$ is a scalar for computing the finite difference of the cost, and $a_i$ is a constant that defines the update rate.

\section{Experiment}
To analyze the performance of the proposed algorithms on simple tasks,
we first evaluate the proposed algorithm on tasks with three-link manipulator in 2D space.
We then evaluate the proposed framework on motion planning tasks with manipulators having six and seven DoFs in 3D space.
To evaluate our method on tasks on 3D space, we set tasks with UR3, which has 6 DoFs, and KUKA Light Weight Robot~(LWR), which has 7 DoFs.

\begin{table}[h]
	\small\sf\centering
	\caption{Hyperparameters Used in the Experiments.}
	\label{tbl:param}
	\begin{tabular}{lll}
		\toprule
		Param. Name & Symbol  & value \\
		\midrule
		\makecell{Max. $\#$ of \\ Solutions} & O in Alg.~\ref{alg:MulTrajOpt} & 10 \\
		$\#$ of samples & N in Alg.~\ref{alg:MulTrajOpt}& \makecell{800 (KUKA LWR) \\ 500 (UR3 $\&$ 2D)}\\
		\makecell{Max. $\#$ of \\ iterations} & K in Alg.~\ref{alg:MulTrajOpt} & 3 \\
		\makecell{Dimension after \\ dim. reduction } &  & 10 \\
		\makecell{$\#$ of steps \\ in a trajectory} & $T$ & 50 \\
		\bottomrule
	\end{tabular}\\[10pt]
\end{table}

\subsection{Implementation Details}
Before describing our experiments, we explain the implementation details\footnote{For reproducibility of the results, codes are available at https://github.com/TakaOsa/SMTO}.
We provide the hyperparameters used in our experiments in Table~\ref{tbl:param}.
In all the tasks described in the following section, the initial trajectory for motion planning is generated by linearly interpolating the given start and goal configurations.
We stop the iteration of updating trajectories when the algorithm finds multiple solutions that satisfy the threshold values of the collision cost.
To perform dimensionality reduction in Line 7 of Algorithm~\ref{alg:MulTrajOpt}, we implement Laplacian Eigenmaps~\citep{Belkin03}, based on the implementation in~\cite{Sugiyama15}.
For cost-weighted density estimation, we draw more samples on tasks with KUKA LWR than on tasks with UR3 and a 2D manipulator.
We observed that more samples are necessary to find proper solutions when the degree of freedom of the manipulator increases.

For motion planning tasks in Section~\ref{sec:2d_link} and \ref{sec:3d_rob}, we used the cost function that was used in CHOMP~\citep{Zucker13}.
To make the paper self-contained, we briefly describe the cost function used in our implementation.

The cost function is given by the sum of the collision cost and the smoothness cost
\begin{align}
\mathcal{C}(\vect{\xi}) = c_{\textrm{obs} }(\vect{\xi}) + \alpha c_{\textrm{smoothenss} }(\vect{\xi}),
\end{align}
where $c_{\textrm{obs} }(\vect{\xi})$ is the collision cost, $c_{\textrm{smoothenss} }(\vect{\xi})$ is the cost associated with the smoothness of the trajectory, and $\alpha$ is a constant.
The smoothness cost  $c_{\textrm{smoothenss} }(\vect{\xi})$ is defined as
\begin{align}
c_{\textrm{smoothenss} }(\vect{\xi}) = \sum^T_{t=1} \left\| \ddot{\vect{q}}_t \right\|^2. 
\end{align}
Denoting by $u$ the index of the body point and by $\vect{x}_{u}(t)$ the position of the body point $u$ in task space at time $t$,
the collision cost  $c_{ \textrm{obs} }(\vect{\xi})$ is given by
\begin{equation}
c_{\textrm{obs} }(\vect{\xi}) = \frac{1}{2} \sum_{t} \sum_{u \in \mathcal{B} } c \left( \vect{x}_{u}(\vect{q}_t) \right)  \left\| \frac{d}{dt}\vect{x}_{u}(\vect{q}_t) \right\|,
\end{equation}
where $\mathcal{B}$ is a set of body points which comprises the robot body. The local collision cost function $c(\vect{x}_{u})$ is defined as
\begin{equation}
c ( \vect{x}_{u} ) =
\left\{
\begin{array}{cll}
0 , & \textrm{if} & d(\vect{x}_{u}) >  \epsilon, \\
\frac{1}{2  \epsilon} (d (\vect{x}_{u}) -   \epsilon)^{2} , & \textrm{if} & 0 < d(\vect{x}_{u}) < \epsilon, \\
- d(\vect{x}_{u})+\frac{1}{2}\epsilon,& \textrm{if} & d(\vect{x}_{u})< 0,
\end{array}
\right. 
\end{equation}
where $\epsilon$ is a constant that scales the margin, 
and $d( \vect{x}_{u} )$ represents a signed distance between the body point $u$ and the nearest obstacle.
$d( \vect{x}_{u} )$ is negative when the body point is inside obstacles, and zero at the boundary.
The cost function proposed by \cite{Zucker13} is designed to be differentiable in areas close to obstacles.
Therefore, when we employ this cost function, the cost function changes smoothly.

For cost-weighted density estimation, we used the following form for $f(\cdot)$ in \eqref{eq:scale_f}:
\begin{align}
	f(\vect{x}) = \exp\left( - \alpha \frac{c - c_{\textrm{max}} }{c_{\textrm{max}} - c_{\textrm{min}} }  \right)
	\label{eq:form_f}
\end{align}
where $c_{\textrm{max}}$ and $c_{\textrm{min}}$ are the maximum and minimum cost values of the samples.
$\alpha$ is a constant that controls the scaling of the cost as in \citep{Haarnoja18}.
We evaluate the effect of $\alpha$ in the following section.

\begin{figure}[t]
	\centering
	\begin{subfigure}[b]{0.465\columnwidth}
		\centering
		\includegraphics[width=\textwidth]{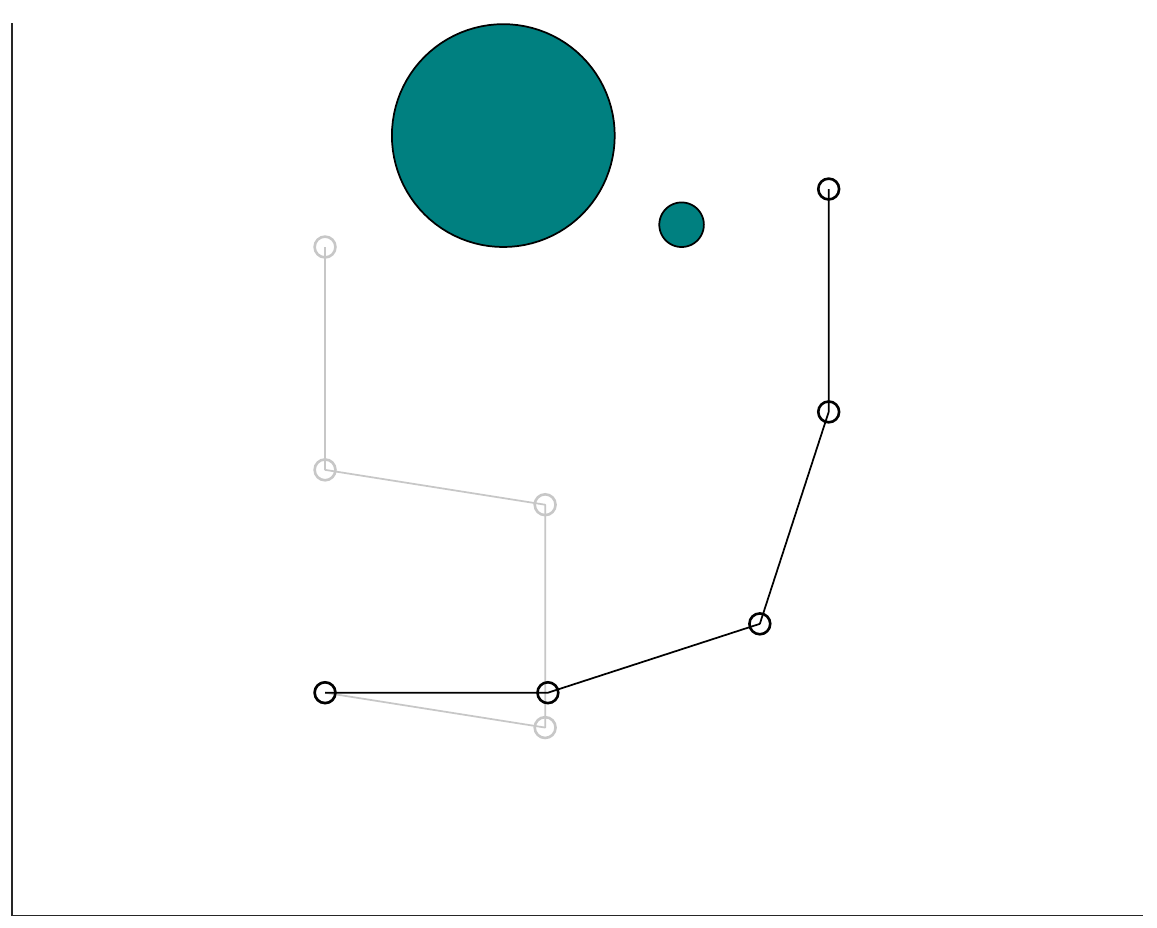}
		\caption{Problem setting.}
	\end{subfigure}
	\begin{subfigure}[b]{0.465\columnwidth}
		\centering
		\includegraphics[width=\textwidth]{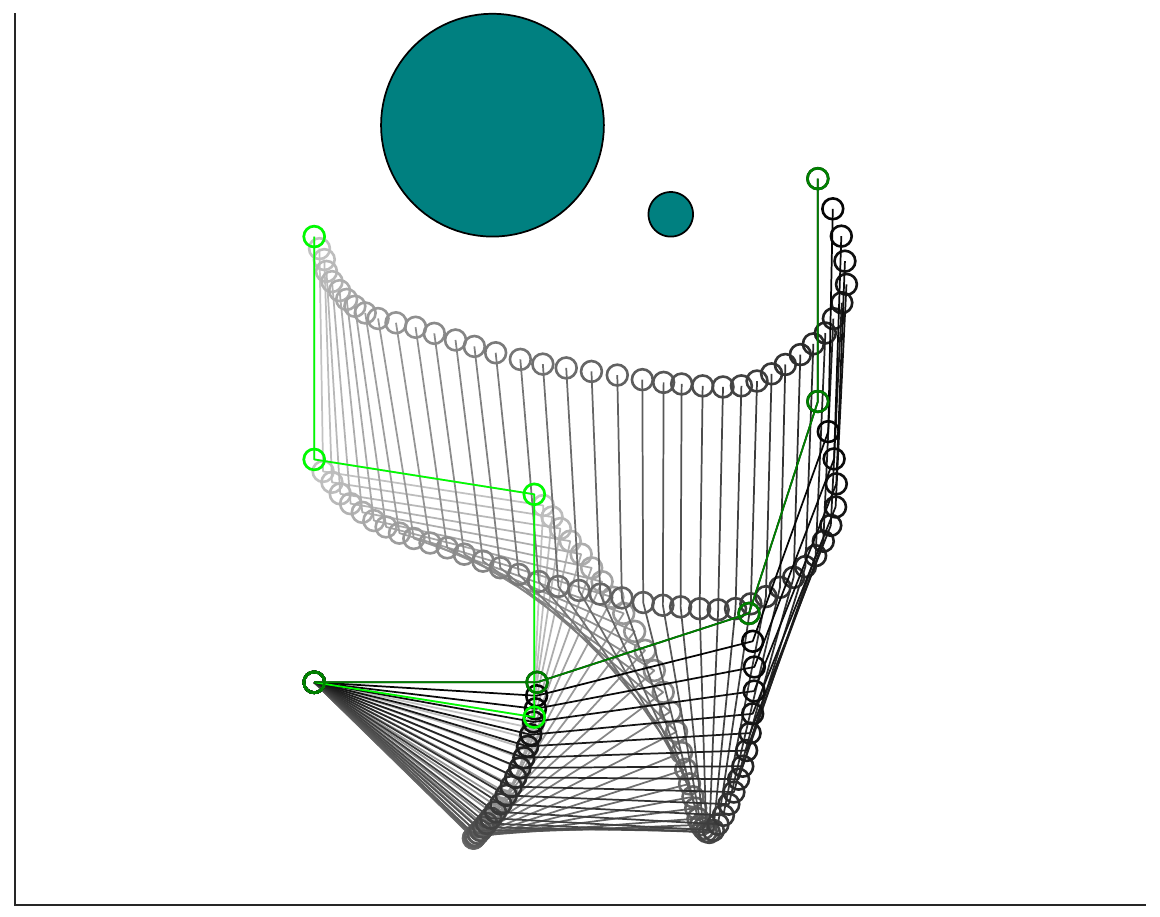}
		\caption{Solution $\#$1.}
	\end{subfigure}
	\begin{subfigure}[b]{0.465\columnwidth}
		\centering
		\includegraphics[width=\textwidth]{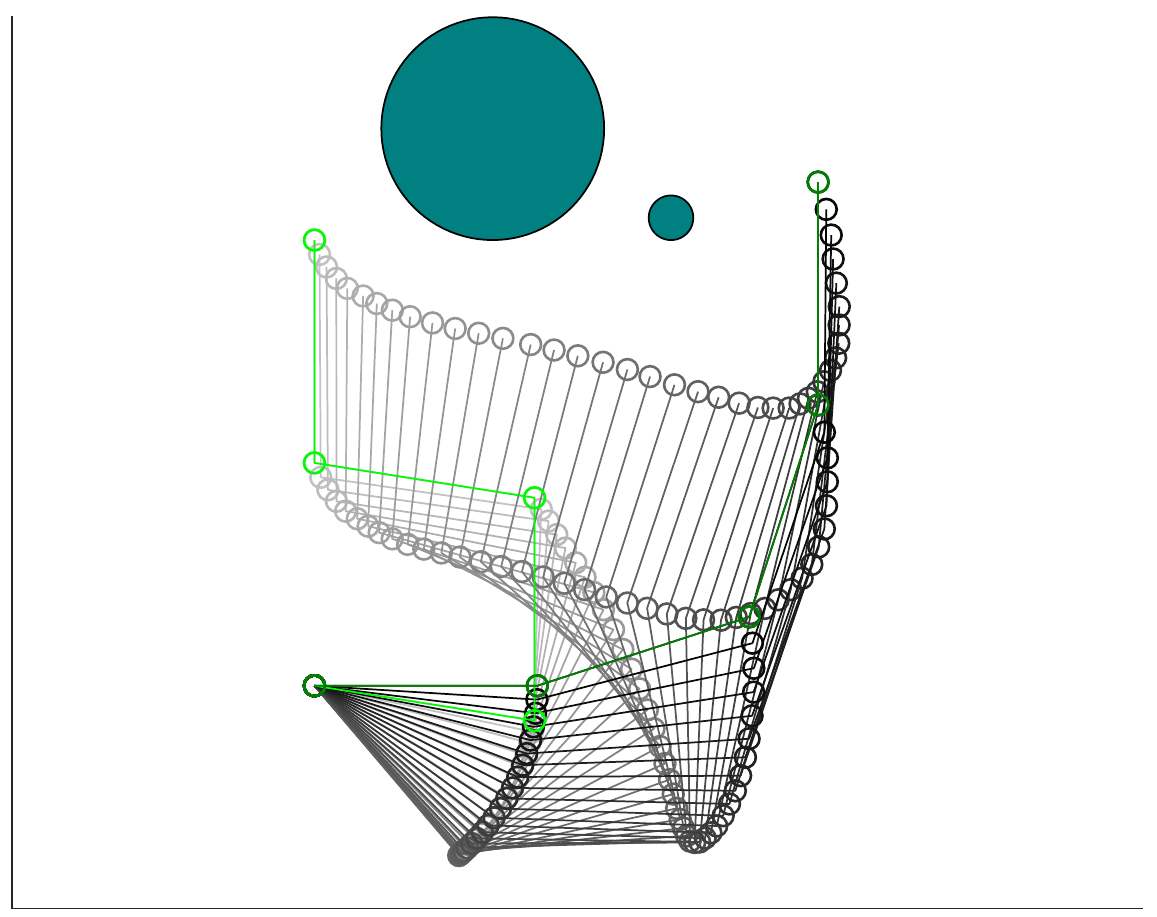}
		\caption{Solution $\#$2.}
	\end{subfigure}
	\begin{subfigure}[b]{0.465\columnwidth}
		\centering
		\includegraphics[width=\textwidth]{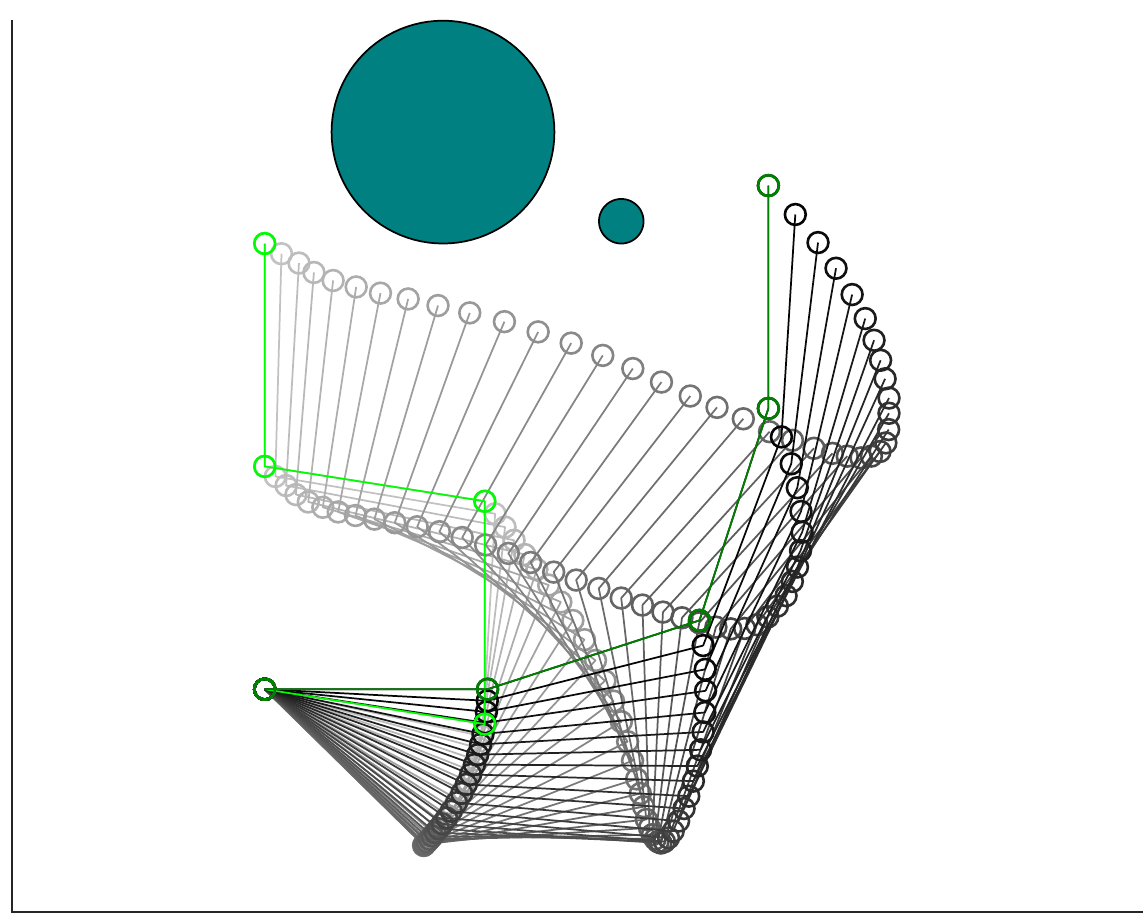}
		\caption{Solution $\#$3.}
	\end{subfigure}
	\caption{An example of the multimodal trajectory optimization with fixed end points. (a) shows given start and goal configurations and obstacles. (b)-(d) show the trajectories obtained from the proposed method. All of them successfully avoid the obstacle in the scene and achieve comparable  costs. 
		Green configurations represent given start and goal positions in (b)-(d).
	} 
	\label{fig:multiTraj2D}
\end{figure}

\subsection{Four-Link Manipulator Tasks in 2D Space}
\label{sec:2d_link}
We first evaluate the proposed SMTO algorithm on tasks with a 2D four-link manipulator.
Although the benefits of SMTO are more prominent on 3D tasks, 
we first show the evaluation on 2D tasks to analyze the performance. 

\subsubsection{Evaluation of Multimodal Trajectory Optimization with Fixed End Points}
In this experiment, the goal configurations are fixed.
Figure~\ref{fig:multiTraj2D} show an example of the behavior of the proposed multimodal trajectory optimization. 
In Fig.~\ref{fig:multiTraj2D}, green solid circles represent obstacles, (a) shows given start and goal configurations and obstacles while (b)-(d) show the trajectories obtained from the proposed method.

As shown, SMTO finds three solutions in this task. All the resulting trajectories achieve collision free motion, while achieving comparable smoothness.
This result indicates that there are multiple modes in the cost function even in this simple motion planning task.
Although the topological shapes of these three trajectories are different, the values of the cost function corresponding to these solutions are comparable.
Therefore, when selecting one of these solutions, one can consider factors that is not encoded in the cost function, which may not be obvious before executing a motion planning algorithm.
Existing motion planning frameworks usually generate only a single solution and ignore other solutions. 
As a result, a user may need to try different initialization or manual tuning of the cost function to obtain different types of solutions. We think that the benefits of our method can be illustrated even in this simple motion planning task.

\begin{figure}[]
	\centering
	\begin{subfigure}[b]{0.45\columnwidth}
		\centering
		\includegraphics[width=\textwidth]{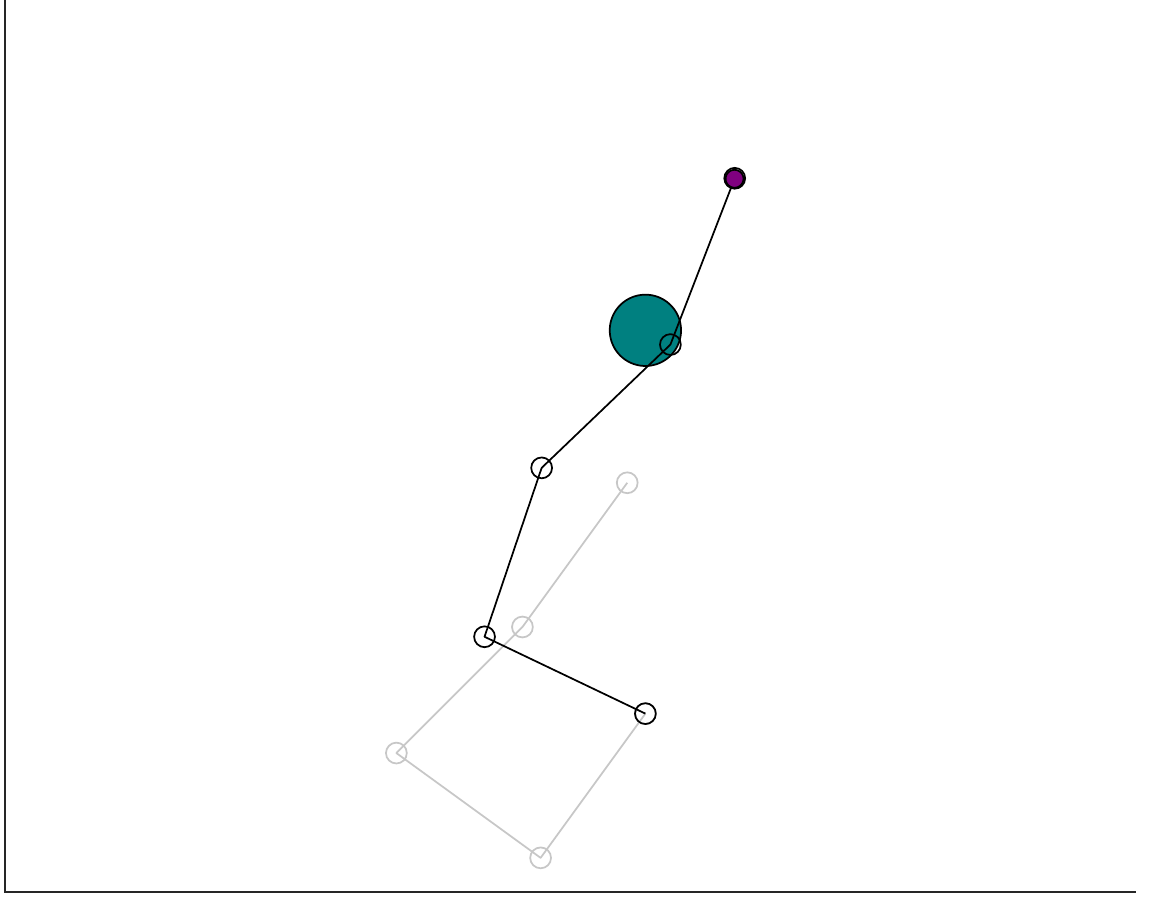}
		\caption{}
	\end{subfigure}
	
	\begin{subfigure}[b]{0.465\columnwidth}
		\centering
		\includegraphics[width=\textwidth]{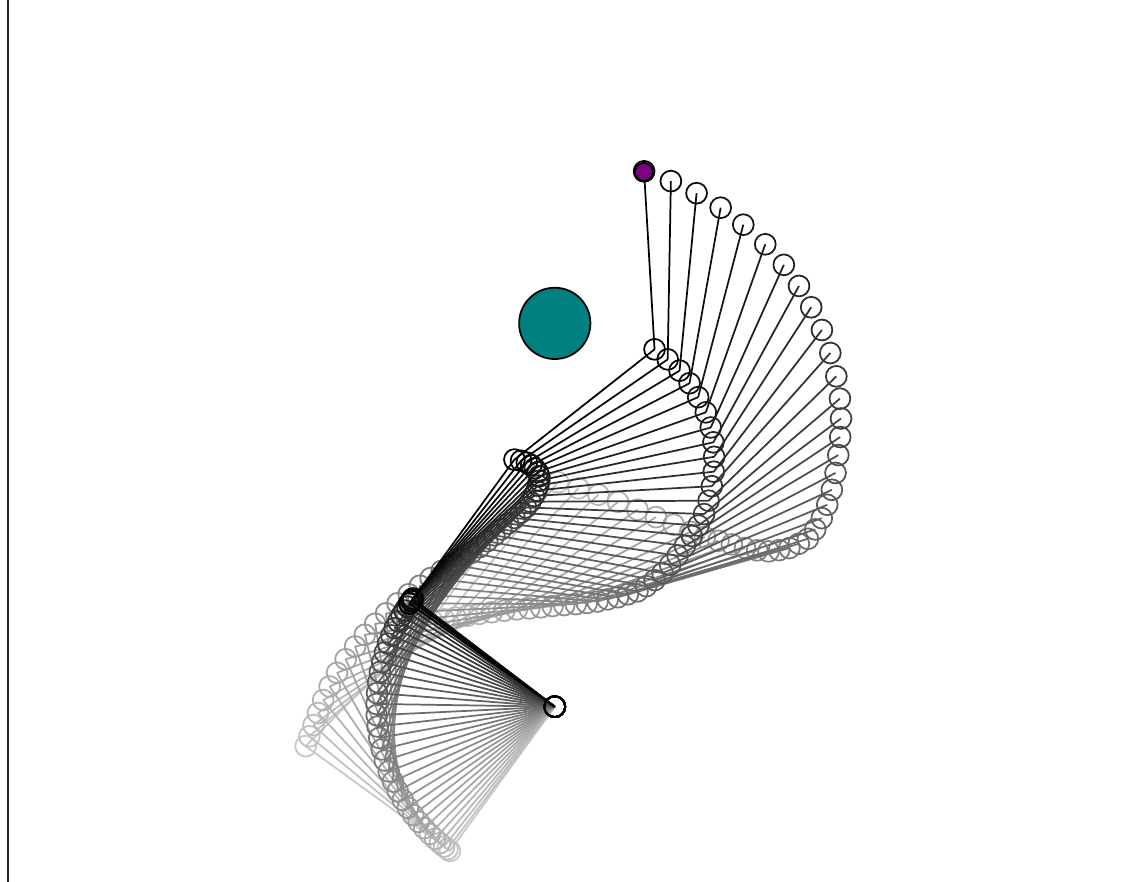}
		\caption{}
	\end{subfigure}
	\hfill
	\begin{subfigure}[b]{0.465\columnwidth}
		\centering
		\includegraphics[width=\textwidth]{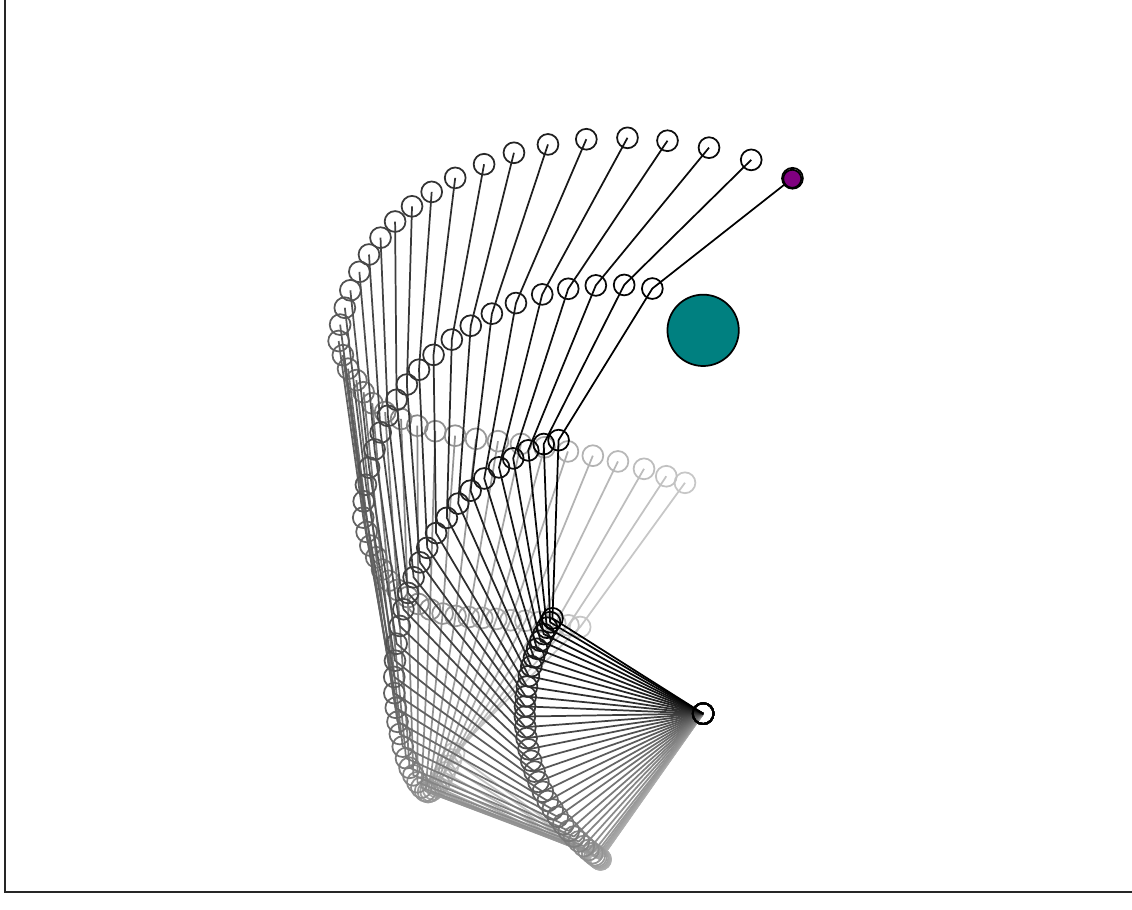}
		\caption{}
	\end{subfigure}
	\caption{An example of the multimodal trajectory optimization with a rotational freedom at the end point. (a) shows given start and goal configurations and an obstacle. (b) and (c) show the trajectories obtained from the proposed method.  The purple point indicates the position of the end-effector in task space of the given goal configuration.
	} 
	\label{fig:multiTrajend2D}
\end{figure}

\subsubsection{Evaluation of Multimodal Trajectory Optimization with a Rotational Freedom at the Goal Point }
In this experiment, the goal configurations have a rotational freedom around the axis vertical to the page. 
Fig.~\ref{fig:multiTrajend2D} shows an example of the behavior of the proposed multimodal trajectory optimization. 
Fig.~\ref{fig:multiTrajend2D}(a) shows the start and goal configurations and obstacles. 
We provided a goal configuration that collided with the given obstacle in order to verify that our method could successfully find collision-free goal configurations.

Fig.~\ref{fig:multiTrajend2D}(b) and (c) show the results of SMTO with a flexible goal configuration. 
The proposed method finds two trajectories that can achieve collision-free motions with orientations different from the given goal configuration. 
In both solutions, the position of the end-effector in task space at the goal configuration is the same as that of the initial goal configuration.
This result indicates that our method can deal with multiple modes of the cost function even when we need to optimize the goal configuration along a given rotational freedom.

\subsection{Motion Planning with a Manipulator in 3D Space}
\label{sec:3d_rob}

To evaluate our motion planning framework in 3D space, we applied our proposed method to UR3 which has 6 DoFs and KUKA Light Weight Robot~(LWR) which has 7 DoFs. 
Since KUKA LWR is a redundant manipulator, we applied the optimization of the null space motion in \eqref{eq:update_null}.
For visualizing the motions of manipulators, we used V-REP as a simulator~\citep{Rohmer13}.

\subsubsection{Evaluation of Multimodal Trajectory Optimization with Fixed End Points}
\begin{figure*}[]
	\centering
	\begin{subfigure}[b]{0.46\textwidth}
		\centering
		\includegraphics[width=\textwidth]{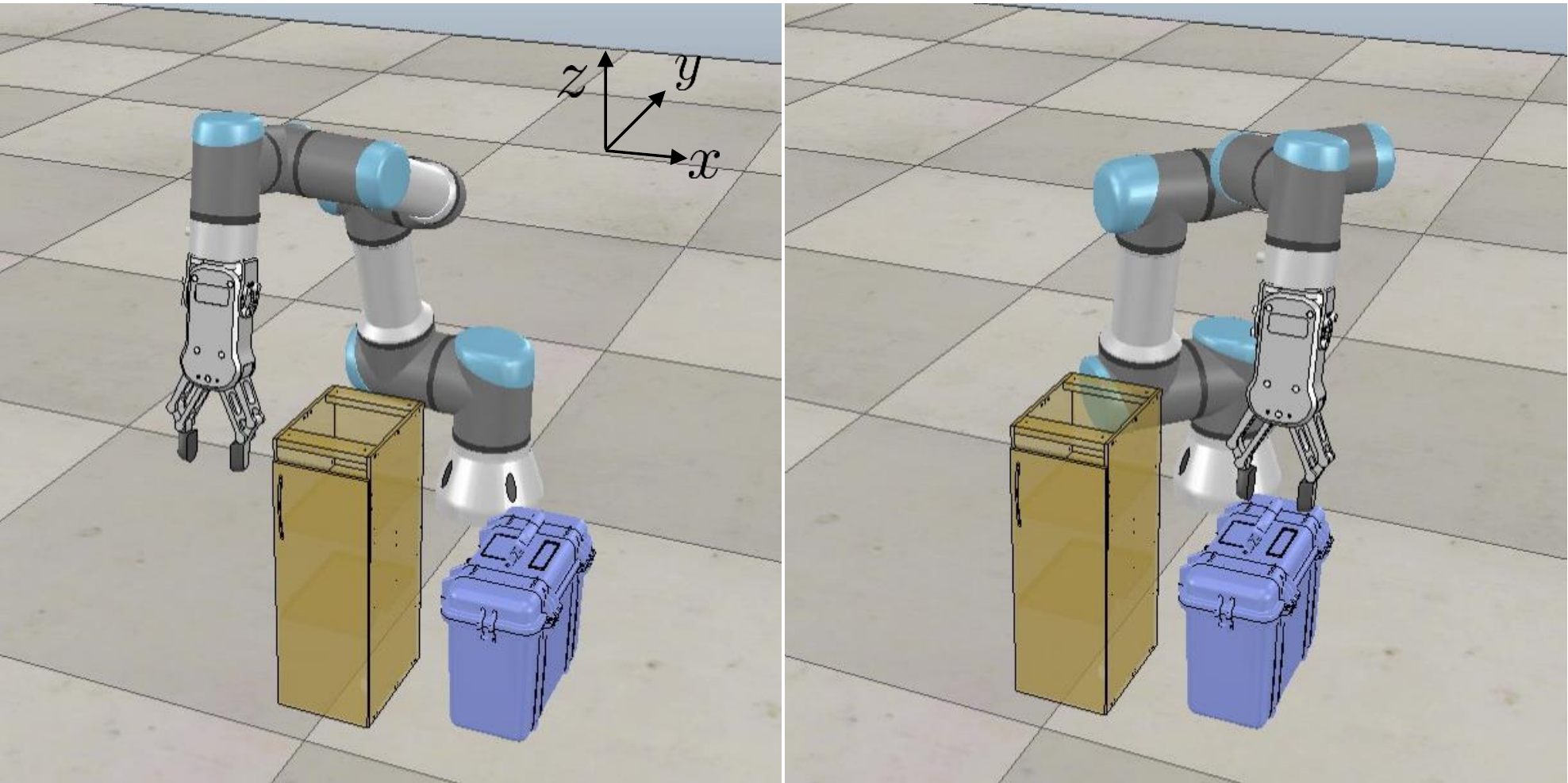}
		\caption{Problem setting of the bag grasping task with UR3.  }
		\label{fig:UR_case_setting}
	\end{subfigure}
	\hfill
	\begin{subfigure}[b]{0.46\textwidth}
		\centering
		\includegraphics[width=0.65\textwidth]{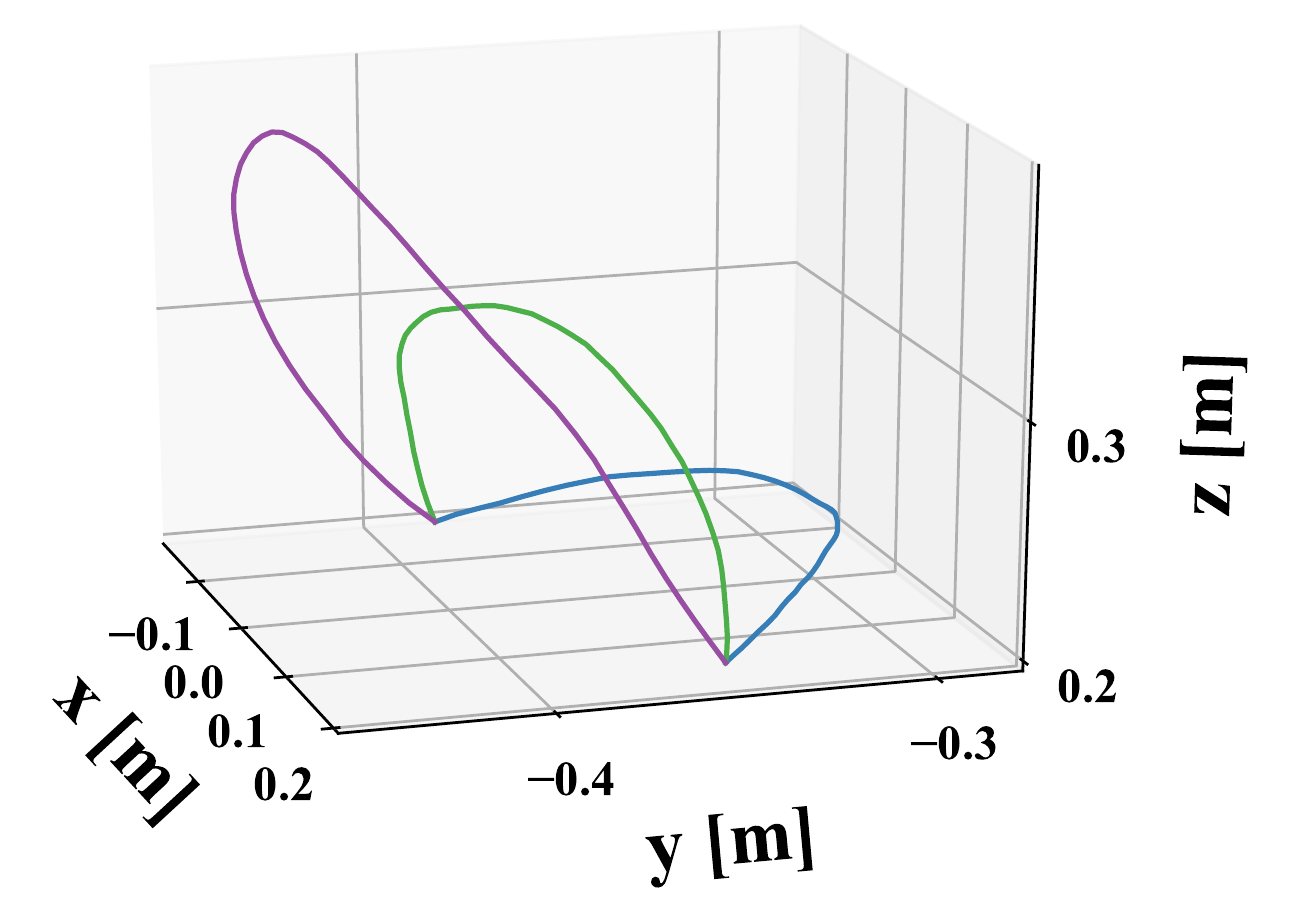}
		\caption{Trajectories in task space.  }
		\label{fig:UR3_case_task}
	\end{subfigure}
	\hfill
	\begin{subfigure}[b]{\textwidth}
		\centering
		\includegraphics[width=\textwidth]{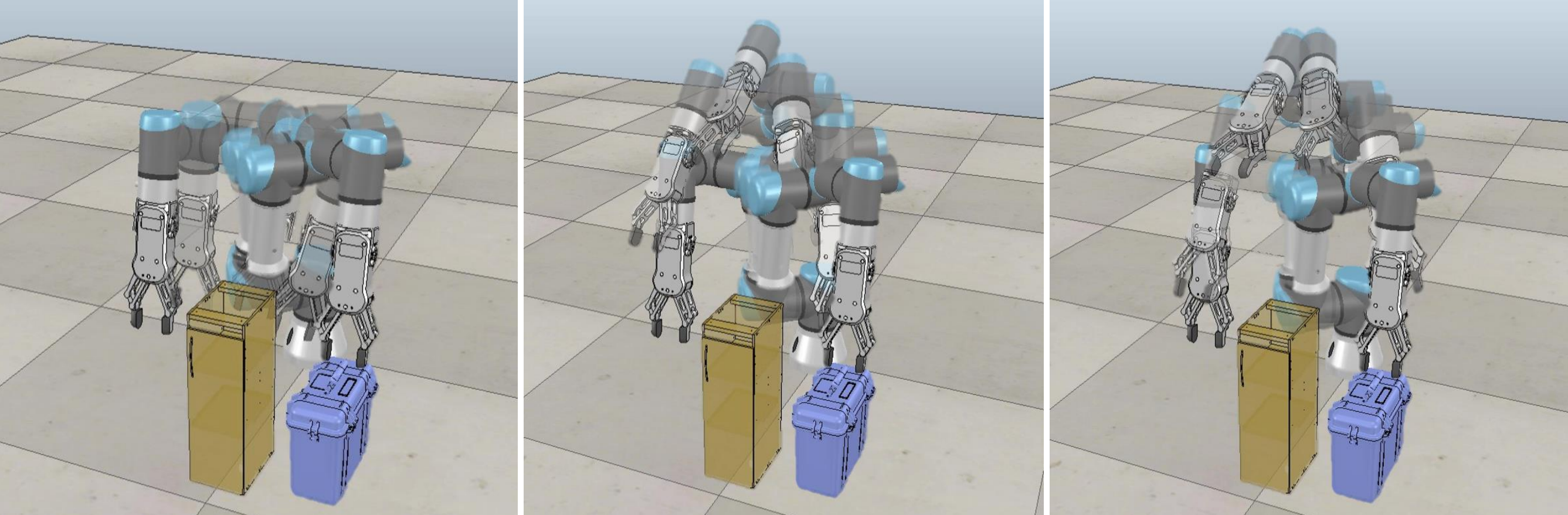}
		\caption{Results of the task with UR3. Three trajectories planned for the case grasping task are visualized. All the trajectories reach the goal configuration without collision.  The values of the cost function for the obtained solutions are, from left to right, 3.1, 2.36, 2.33, respectively. }
		\label{fig:UR_case_sequence}
	\end{subfigure}
	\hfill
	\begin{subfigure}[b]{0.46\textwidth}
		\centering
		\includegraphics[width=\textwidth]{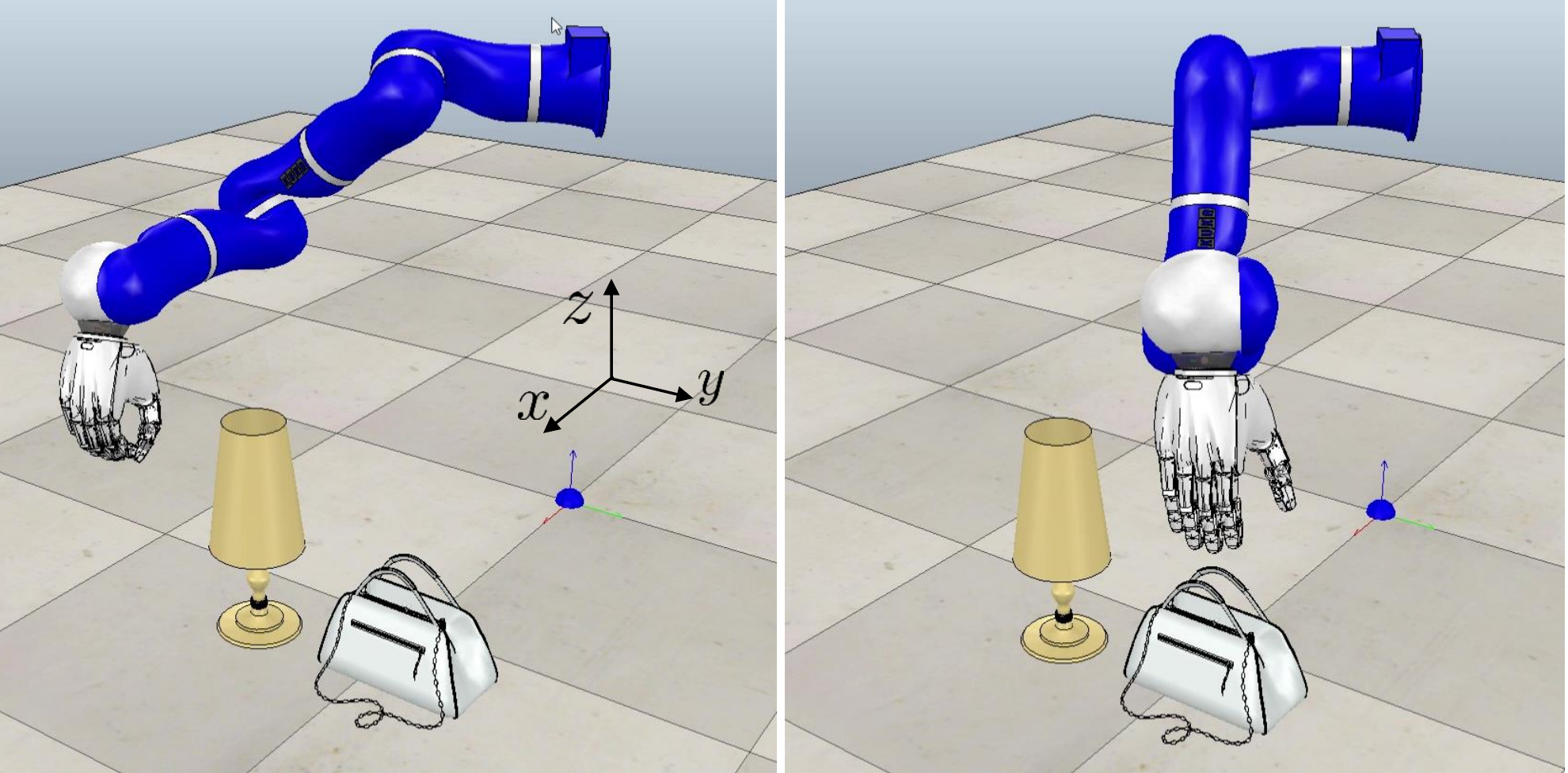}
		\caption{Problem setting of the bag grasping task with KUKA LWR.  }
		\label{fig:Darias_bag_setting}
	\end{subfigure}
	\hfill
	\begin{subfigure}[b]{0.46\textwidth}
		\centering
		\includegraphics[width=0.65\textwidth]{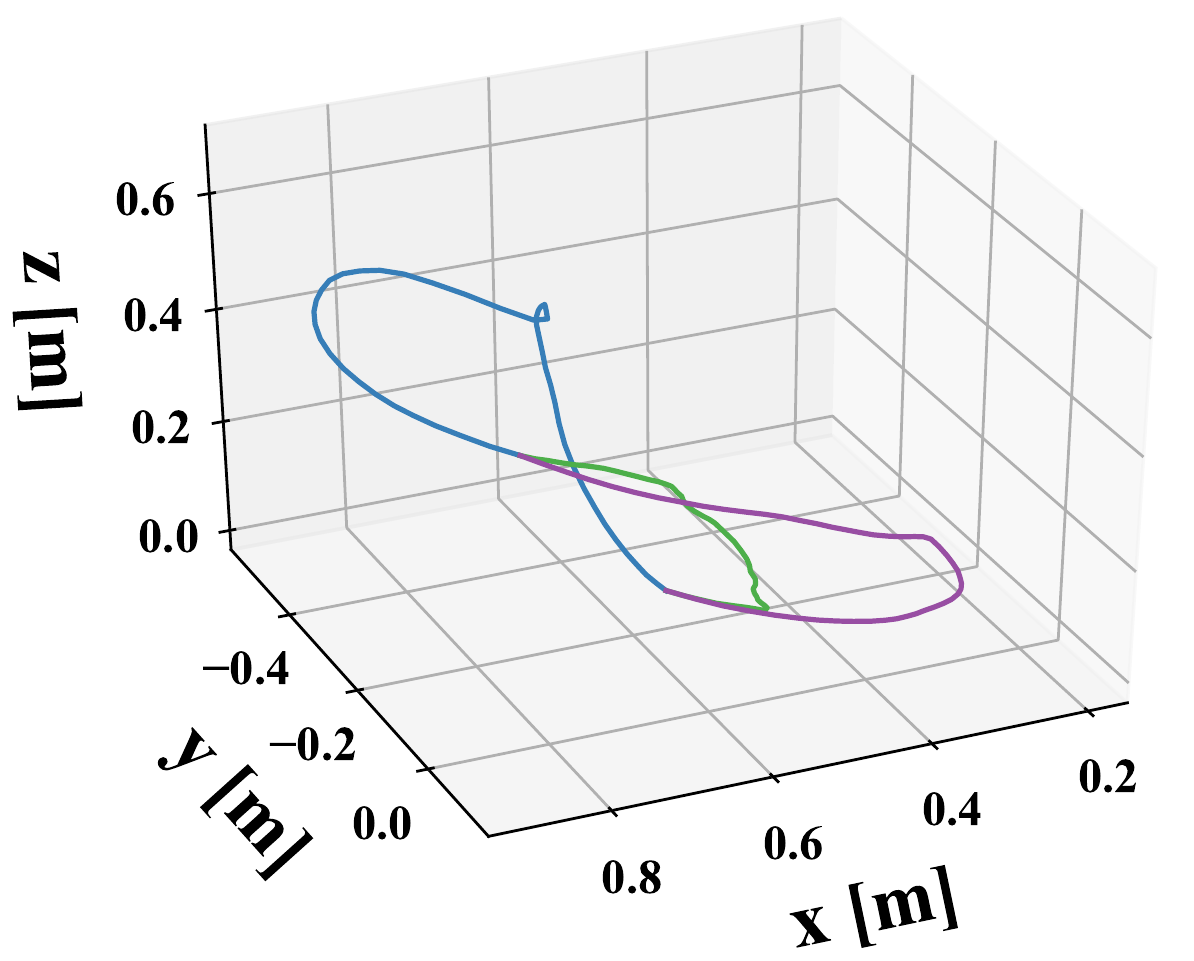}
		\caption{Trajectories in task space.  }
		\label{fig:Darias_bag_task}
	\end{subfigure}
	\hfill
	\begin{subfigure}[b]{\textwidth}
		\centering
		\includegraphics[width=\textwidth]{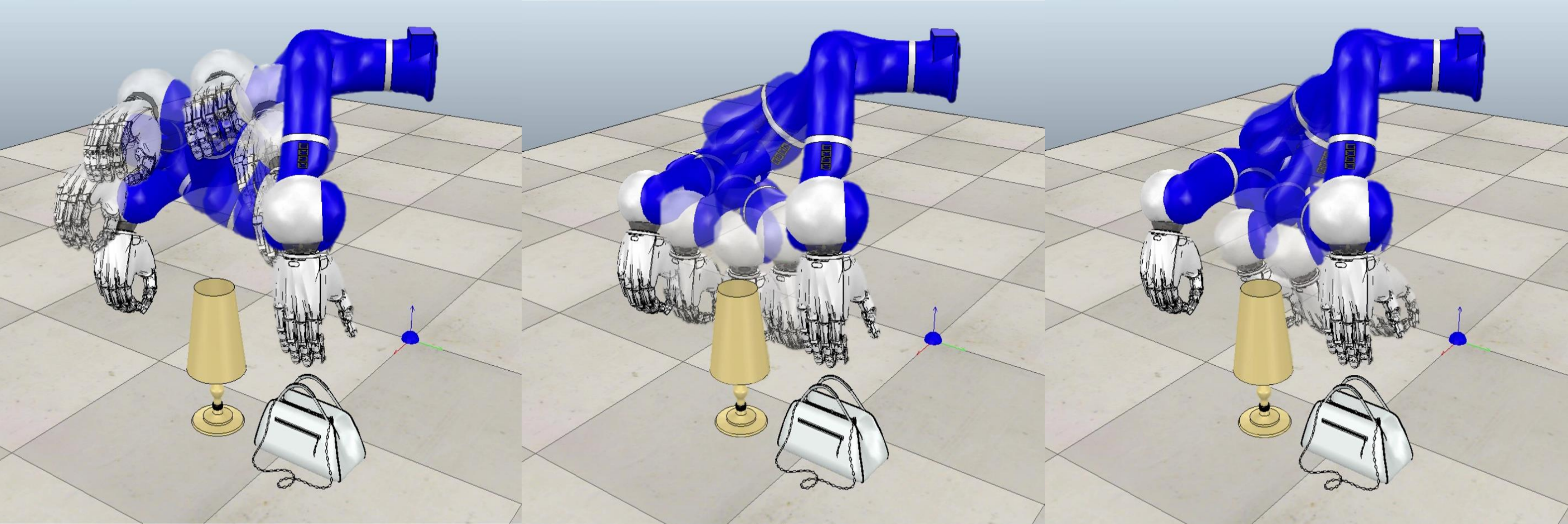}
		\caption{Results of the task with KUKA LWR. Three trajectories planned for the bag grasping task are visualized. All the trajectories reach the goal configuration without collision. The values of the cost function for the obtained solutions are, from left to right, 1.75, 1.80, 1.85, 2.1, respectively.
		}
		\label{fig:Darias_bag_sequence}
	\end{subfigure}
	\caption{Experiments on tasks with fixed goal configurations. In (a) and (b), the left and right figures show the given start and goal configurations, respectively. Diverse solutions are obtained by our method. (b) and (e) show the trajectories of the end effector in task space.}
\end{figure*}

First, we applied our proposed method without the goal configuration optimization.
The problem settings are shown in Figs.~\ref{fig:UR_case_setting} and \ref{fig:Darias_bag_setting}.
The goal of this task is to plan a collision-free trajectory to reach the pre-grasp position for grasping a case in the scene.

In this planning task with UR3, three trajectories are obtained from the proposed motion planning framework as shown in Fig.~\ref{fig:UR_case_sequence}.
The hand goes over the box in one of the solutions, and behind the box in the other two solutions.
Similar solutions are obtained when the proposed method is applied to a task with KUKA LWR, as shown in Fig.~\ref{fig:Darias_bag_setting}.
As shown in Fig.~\ref{fig:Darias_bag_sequence}, SMTO finds three solutions for this task.
One can see that these obtained solutions are qualitatively acceptable for performing the task.

This result is interesting in the sense that it shows that the cost function used in the existing motion planning method can have multiple modes in practice.
As described in the previous section, we used the cost function used in CHOMP~\citep{Zucker13}.
Variants of this cost function have been employed in other motion planning methods such as~\citep{Kalakrishnan11,Mukadam18} 
Our result indicates that our proposed method can handle multiple modes of the cost function, which has often been neglected in prior work.

With respect to the cost, the trajectory drawn in the left of Fig.~\ref{fig:Darias_bag_sequence} has the lowest cost among the obtained solutions.
However, we think that a user may actually prefer one of the other solutions, as shown Fig.~\ref{fig:Darias_bag_sequence} if she/he thinks that the motion going over the lamp is not preferable.
The cost function implemented in a motion planning program usually encodes only commonly used factors, e.g. the collision and smoothness cost. 
The actual values of the cost function are defined by the weights of multiple terms, which are often determined in a tedious way.
It is beneficial for users if the motion planning framework can handle the multimodality of the cost function and can propose multiple solutions.

\begin{table}[!t]
	\small\sf\centering
	\caption{Comparison of the average smoothness of resulting trajectories.}
	\label{tbl:smoothness}

	\begin{tabular}{cccc}
		\hline
		\makecell{SMTO \\ (Ave.)} & \makecell{SMTO \\ (Best)} & CHOMP & STOMP \\
		\hline
		1.404 &	1.402 &	1.404 &	1.402 \\
		\hline
	\end{tabular}
\end{table}

\begin{table}[!t]
	\small\sf\centering
	\caption{Comparison of the average computation time.}
	\label{tbl:time}
	\begin{tabular}{cccc}
		\toprule
		& SMTO & CHOMP & STOMP \\
		\midrule
		Time [s] & 72.9 & 19.9 & 46.3 \\
		\bottomrule
	\end{tabular}
\end{table}

Comparison of the computation time and the smoothness of the resulting trajectories with existing motion planning methods are shown in Tables~\ref{tbl:smoothness} and \ref{tbl:time}.
To quantify smoothness, we show $\frac{1}{N} \sum^{T}_{t=1} \left\| \ddot{\vect{q}}_t \right\|^2$ in Tables~\ref{tbl:smoothness}.
When measuring the computation time, we used a computer with Core-i7-8750H and 16.0GB memory.
The smoothness of the resulting trajectories are comparable on these tasks.
Since our method SMTO generates multiple solutions, we show the average of the smoothness of all solutions and the smoothness of the most smooth trajectory in Tables~\ref{tbl:smoothness}.
On the other hand, the computational time required for our proposed method is longer than those required for CHOMP and STOMP. 
While the dominant factor of the computational cost in CHOMP and STOMP is the number of trajectory updates,
that of our method is the number of trajectory samples for cost-weighted density estimation.

The most computationally expensive part of our method is sampling and evaluating trajectories for performing the cost-weighted density estimation. 
When the number of DoFs increases, the necessary number of samples also increases. 
Therefore, the computational time is longer for the task with KUKA LWR than for UR3.

From this experiment, it is seen that our method can find multiple solutions for these tasks, which are not feasible in most of the existing motion planning methods.
When using an existing motion planning framework, obtaining different types of solutions requires manual tuning of hyperparameters in practice.
As compared with such efforts, we think that the computation time required for our motion planning is negligible.

\begin{figure*}[]
	\centering
	\begin{subfigure}[b]{\columnwidth}
		\centering
		\includegraphics[width=\textwidth]{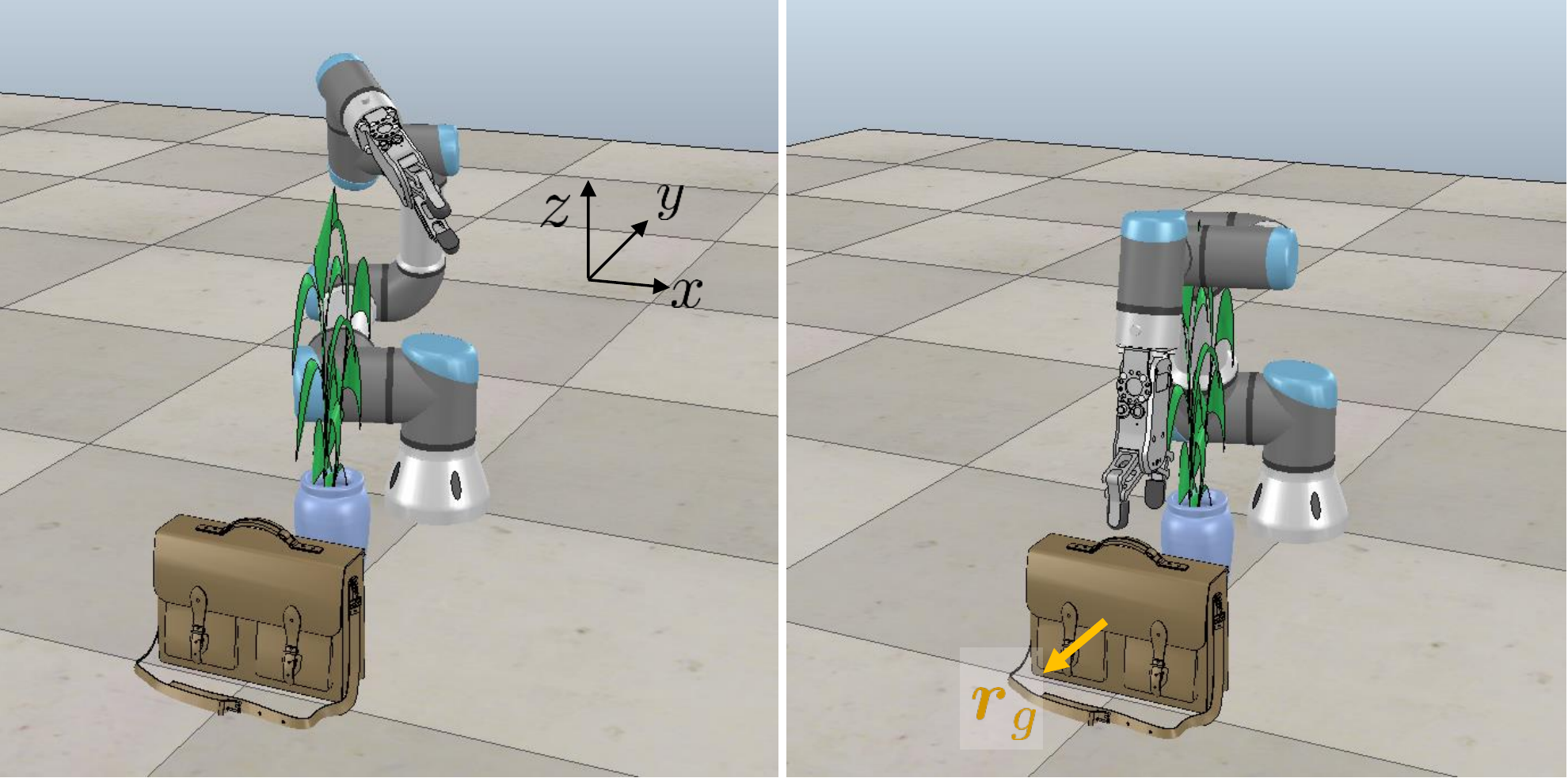}
		\caption{Problem setting of the bag grasping task with UR3.  }
		\label{fig:UR_bag_setting}
	\end{subfigure}
	\hfill
	\begin{subfigure}[b]{\columnwidth}
		\centering
		\includegraphics[width=\textwidth]{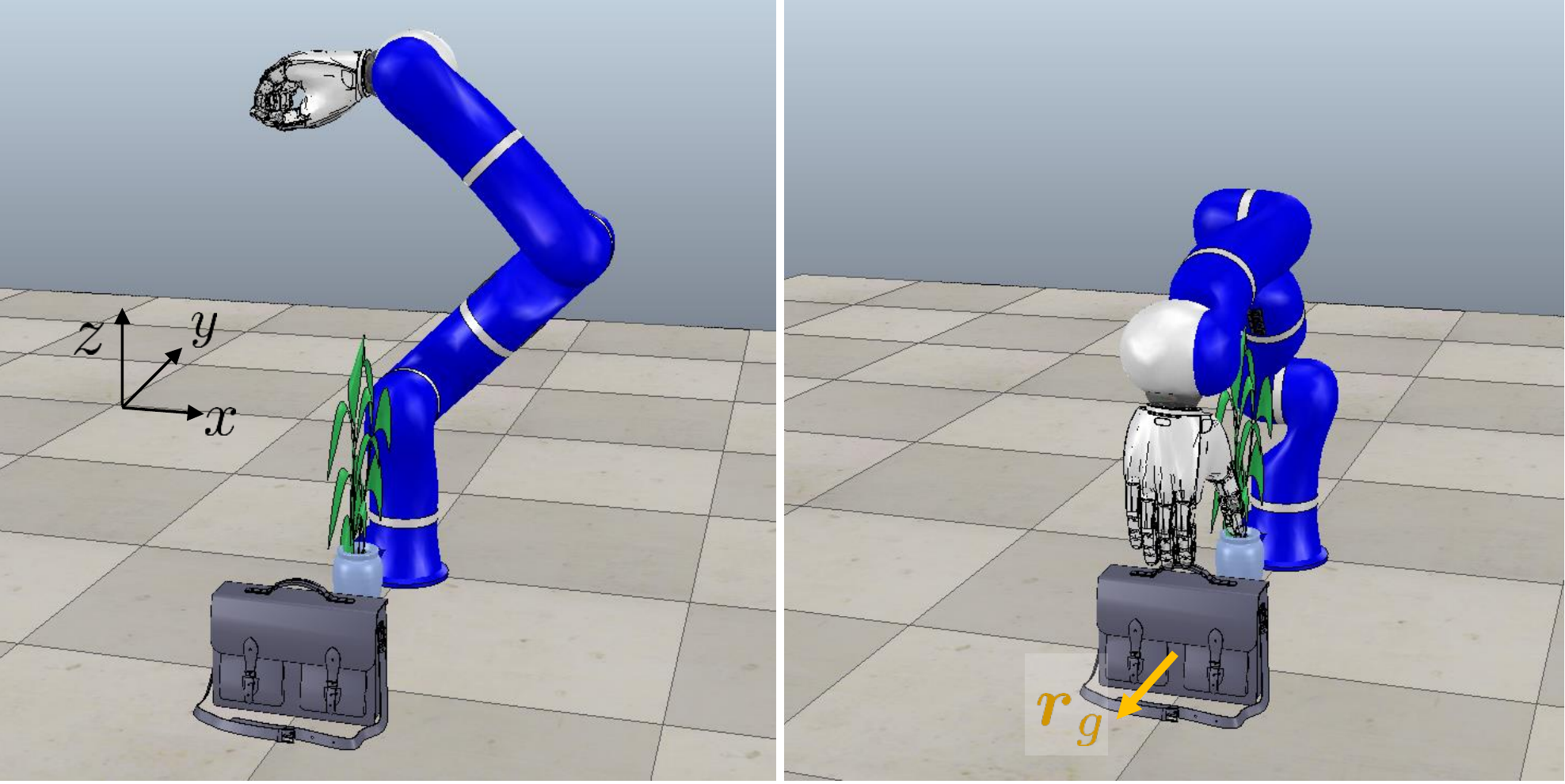}
		\caption{Problem setting of the bag grasping task with KUKA LWR.}
		\label{fig:Darias_bag2_setting}
	\end{subfigure}
	\hfill
	\begin{subfigure}[b]{\columnwidth}
		\centering
		\includegraphics[width=\textwidth]{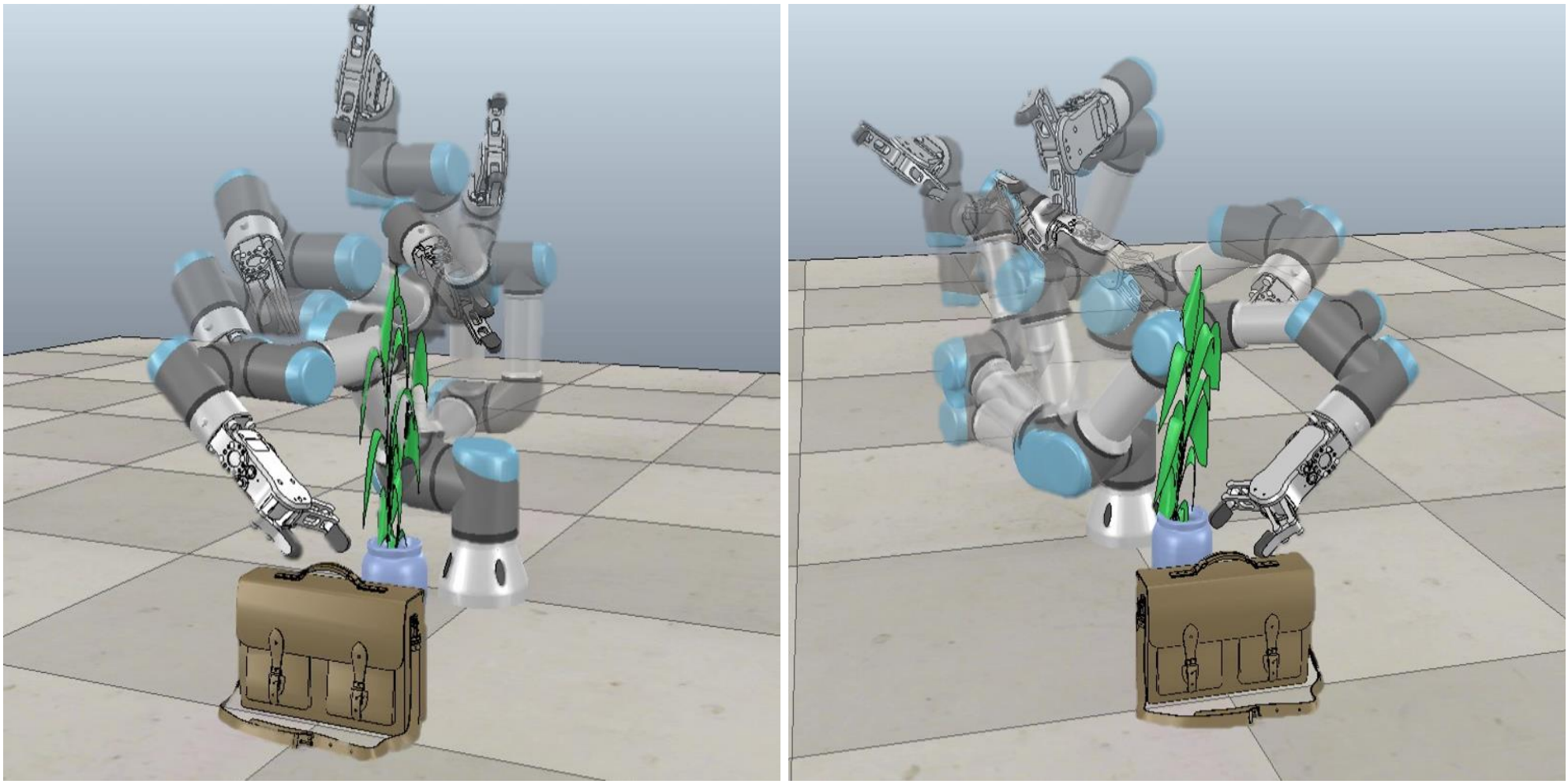}
		\caption{Results of the bag grasping task with UR3. The proposed method finds two solutions as shown in the left and right figures. The viewpoints are different in the left and right figures. The values of the cost function for the obtained solutions in the left and right figures are 3.52 and 3.58, respectively.}
		\label{fig:UR_bag_sequence}
	\end{subfigure}
	\hfill
	\begin{subfigure}[b]{\columnwidth}
		\centering
		\includegraphics[width=\textwidth]{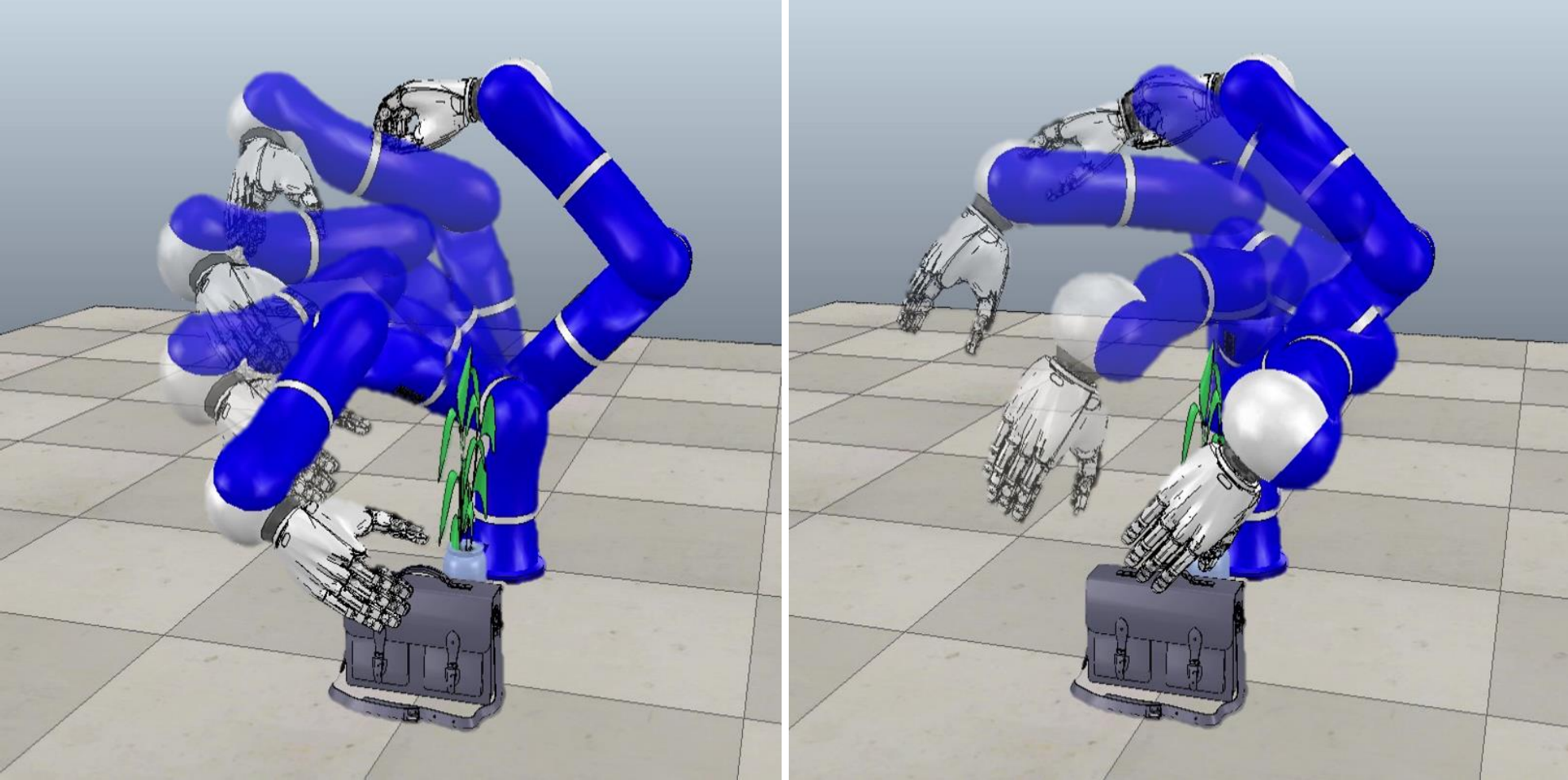}
		\caption{Results of the bag grasping task with KUKA LWR. The proposed method finds two solutions as shown in the left and right figures. The values of the cost function for the obtained solutions in the left and right figures are 1.02 and 2.25, respectively.}
		\label{fig:Darias_bag2_sequence}
	\end{subfigure}
	\hfill
	\begin{subfigure}[b]{\columnwidth}
		\centering
		\includegraphics[width=0.7\textwidth]{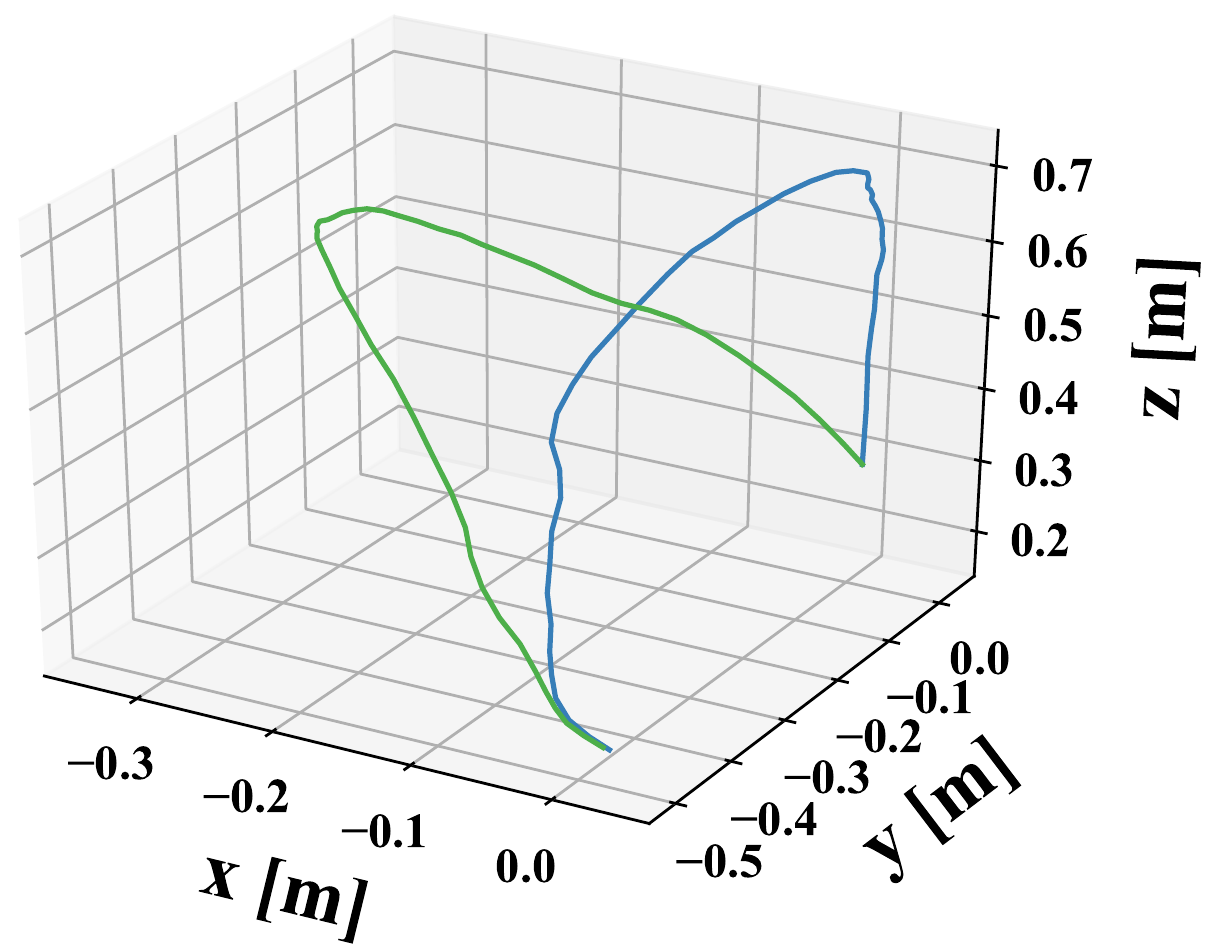}
		\caption{Trajectories of the end effector in task space on the task with UR3.}
		\label{fig:UR_bag_task}
	\end{subfigure}
	\hfill
	\begin{subfigure}[b]{\columnwidth}
		\centering
		\includegraphics[width=0.7\textwidth]{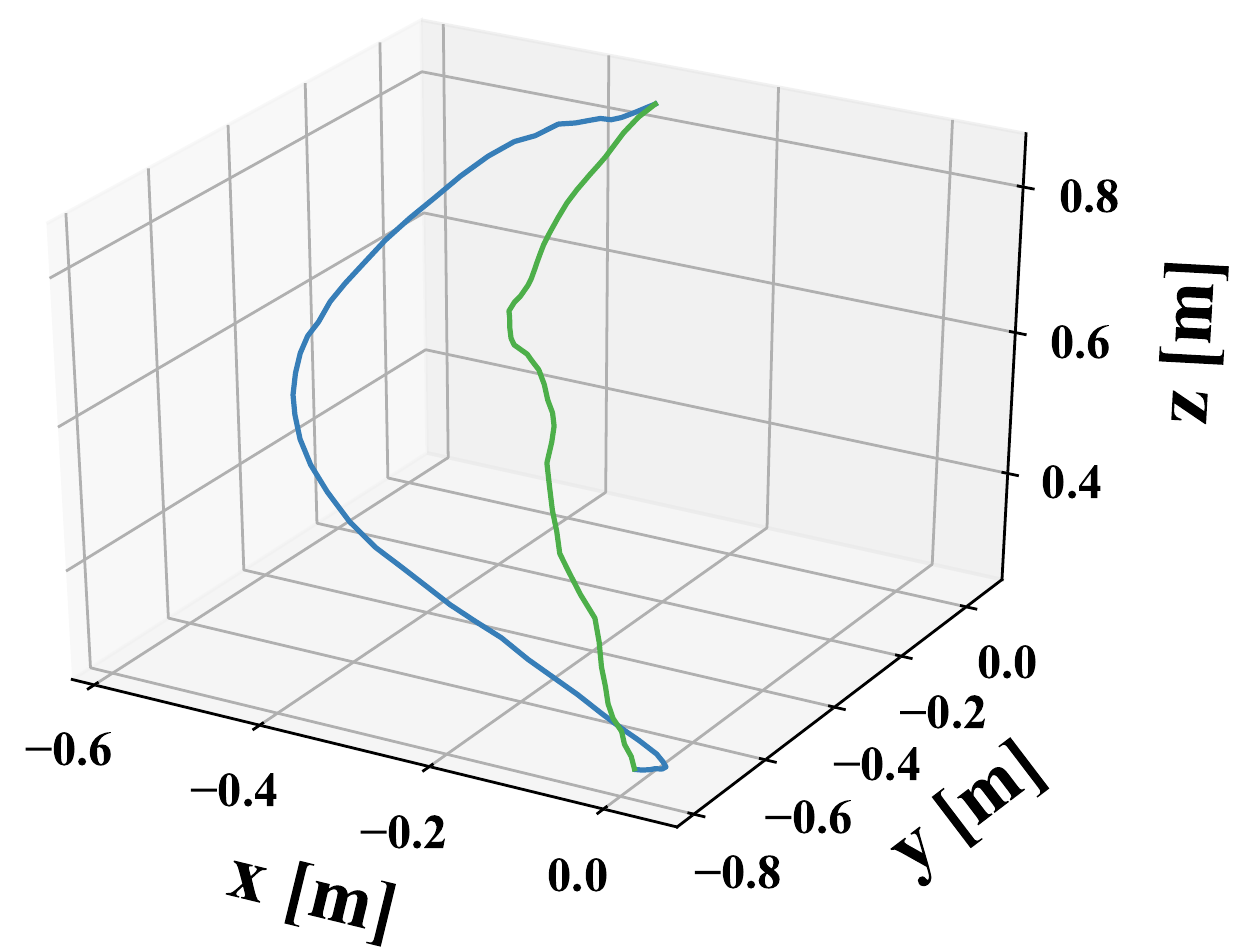}
		\caption{Trajectories of the end effector in task space on the task with KUKA LWR.}
		\label{fig:Darias_bag2_task}
	\end{subfigure}
	\caption{Experiments on tasks with a rotational freedom at the goal configuration. In (a) and (b), the left and right figures show the given start and goal configurations, respectively.  }
\end{figure*}

\subsubsection{Evaluation of Multimodal Trajectory Optimization with a Rotational Freedom at the Goal Point }

We evaluated the proposed motion planning method with a rotation freedom at the goal point.
Namely, the goal configuration is explored by using the strategy in \eqref{eq:exploration_end},
and the planned trajectories are projected onto the constraint solution space as in \eqref{eq:projection}.
In this task, the given goal configuration is made to collide with obstacles, and the algorithm aims to find trajectories that achieve the same end-effector position in the task space without any collision.

\begin{figure*}[]
	\centering
	\begin{subfigure}[b]{0.46\textwidth}
		\centering
		\includegraphics[width=\textwidth]{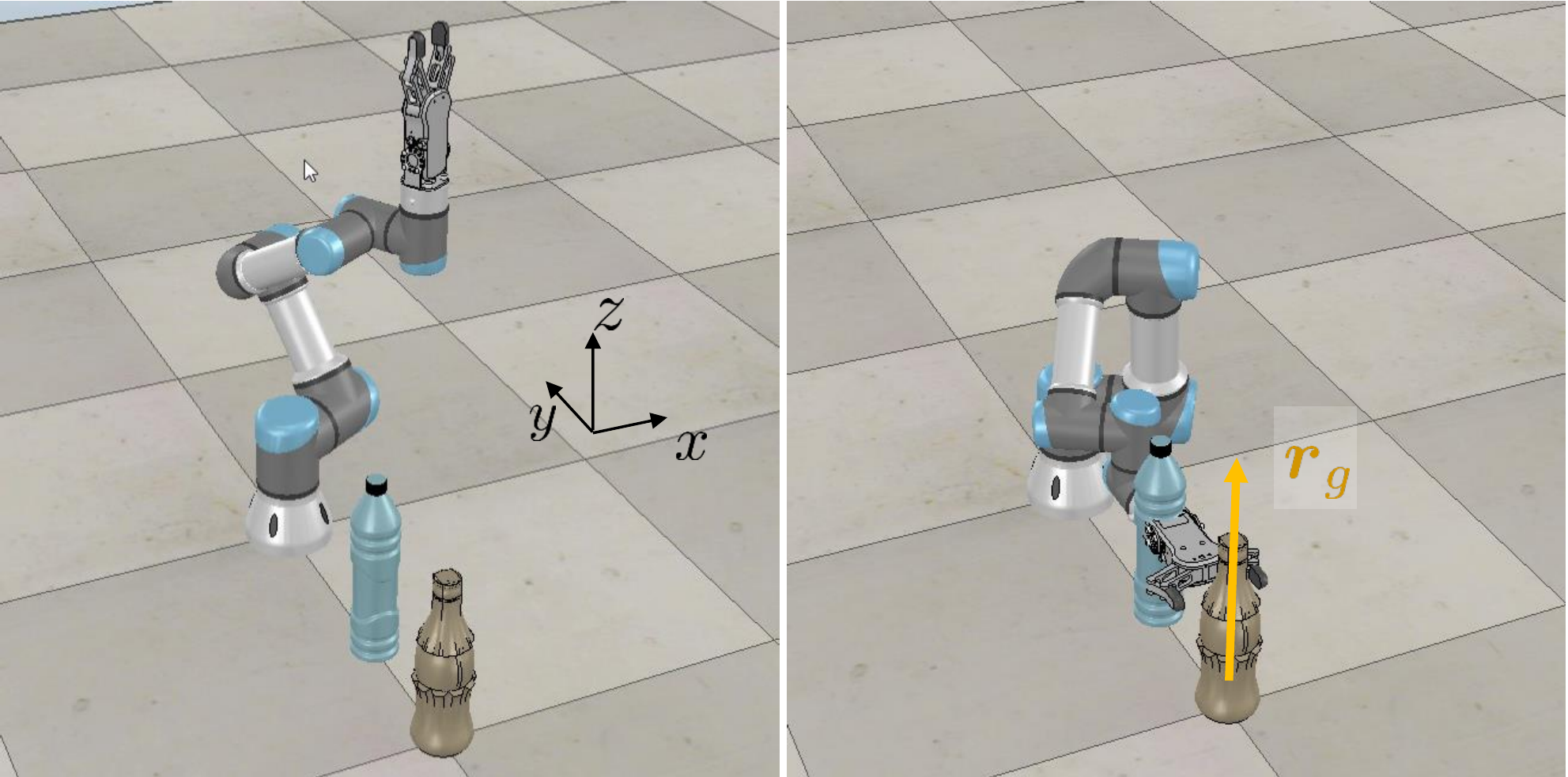}
		\caption{Problem setting for the bottle grasping task. }
		\label{fig:UR_bottle_setting}
	\end{subfigure}
	\hfill
	\begin{subfigure}[b]{0.46\textwidth}
		\centering
		\includegraphics[width=0.55\textwidth]{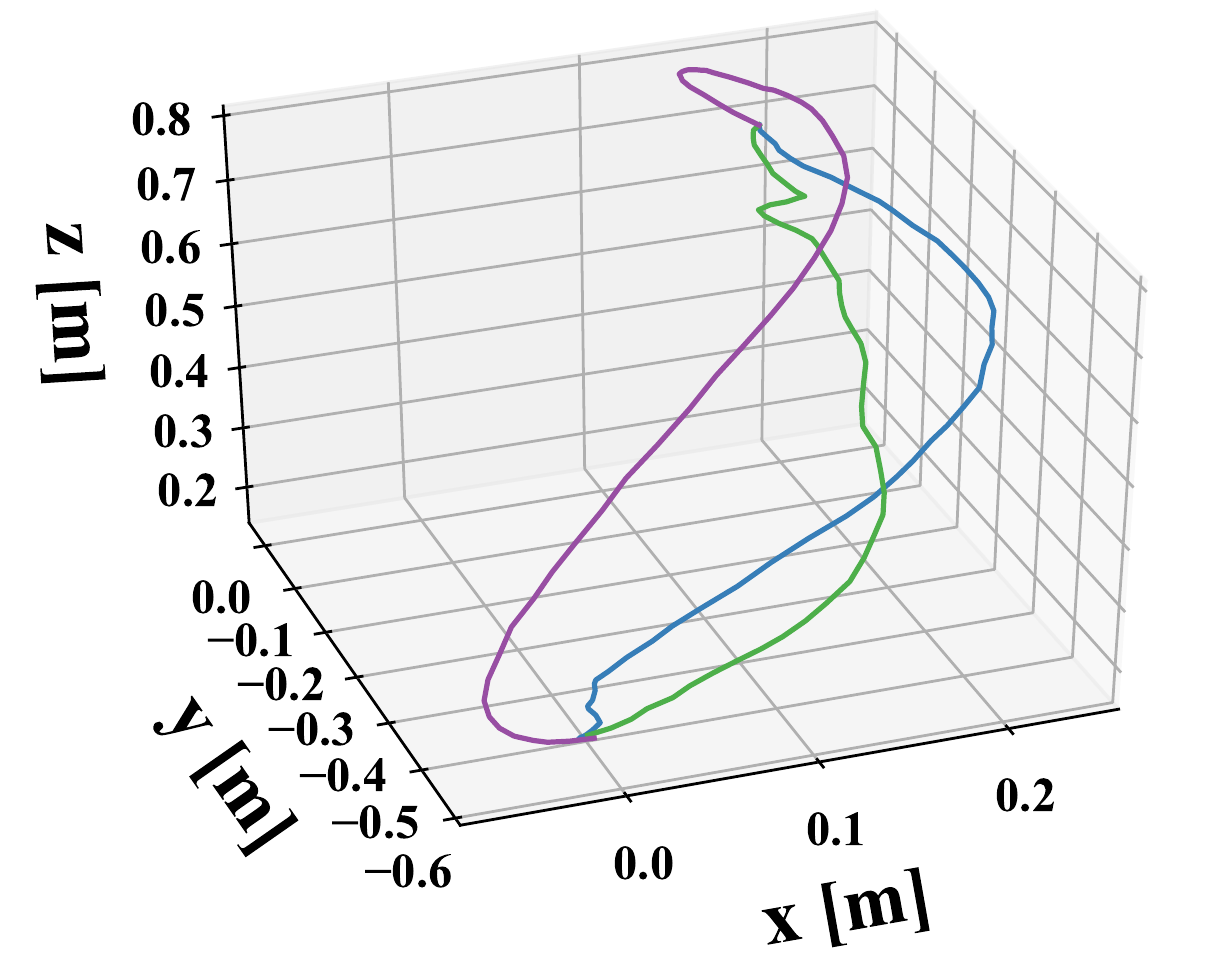}
		\caption{Trajectories of the end effector in task space on the task with UR3.  }
		\label{fig:UR_bottle_task}
	\end{subfigure}
	\hfill
	\begin{subfigure}[b]{\textwidth}
		\centering
		\includegraphics[width=\textwidth]{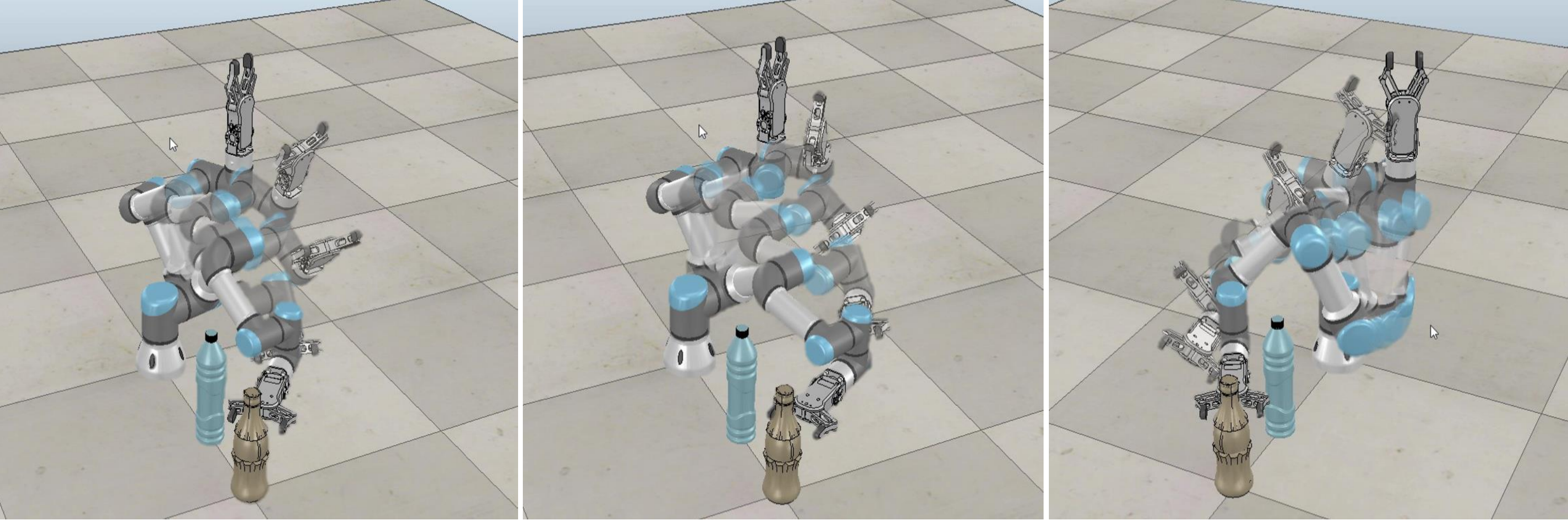}
		\caption{Results of the task with UR3. Three trajectories planned for the bottle grasping task are visualized.The values of the cost function for the obtained solutions are, from left to right, 3.74, 1.27, 2.24, respectively.}
		\label{fig:UR_bottle_sequence}
	\end{subfigure}
	\hfill
	\begin{subfigure}[b]{0.46\textwidth}
		\centering
		\includegraphics[width=\textwidth]{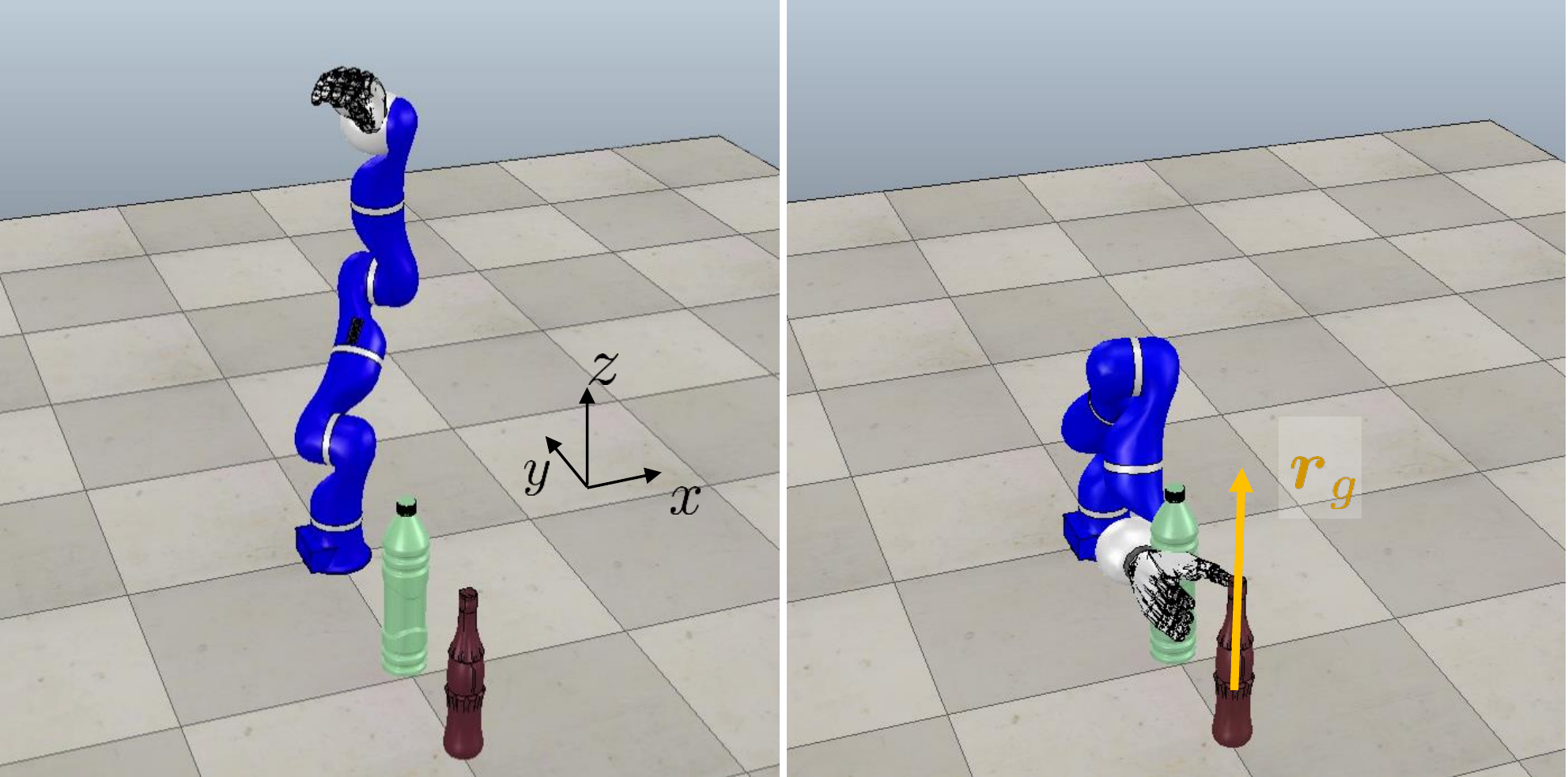}
		\caption{Problem setting for the bottle grasping task.  }
		\label{fig:Darias_bottle_setting}
	\end{subfigure}
	\hfill
	\begin{subfigure}[b]{0.46\textwidth}
		\centering
		\includegraphics[width=0.55\textwidth]{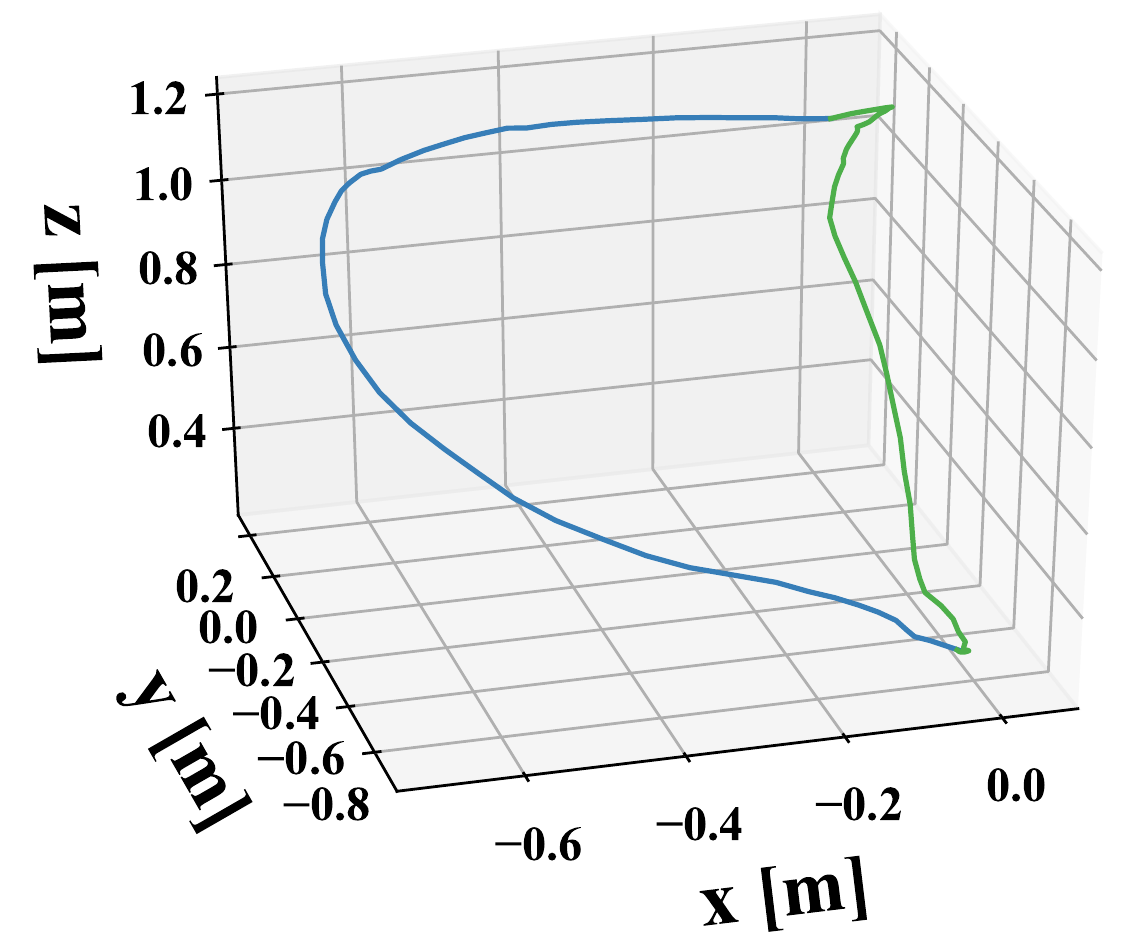}
		\caption{Trajectories of the end effector in task space on the task with UR3.  }
		\label{fig:Darias_bottle_task}
	\end{subfigure}
	\hfill
	\begin{subfigure}[b]{\textwidth}
		\centering
		\includegraphics[width=0.75\textwidth]{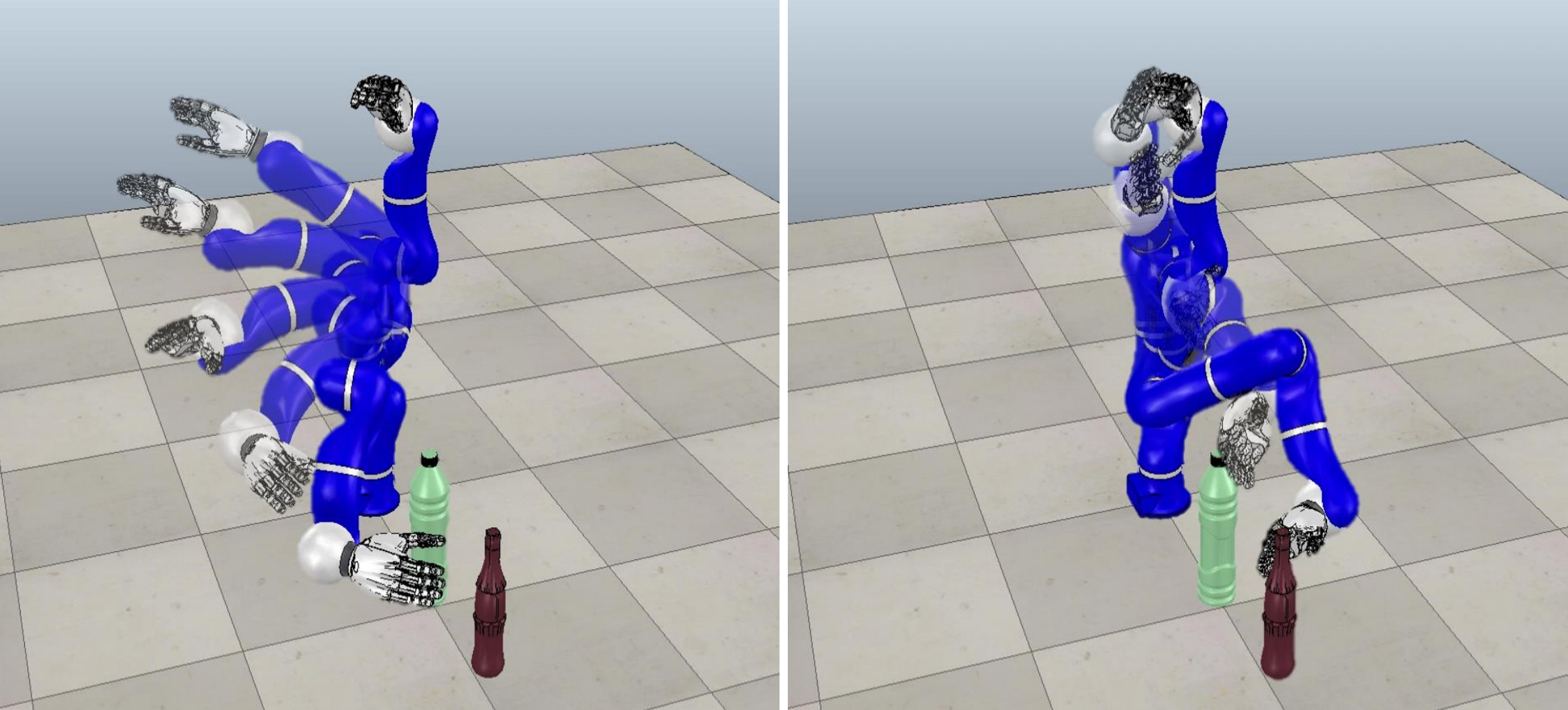}
		\caption{Results of the task with KUKA LWR. Two trajectories planned for the bottle grasping task are visualized.  The values of the cost function for the obtained solutions in the left and right figures are 1.83 and 3.01, respectively.}
		\label{fig:Darias_bottle_sequence}
	\end{subfigure}
	\caption{Experiments on tasks with a rotational freedom at the goal configuration. In (a) and (b), the left and right figures show the given start and goal configurations, respectively. We set the free rotational axis at the goal configuration as indicated as a yellow arrow in (a) and (b).}
	\label{fig:bag}
\end{figure*}

Task settings are shown in~Figs.~\ref{fig:UR_bag_setting} and \ref{fig:Darias_bag2_setting}. 
We set the free rotational axis at the goal configuration as indicated with a yellow arrow in Figs.~\ref{fig:UR_bag_setting} and \ref{fig:Darias_bag2_setting}.
In both tasks, the given goal configurations collide with an obstacle, and therefore the motion planning algorithm needs to find collision-free goal configurations.
The goal of this task is to plan a collision-free motion to reach a pre-grasp position for grasping the bag in the scene.
Please note that the size of the obstacles are different in Figs.~\ref{fig:UR_bottle_setting} and \ref{fig:Darias_bottle_setting}. Since the size of the manipulator is different, we used similar but different objects for UR3 and KUKA LWR.

As shown in Fig.~\ref{fig:UR_bag_sequence}, the proposed method finds two solutions for the task with UR3.
Likewise, our method finds two collision-free trajectories on the task with KUKA LWR as shown in Fig.~\ref{fig:Darias_bag2_sequence}.
Computation time was 63.2 seconds for the task with UR3 and 93.5 secs for the task with KUKA KWR.
The results are aligned with our intuition, that suggest that there should be approaches from both the right and left sides of the obstacle.

Another case study is shown in Fig.~\ref{fig:bag}.
The goal of the task is to plan a collision-free motion to reach a pre-grasp position for grasping the bottle presented in the scene as show in Figs.~\ref{fig:UR_bottle_setting} and \ref{fig:Darias_bottle_setting}.
As shown in Fig.~\ref{fig:UR_bottle_sequence}, three solutions are found for this task with UR3.
Two of the solutions approach the target object from the left-hand side, and the other one approaches the target from the right-hand side.
Our method also finds three trajectories for the task with KUKA LWR as shown in  Fig.~\ref{fig:Darias_bottle_sequence}.
Computation time was 28.4 seconds for the task with UR3 and 46.6 secs for the task with KUKA KWR.
We believe that the results presented in this section indicate that our method can find diverse solutions.

It is worth noting that our method finds multiple solutions even if the values of the cost function are not necessarily comparable. For example, regarding the solutions shown in Fig.~\ref{fig:Darias_bottle_sequence},
the cost of the solution shown in the right figure is clearly lower than that of the solution shown in the left figure.
This property of our method is attributed to the property of the density estimation with VBEM.
When performing Gaussian mixture fitting with VBEM, clusters of samples can be found even if the density of each cluster is different.
In our framework, this property enables us to find multiple solutions even if the value of the cost function corresponding to each solution is reasonably different.

\begin{table*}[!t]
	\small\sf\centering
	\caption{Performance of SMTO with different hyperparameters. The first three rows represent the combination of the hyperparameters. The averages over three executions of SMTO with different random seeds are shown. }
	\label{tbl:param_analysis}
		\begin{tabular}{cccccccccc}
			\multicolumn{10}{c}{Computation time [sec]} \\
			\hline
			$O$ &        5& 10 & 20 & 10 & 10 & 10 & 10 & 10 & 10 \\
			$N$ & 800& 800& 800& 500& 1000& 1500& 800& 800& 800 \\
			$\alpha$ & 20 & 20 & 20 & 20 & 20 & 20 &  5 & 10 & 50\\
			\hline
			\hline
			Task &&& &&& &&&  \\
			\hline
			UR3 (Fig. \ref{fig:UR_case_setting}) &40.9 & 42.1 &	45.0 & 29.4 & 52.3 & 77.8 &	71.4 & 70.6 & 44.3 \\
			\hline
			KUKA LWR (Fig. \ref{fig:Darias_bag_setting}) &56.7 & 58.8 & 62.1 & 28.9 & 72.1 & 78.8 & 115.5 & 56.6 & 42.7  \\
			\hline  
			UR3 (Fig. \ref{fig:UR_bag_setting})&88.9 & 89.8 & 99.2 & 63.6 & 93.5 &162.5 & 79.1 & 92.1 & 90.8  \\
			\hline
			KUKA LWR (Fig. \ref{fig:Darias_bag2_setting}) &42.1 & 44.0 & 46.5 &	28.5 & 54.4 & 78.0 & 43.8 & 44.2 & 44.3 \\
			\hline
			UR3 (Fig. \ref{fig:UR_bottle_setting})&58.5 & 60.2 & 80.1 & 31.0 & 95.1 & 131.4 & 108.4 & 92.4 & 42.8  \\
			\hline
			KUKA LWR (Fig. \ref{fig:Darias_bottle_setting}) &98.4 & 100.5 & 124.4 & 59.3 & 126.7 & 133.5 & 73.5 & 88.8 & 71.8 \\
		\end{tabular}
	
		\begin{tabular}{cccccccccc}
			\multicolumn{10}{c}{Number of solutions} \\
			\hline
			$O$ &        5& 10 & 20 & 10 & 10 & 10 & 10 & 10 & 10 \\
			$N$ & 800& 800& 800& 500& 1000& 1500& 800& 800& 800 \\
			$\alpha$ & 20 & 20 & 20 & 20 & 20 & 20 &  5 & 10 & 50\\
			\hline
			\hline
			Task &&& &&& &&&  \\
			\hline
			UR3 (Fig. \ref{fig:UR_case_setting}) &2.33 & 2.33 & 2.33 & 2.33 & 2.00 & 2.33 & 2.00 & 2.00 & 3.00  \\
			\hline
			KUKA LWR (Fig. \ref{fig:Darias_bag_setting}) &3.00 & 3.00 & 3.33 & 2.67 & 3.00 & 2.67 & 2.00 & 2.00 & 2.00 \\
			\hline  
			UR3 (Fig. \ref{fig:UR_bag_setting})&1.67 & 2.00 & 2.00 & 1.33 & 1.67 & 2.00 & 1.67 & 2.00 & 2.33 \\
			\hline
			KUKA LWR (Fig. \ref{fig:Darias_bag2_setting}) &2.33 & 2.33 & 2.67 & 2.00 & 2.33 & 3.00 & 3.00 & 2.67 & 2.33 \\
			\hline
			UR3 (Fig. \ref{fig:UR_bottle_setting})&2.67 & 3.00 & 2.33 & 2.33 & 2.33 & 1.67 & 1.33 & 2.33 & 3.00 \\
			\hline
			KUKA LWR (Fig. \ref{fig:Darias_bottle_setting})  &2.00 & 2.00 & 2.00 & 2.00 & 2.33 & 2.67 & 2.00 & 2.00 & 2.67 \\
		\end{tabular}
\end{table*}

\subsubsection{Effect of hyperparameters} 

To analyze the effect of each hyperparameter, we performed motion planning on the above-mentioned tasks with different $O$, $N$, and $\alpha$.
The results are summarized in Table~\ref{tbl:param_analysis}.
As SMTO is a stochastic method, the results differ slightly different with each execution;
we thus display the average over three trials with different random seeds in Table~\ref{tbl:param_analysis}.
As expected, increasing values of $N$ led to longer computation time, although the number of the solutions found by SMTO is not affected by $N$.
Likewise, when $O$ is larger, the computation time also gets longer, while the number of the solutions is not significantly affected by $O$.
This result indicates that density estimation using VBEM is not sensitive to $O$ in our framework.
The number of solutions was the most sensitive to $\alpha$ in our experiments.
This result is reasonable since $\alpha$ changes the landscape of the density induced by $d^{\cost}(\vect{\xi})$ in \eqref{eq:scale_f}.
As listed in Table~\ref{tbl:param_analysis}, larger values of $\alpha$ often lead to the more solutions
since larger values of $\alpha$ make the cost-weighted density estimation more sensitive to small gaps in the cost function.
To see the effect of the null space optimization presented in Section~\ref{sec:goal_opt}, we performed motion planning with SMTO without the null space optimization on the tasks in Figs.~\ref{fig:Darias_bag_setting} and \ref{fig:Darias_bottle_setting}. 
In this experiment, we set $O=10$, $N=800$ and $\alpha=20$. 

\begin{figure}[]
	\centering
	\begin{subfigure}[b]{0.24\textwidth}
		\centering
		\includegraphics[width=\textwidth]{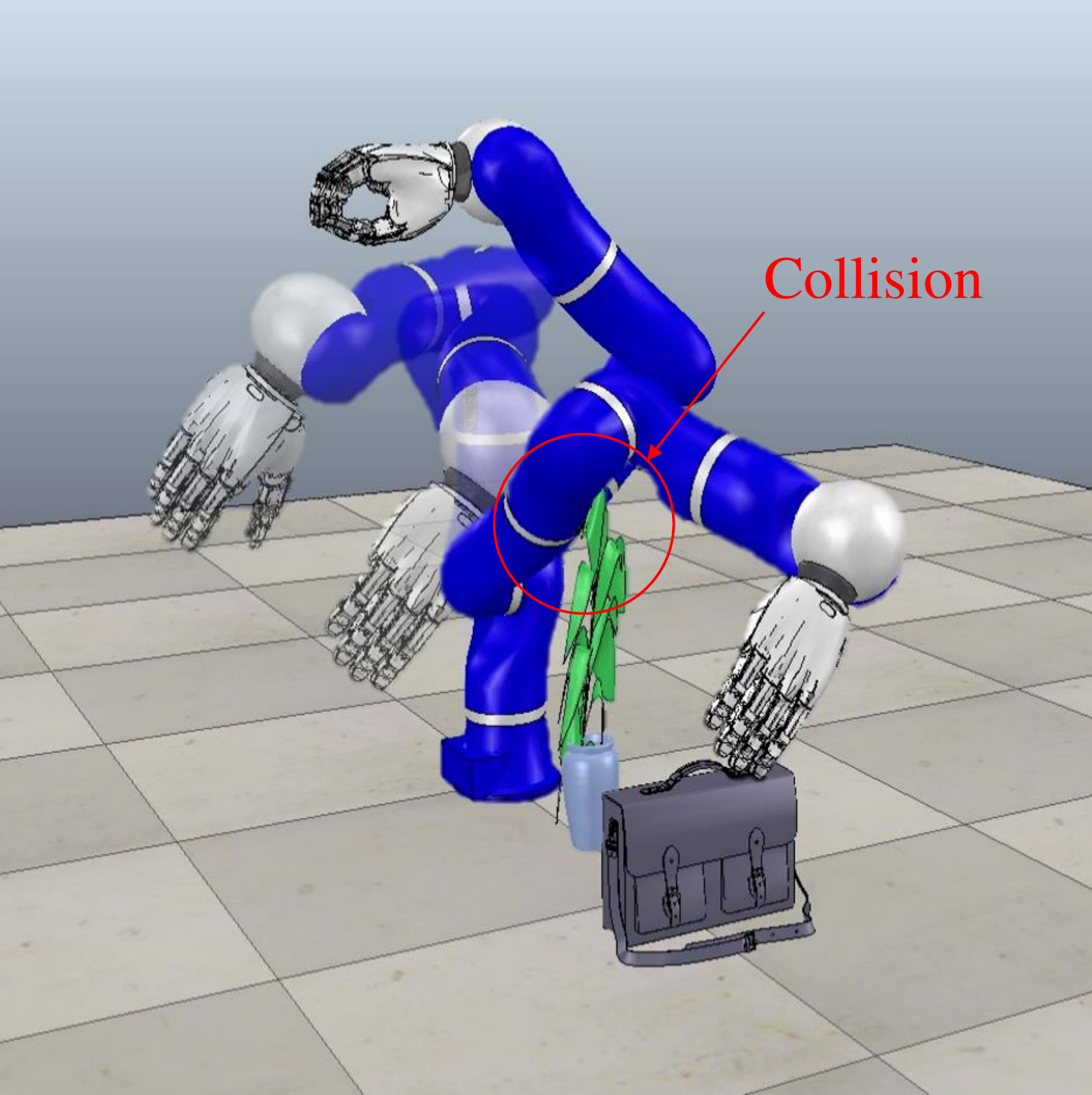}
		\caption{An example of local minima on the task in Fig.~\ref{fig:Darias_bag2_setting}. }
		\label{fig:null-off_bag}
	\end{subfigure}
	\hfill
	\begin{subfigure}[b]{0.24\textwidth}
		\centering
		\includegraphics[width=\textwidth]{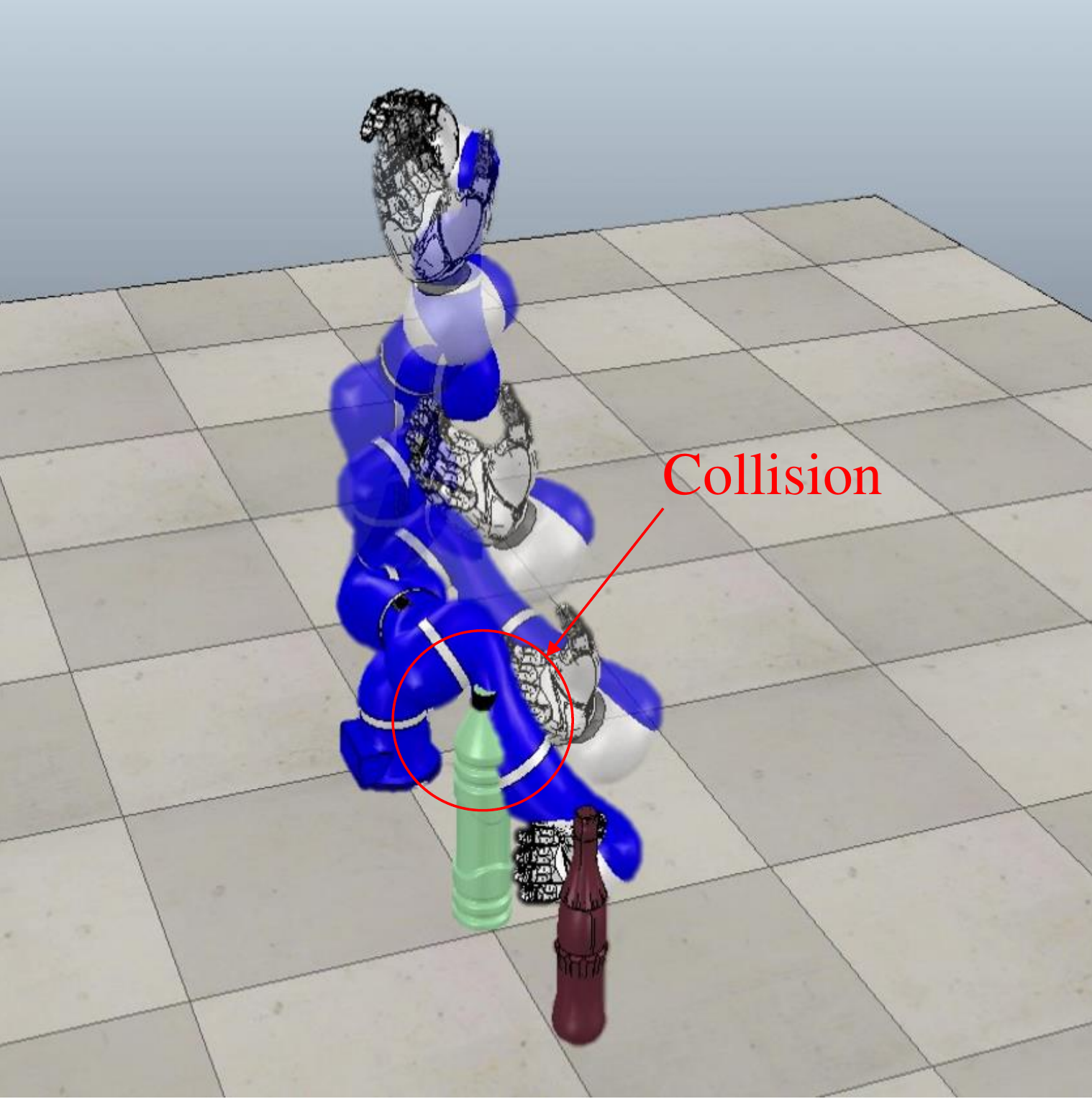}
		\caption{An example of local minima on the task in Fig.~\ref{fig:Darias_bottle_setting}.  }
		\label{fig:null-off_bottle}
	\end{subfigure}
	\caption{When the null space optimization in Section~\ref{sec:goal_opt} is not implemented, the trajectory optimization occasionally falls into local minima, resulting a fewer number of solutions.} 
	\label{fig:null-off}
\end{figure}

The average computation time over three trials with different random seeds for the tasks in Figs.~\ref{fig:Darias_bag_setting} and \ref{fig:Darias_bottle_setting} was  57.6~s and 117.0~s, respectively.
This result indicates that more computation time was necessary if the null space optimization was not implemented. 
The average number of the solutions found for the tasks in Figs.~\ref{fig:Darias_bag_setting} and \ref{fig:Darias_bottle_setting} were 1.33 and 2.33, respectively.
Without the null space optimization, the trajectory optimization occasionally falls into one of the local minima.
Examples of local minima on these tasks are shown in Fig.~\ref{fig:null-off}. 
As a consequence, SMTO finds fewer solutions for the task in Fig.~\ref{fig:Darias_bottle_setting} if the proposed null space optimization is not implemented.
For the task in Figs.~\ref{fig:Darias_bag_setting}, while variants of the solution corresponding to the left figure of Fig.~\ref{fig:Darias_bag2_sequence} were found, the solution corresponding to the right figure of Fig.~\ref{fig:Darias_bag2_sequence} was not found without the null space optimization.
Therefore, we think that null space optimization in Section~\ref{sec:goal_opt} improves the diversity of solutions found by SMTO.

These results indicate that our proposed method can optimize trajectories including the goal configuration by considering the multimodality of the cost function.
This functionality of our method can reduce the effort taken in manually tuning the goal configuration when deploying a robotic manipulator in practice.

\section{Discussion}

One can interpret that our approach divides the manifold of the trajectories based on the modes of the cost function and finds local solutions in each region. This concept is closely related to that of the reinforcement learning methods presented in~\citep{Ghosh18,Osa19}.
Our method alternates between gradient-free and gradient-based trajectory updates, and our gradient-free update is based on density estimation as described in Section~\ref{sec:MulTrajOpt}.
The gradient-based trajectory update can be viewed as a step to improve the proposal distribution for importance sampling for density estimation.

Another perspective on our method is that it alternates the model-free and model-based trajectory updates.
Alternation of the model-free and model-based updates appears in prior work on reinforcement learning~\citep{Chebotar17}.
In our method, the trajectory update based on the cost-weighted density estimation does not require explicit task information, except the costs of the sampled trajectories, while the gradient-based trajectory update explicitly requires the task information such as the position of the obstacles and the manipulator.
The combination of these trajectory updates enables flexible and efficient trajectory motion planning in the sense that the number of solutions are automatically determined and that the objective function can be directly optimized.

Recent studies have proposed sophisticated methods such as TrajOpt~\citep{Schulman14} or GPMP~\citep{Mukadam18} for optimizing a trajectory under constraints.
We do not claim that the proposed SMTO algorithm can replace other such motion planning framework.
Rather, the benefits of our proposed method will be enhanced in combination with such methods.
While we employed CHOMP-based optimization as a part of the motion planning in this study, the gradient-based optimization process could be replaced with other methods such as GPMP.
The contribution of our work is to throw lights on motion planning from multimodal optimization.
We show that multiple solutions can be found by considering the multiple modes of the cost function, which have often been ignored by prior studies.
To our knowledge, SMTO is the first method that explicitly considers the multimodality of the cost function and finds multiple trajectories that correspond to the modes of the cost function.

We think that our work could be used for task-level motion planning~\citep{Lozano-Perez89}, which aims to plan a sequence of sub-tasks.
The motion planning framework presented in this study was able to find multiple options for performing a sub-task.
When using our method for task-level motion planning, it is possible to select a trajectory among multiple options that optimizes the entire sequence of sub-tasks.
We think that this task-level motion planning with our motion planning framework is a promising way for achieving a complex task, which is hard to initiate with a single trajectory with the existing motion planning framework.
In addition, recent studies such as~\citep{Bobu18} make use of motion planners such as TrajOpt for sampling trajectories and inferring human intention within the context of human-robot interaction. 
Our method has the potential to contribute to such domains.

As our method SMTO was specifically designed to address motion planning problems, it is not applicable to control problems that are normally addressed by recent RL methods.
However, the concept of converting an optimization problem into a density estimation problem can be adapted to other domains.
For example, our prior work in \citep{Osa19} adapted this idea to deep hierarchical RL for learning option policies that correspond to the different modes of the Q-function. 
Extending the concept of cost-weighted density estimation to other problem settings should be addressed in future work.

\section{Conclusion}
In this study, we proposed the stochastic mutimodal trajectory optimization algorithm that determines multiple solutions in motion planning.
Our method identifies the multiple modes of the cost function and determines the trajectories corresponding to the modes of the cost function.
We derived this approach by formulating the trajectory optimization problem as a density estimation problem and introducing the importance sampling based on the cost function.
The proposed approach can be interpreted as a method that divides the manifold of the trajectory plan and determines solutions in each region.
Our framework enables users to select a preferable solution from multiple candidate trajectories,
and therefore eases the workload involved in tuning the cost function for obtaining a satisfactory solution.
We evaluated our proposed method with a 2D-three-link manipulator and manipulators with six and seven DoFs.
Our experiments show that our proposed algorithm was able to determine multiple solutions even when we used the cost function in a motion planning framework that finds only a single solution.
The results shed light on motion planning from the perspective of multimodal optimization.
In the future, we will investigate task-level motion planning based on the diverse behaviors obtained by our motion planning framework.

\begin{acks}
This work is partially supported by KAKENHI 19K20370.
\end{acks}

%
%
%


\appendix
\section{Implementation Details of VBEM with Importance Weights}
\label{sec:VBEM}

Here we describe details of  the VBEM algorithm with importance weights.
The basic implementation follows the algorithm described in \cite{Bishop06,Sugiyama15}.
For details related to the VBEM algorithm, please refer to \cite{Bishop06,Murphy12}.

For a given samples $\mathcal{D} = \{\vect{x}_{i}\}_{i=1}^{n}$, we consider the mixture of $m$ Gaussian models:
\begin{align}
q( \vect{x} | \mathcal{K}, \mathcal{M}, \mathcal{S} ) = \sum_{\ell=1}^{m} k_{\ell} \mathcal{N}( \vect{x} | \vect{\mu}, \vect{S}^{-1} ),
\end{align}
where $\mathcal{K} = \{k_1,...,k_m \}$ is a set of mixture coefficients, $\mathcal{M}=\{ \vect{\mu}_1,...,\vect{\mu}_m \}$ is a set of the means, and $\mathcal{S}=\{\vect{S}_1,...,\vect{S}_m \}$ is a set of the precision matrices.
Subsequently, we consider latent variables $O = \{ o_1,...o_n \}$ and the variational distribution $q(O)q(\mathcal{K}, \mathcal{M}, \mathcal{S})$. 
We choose a Dirichlet distribution as a prior for the mixing coefficients $\mathcal{K}$ and a Gaussian-Wishart distribution as a prior for each Gaussian component. 
The goal of VBEM is to maximize the lower bound of the marginal log likelihood $\log p(\mathcal{D})$ by repeating processes called VB-E and VB-M steps.

In the VB-E step, we compute the distribution of the latent variable $O$ from the current solution as
\begin{align}
q(O) = \prod_{i=1}^{n} \prod_{\ell = 1}^{m} \hat{\eta}^{o_{i \ell}}_{i,\ell}, 
\label{eq:VB-E}
\end{align}
where $\hat{\eta}_{i,\ell}$ is the responsibility of the sample $i$ on the $\ell$th cluster.  
The responsibility  $\hat{\eta}_{i,\ell}$ is given by 
\begin{align}
\hat{\eta}_{i,\ell}=\frac{ \hat{\rho}_{i,\ell}}{\sum_{\ell'=1}^{m} \hat{\rho}_{i,\ell'}},
\end{align}
where $\hat{\rho}_{i,\ell}$ is computed as
\begingroup\makeatletter\def\f@size{9}\check@mathfonts
\def\maketag@@@#1{\hbox{\m@th\large\normalfont#1}}
\begin{align}
& \hat{\rho}_{i,\ell}  = \exp \left( \psi(\hat{\alpha}_{\ell}) - \psi\left( \sum_{\ell'=1}^{m} \hat{\alpha}_{\ell'} \right) +
\frac{1}{2}\sum_{j=1}^{d} \psi \left( \frac{\hat{\nu_{\ell}} + 1 - j}{2}  \right) \right. \nonumber \\
& + \frac{1}{2} \log \det( \hat{\vect{K}}_{\ell} ) - \frac{d}{2\hat{\beta}_{\ell}} - \frac{\hat{\nu_{\ell}}}{2} 
( \vect{x}_{i} - \hat{\vect{h}}_{\ell} )\hat{\vect{K}}_{\ell}( \vect{x}_{i} - \hat{\vect{h}}_{\ell} )^{\top}
\Bigg),
\label{eq:responsibility}	
\end{align}
\endgroup

\noindent and $\psi(\alpha)$ is the digamma function defined as the log-derivative of the gamma function.

\begin{algorithm}[t]
	\caption{VBEM with Importance Weights }
	\begin{algorithmic}
		\STATE{
			\textbf{Input:} Data samples $\mathcal{D} = \{\vect{x}_{i}\}_{i=1}^{n}$, importance weights $\vect{w} = \{w_{1},...,w_n\}$, and hyperparameters
			$\alpha_{0}$, $\beta_{0}$, $\nu_0$, and $\vect{K}_{0}$  \\
			Initialize parameters \\
		}
		\REPEAT
		\STATE{ 
			VB-E step: \\
			\ \ \ \  Compute the distribution of the latent variable $q(O)$ \\
			\ \ \ \  by updating the responsibility $ \{\hat{\eta}_{i, \ell}\}_{i=1,\ell=1}^{n, \ \ \ m}$ with\\
			\ \ \ \  $ \{ \hat{\alpha}_{\ell}, \hat{\beta}_{\ell}, \hat{\vect{h}}_{\ell}, \hat{\nu}_{\ell}, \hat{\vect{K}}_{\ell} \}$ \\
			VB-M step: \\
			\ \ \ \  Compute the joint distribution $q(\mathcal{K}, \mathcal{M}, \mathcal{S})$ by \\ 
			\ \ \ \  updating $ \{ \hat{\alpha}_{\ell}, \hat{\beta}_{\ell}, \hat{\vect{h}}_{\ell}, \hat{\nu}_{\ell}, \hat{\vect{K}}_{\ell} \}$ with $ \{\hat{\eta}_{i, \ell}\}_{i=1,\ell=1}^{n, \ \ \ m}$ \\
			
		}
		\UNTIL{ convergence}
	\end{algorithmic}
	\label{alg:VBEM}
\end{algorithm}

In the VB-M step, we compute the joint distribution $q( \mathcal{K}, \mathcal{M}, \mathcal{S} )$
from the responsibilities $ \{\hat{\eta}_{i, \ell}\}_{i=1,\ell=1}^{n, \ \ \ m}$ and the weights $\{ w_i \}^{n}_{i=1}$.
In our framework, the weight $\{ w_i \}^{n}_{i=1}$ can be computed as $w_i = \tilde{W}(\vect{s}_i, \vect{\xi}_i)$. 
The joint distribution $q( \mathcal{K}, \mathcal{M}, \mathcal{S} )$ is given by
\begingroup\makeatletter\def\f@size{8.5}\check@mathfonts
\def\maketag@@@#1{\hbox{\m@th\large\normalfont#1}}
\begin{align}
q( \mathcal{K}, \mathcal{M}, \mathcal{S} )  =  \textrm{Dir}(\mathcal{K}|\hat{\alpha})
\prod_{\ell=1}^{m} 
\mathcal{N}( \vect{\mu}_{\ell} | \hat{\vect{h}}_{\ell}, ( \hat{\beta}_{\ell} \vect{S}_{\ell} )^{-1} ) \mathcal{W}( \vect{S}_{\ell} | \hat{K}_{\ell}, \hat{\nu}_{\ell} ),
\label{eq:VB-M}
\end{align}
\endgroup
where  
\begingroup\makeatletter\def\f@size{9}\check@mathfonts
\def\maketag@@@#1{\hbox{\m@th\large\normalfont#1}}
\begin{equation}
\hat{\gamma}_{\ell}  = \sum_{i=1}^{n} w_i \hat{\eta}_{i,\ell} , \ \ 
\hat{\vect{c}}_{\ell} = \frac{1}{\hat{\gamma}_{\ell}} \sum_{i=1}^{n} w_i \hat{\eta}_{i,\ell} \vect{x}_{i}, \ \
\hat{\vect{h}_{\ell}} = \frac{\hat{\gamma_{\ell}}}{\hat{\beta_{\ell}}} \hat{\vect{c}}_{\ell}, \nonumber
\end{equation}
\endgroup
\begin{equation}
\hat{\alpha}_{\ell}  = \alpha_{0} + \hat{\gamma_{\ell}}, \ \
\hat{\beta_{\ell}} = \beta_{0} + \hat{\gamma_{\ell}}, \ \
\hat{\nu_{\ell}} = \nu_0 +  \hat{\gamma_{\ell}}, \nonumber
\end{equation}
{\scriptsize
	\begin{equation}
	\hat{\vect{K}}_{\ell}  = \left( \vect{K}_0^{-1} + \sum_{i=1}^{n} w_i \hat{\eta}_{i, \ell} ( \vect{x}_{i} - \hat{\vect{c}}_{\ell} )( \vect{x}_{i} - \hat{\vect{c}}_{\ell} )^{\top} + \frac{\beta_0\hat{\gamma}_{\ell}}{\beta_0 + \hat{\gamma}_{\ell}} \hat{\vect{c}}_{\ell}\hat{\vect{c}}_{\ell}^{\top} \right)^{-1}, \nonumber
	\end{equation}	
}

\noindent and $\mathcal{W}( \vect{S} | \vect{K}, \nu )$ is the Wishart density with $\nu$ degrees of freedom.
$\alpha_{0}$, $\beta_{0}$, $\nu_0$, and $\vect{K}_{0}$ are hyperparameters.

\end{document}